\documentclass[10pt,twocolumn,letterpaper]{article}

\usepackage[pagenumbers]{wacv} %

\usepackage{graphicx}
\usepackage{booktabs}
\usepackage{xspace}
\usepackage{url}
\usepackage{cuted}   %

\usepackage{tabularx}
\usepackage{float}
\usepackage{siunitx}
\usepackage{subcaption}
\usepackage{wrapfig}
\usepackage{pifont}
\usepackage{multirow}
\usepackage{makecell}
\usepackage[table]{xcolor}
\newcommand{\C}[1]{} %
\newcommand{\cmark}{\ding{51}}
\newcommand{\xmark}{\ding{55}}
\newcolumntype{Y}{>{\raggedright\arraybackslash}X}
\newcolumntype{C}{>{\centering\arraybackslash}p{1.35cm}}
\newcommand{\DatasetName}{DF26\xspace}
\newcommand{\ci}[2]{#1 {\scriptsize\textcolor{gray}{[#2]}}}

\usepackage{xcolor}
\usepackage[most]{tcolorbox}
\usepackage{fontawesome5}

\definecolor{hfyellow}{HTML}{FFD21E}   %
\definecolor{hfamber}{HTML}{E5A400}    %
\definecolor{badgefill}{HTML}{FFF7DB}  %
\definecolor{badgetext}{HTML}{4A3800}  %

\newtcbox{\hfbadge}{on line, arc=3pt, boxrule=0.6pt,
  colback=badgefill, colframe=hfamber,
  boxsep=0pt, left=7pt, right=7pt, top=2.5pt, bottom=2.5pt}

\newcommand{\badgelink}[2]{%
  \begingroup\hypersetup{urlcolor=badgetext}\href{#1}{#2}\endgroup}

\newcommand{\datasetbadge}[1]{%
  \hfbadge{\raisebox{-0.13em}{\includegraphics[height=0.85em]{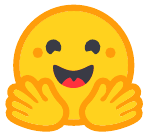}}%
  \hspace{4pt}\badgelink{#1}{\bf \small DF26 on Hugging Face}}}

\definecolor{wacvblue}{rgb}{0.21,0.49,0.74}
\usepackage[pagebackref,breaklinks,colorlinks,allcolors=wacvblue]{hyperref}

\def\wacvPaperID{2036} %
\def\confName{WACV}
\def\confYear{2027}

\title{DF26: We Cannot Tell Fake From Real Anymore}

\author{
Severyn Shykula\textsuperscript{1}, Andrii Yermakov\textsuperscript{3}, Ivan Samarskyi\textsuperscript{3}, Dmytro Mishkin\textsuperscript{2,3},
\\Jan Cech\textsuperscript{3}, Anastasiia Mishchuk\textsuperscript{2,4}
\\{\small \textsuperscript{1}Ukrainian Catholic University} \quad {\small\textsuperscript{2}Hover Inc., USA}
\\{\small \textsuperscript{3}VRG, Faculty of Electrical Engineering, Czech Technical University in Prague}
\\{\small\textsuperscript{4}Institute of Software Systems of the National Academy of Sciences of Ukraine}
\\{\tt\small shykula.pn@ucu.edu.ua} \quad {\tt\small \{yermaand,mishkdmy,cechj\}@fel.cvut.cz} \quad {\tt\small MishchukA@nas.gov.ua}\\
[4pt]\datasetbadge{https://huggingface.co/datasets/DF26/DF26}
}

\begin{document}
\maketitle
\begin{abstract}
We introduce \DatasetName\footnote{\raggedright\url{https://huggingface.co/datasets/DF26/DF26}}, a novel benchmark for detecting AI‑generated videos containing fully synthetic clips produced by recent text-to-video and image-to-video models. The videos capture single-person public-speaking scenarios, spanning direct‑to‑camera recordings, official statements, and studio interviews -- 271 real and 2,420 synthetic videos generated by seven modern video models.
The study on \DatasetName shows that human accuracy in detecting AI-generated videos is 52.6\% while recognizing 76.0\% of real ones, and that state-of-the-art deepfake detectors, using the current evaluation setup, perform close to random chance.
Our results highlight the limitations of current evaluation protocols and motivate the need for benchmarks that explicitly measure robustness to modern generative model distribution shifts.

\end{abstract}
    
\section{Introduction}
\label{sec:intro}

The rapid progress of video generative models has substantially improved the realism of visual content, making deepfake detection increasingly challenging. Recent text-to-video and image-to-video generators produce photorealistic videos that differ drastically from legacy benchmarks organized primarily around face swapping, face reenactment, and facial attribute manipulation. 
One particularly high-risk setting is single-person public-speaking video. We define this setting as a video in which a single person appears to address an audience, a camera, or an interviewer. 

Present benchmarks only partially cover this threat. Modern detectors report strong performance on legacy datasets such as FaceForensics++ \cite{FF++} and DFDC \cite{dolhansky2020deepfakedetectionchallengedfdc}, but they often fail to generalize to unseen generators. Recent benchmarks cover speech-driven or avatar-based synthesis, but they do not isolate full-scene text-to-video and image-to-video generation in controlled public-speaking contexts. Deepfake-Eval-2024~\cite{Deepfake-Eval-2024} finds that existing academic benchmarks are no longer representative of ``in-the-wild'' deepfakes circulating on social media. 

This task is difficult not only for automated tools. Human accuracy in identifying AI-generated videos is near chance, as shown by a meta-analysis of 56 studies~\cite{DIEL2024100538}. This additionally supports the motivation for the creation of reliable automated evaluation methods, especially in high-risk public scenarios.  

\begin{figure*}[t]
\centering
\setlength{\tabcolsep}{2pt}
\renewcommand{\arraystretch}{1.05}

\begin{tabular}{cccccc}
\includegraphics[width=0.155\textwidth]{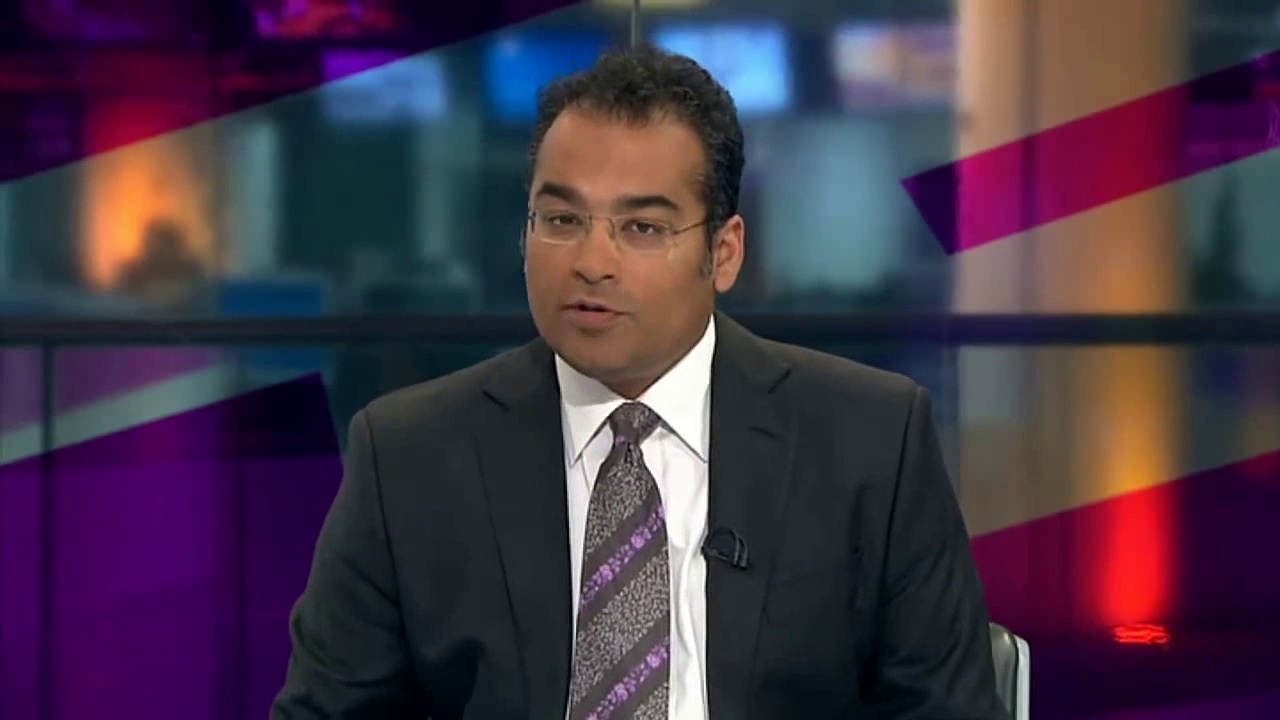}
&
\includegraphics[width=0.155\textwidth]{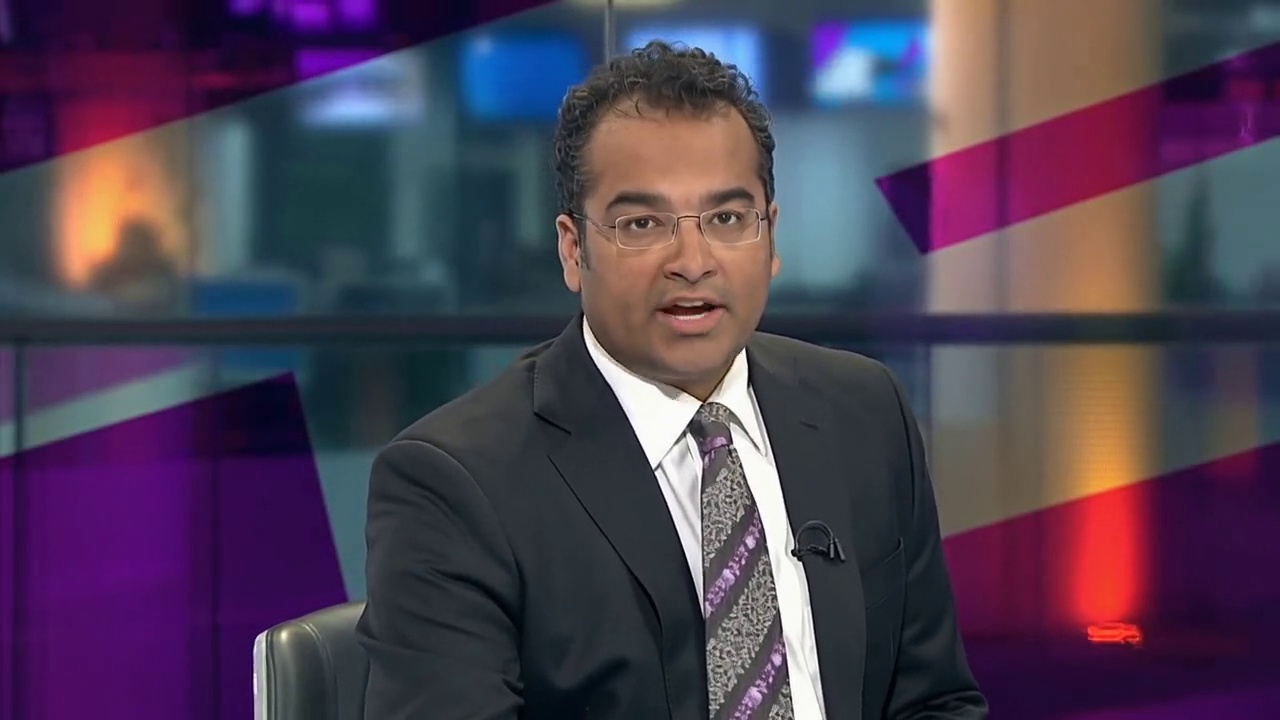}
&
\includegraphics[width=0.155\textwidth]{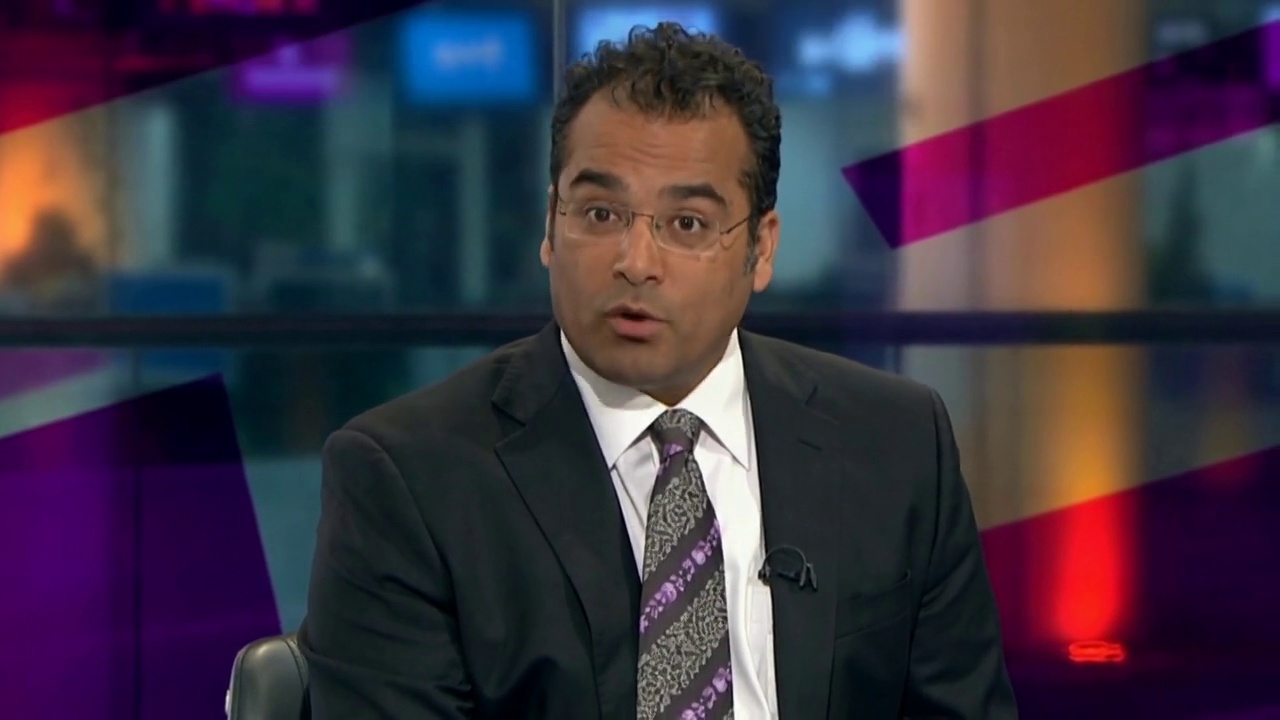}
&
\includegraphics[width=0.155\textwidth]{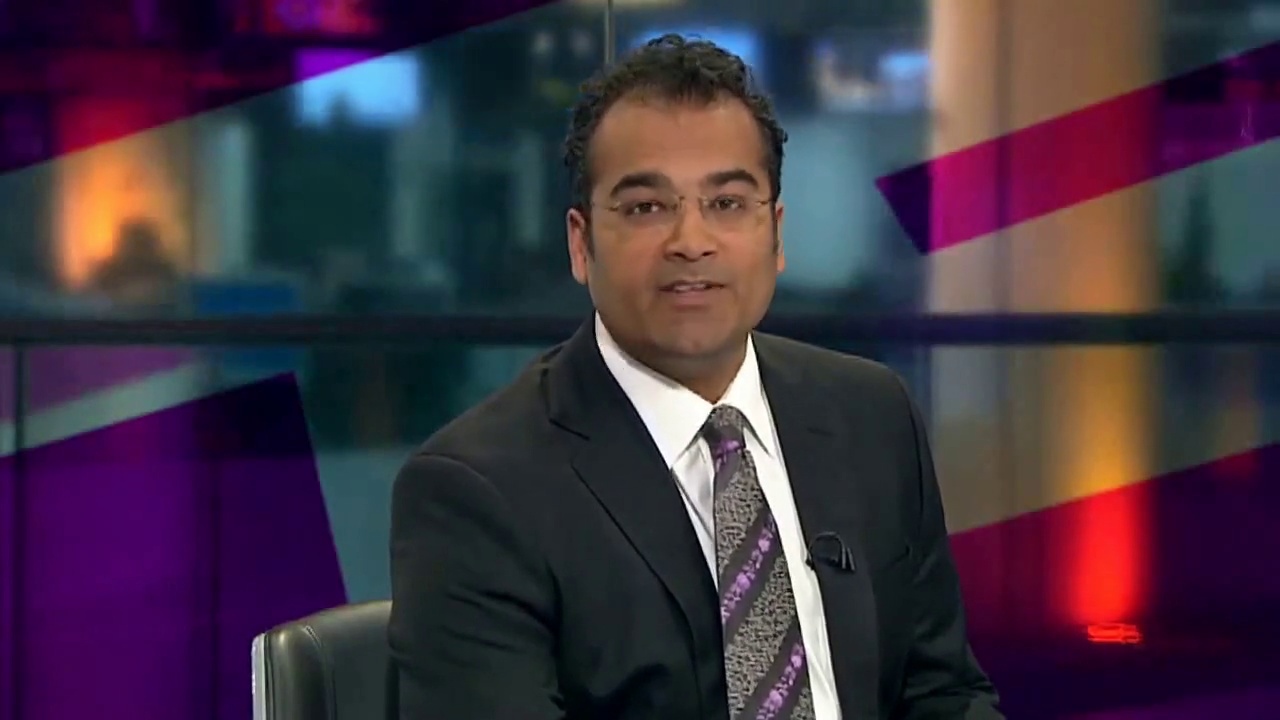}
&
\includegraphics[width=0.155\textwidth]{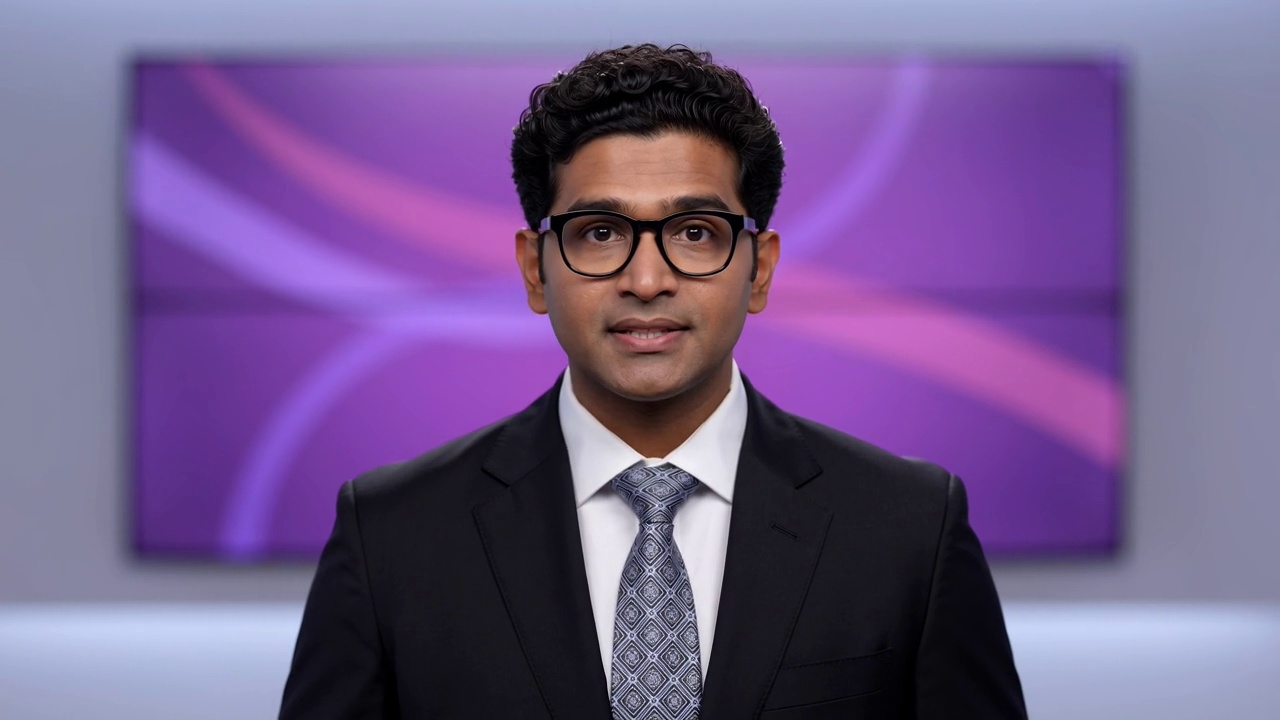}
&
\includegraphics[width=0.155\textwidth]{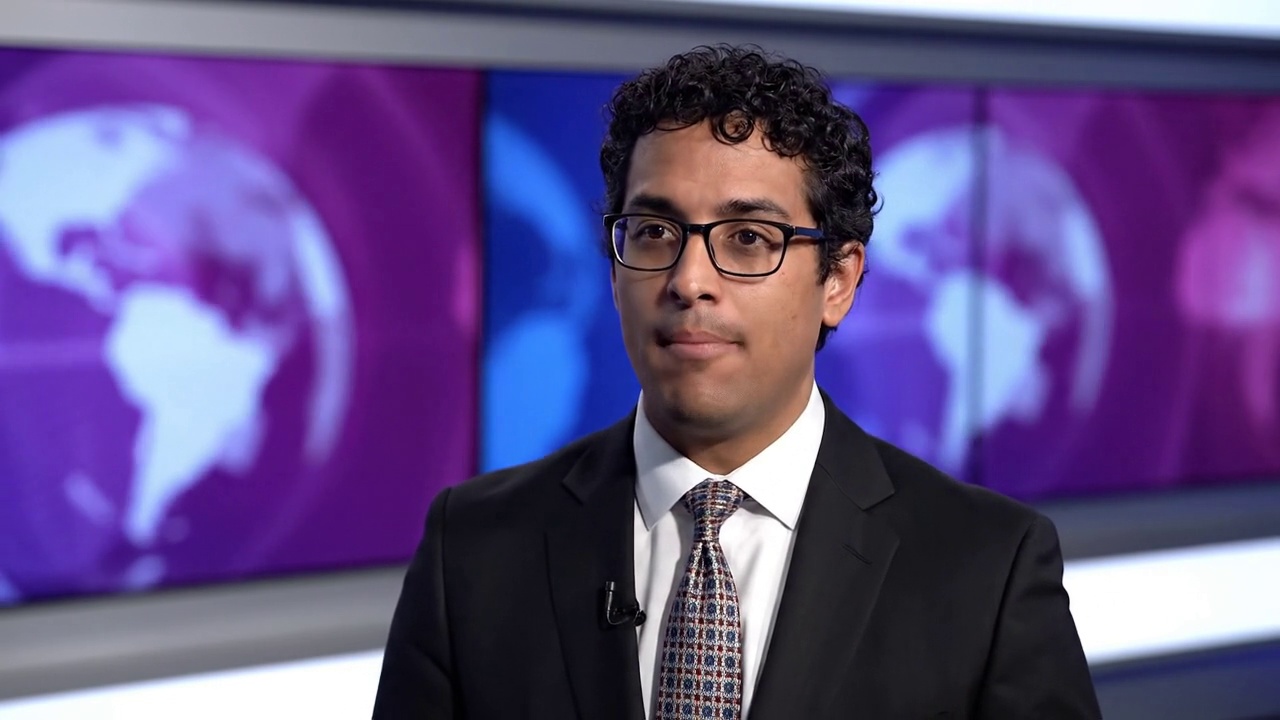}
\\[-1pt]
Real
& HunyuanVideo I2V
& LTX I2V
& Wan 2.2 I2V
& Grok Imagine 1.0
& Veo 3.1
\\[4pt]

&
\includegraphics[width=0.155\textwidth]{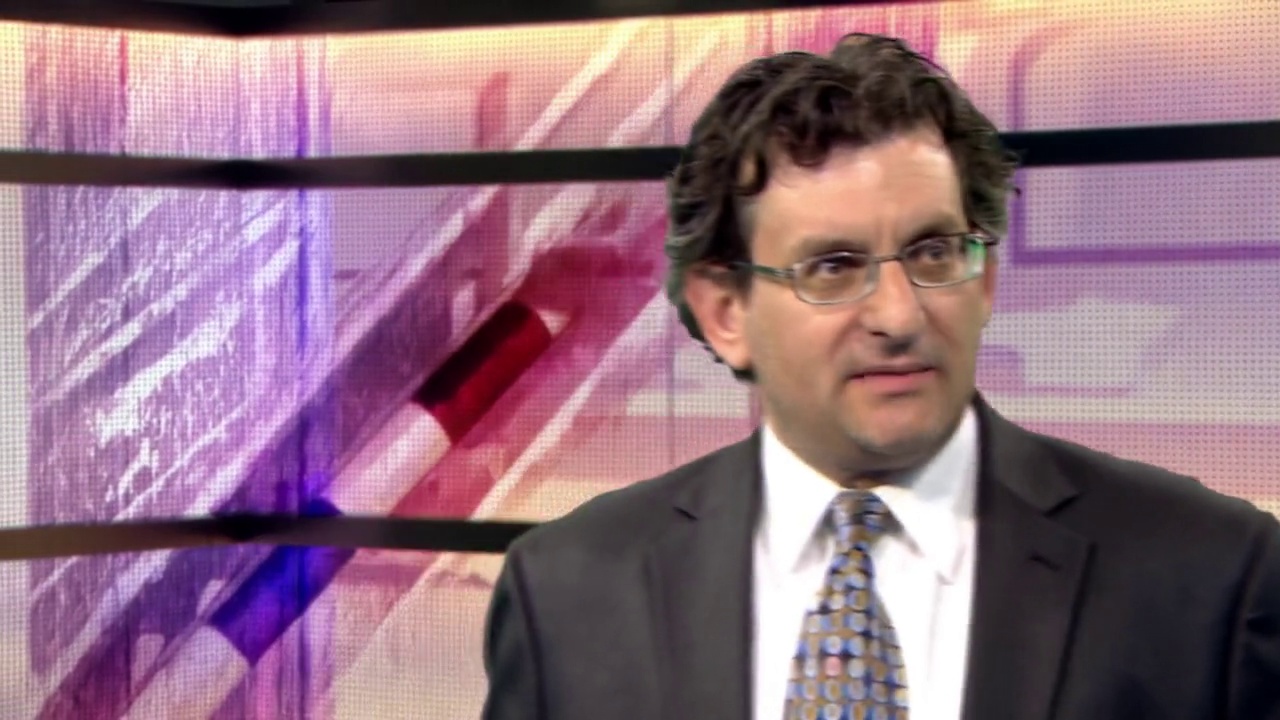}
&
\includegraphics[width=0.155\textwidth]{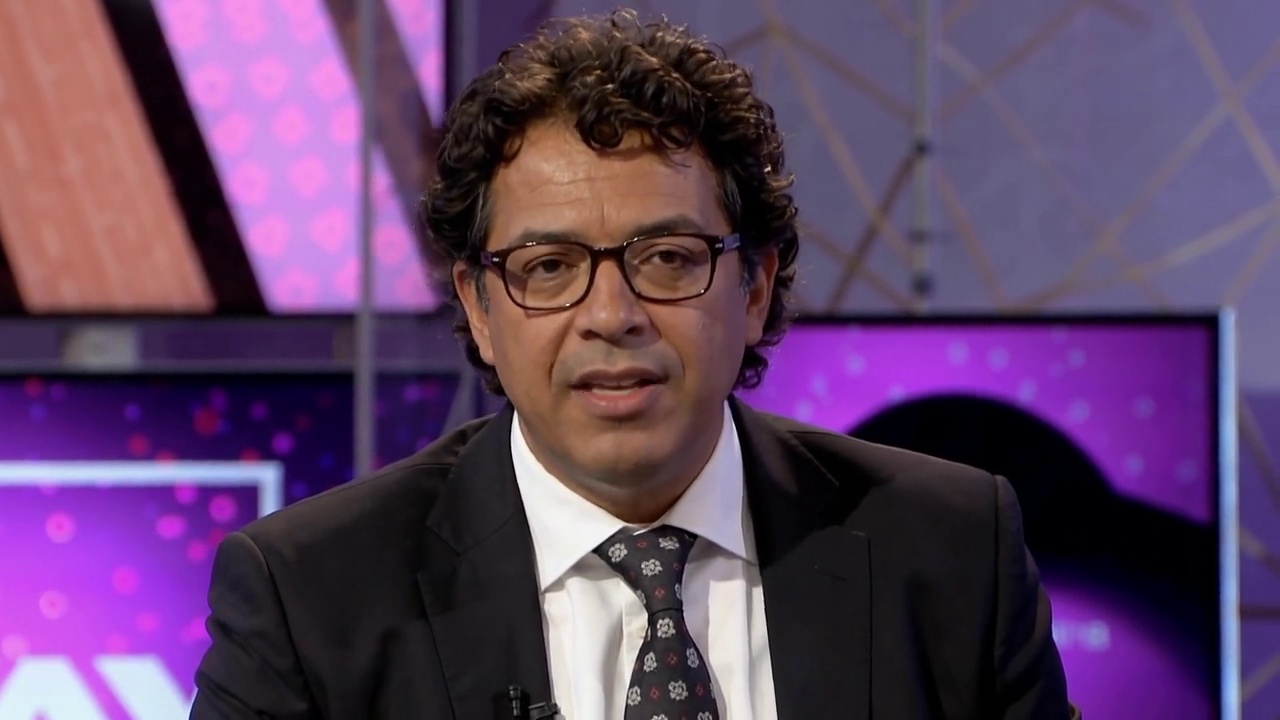}
&
\includegraphics[width=0.155\textwidth]{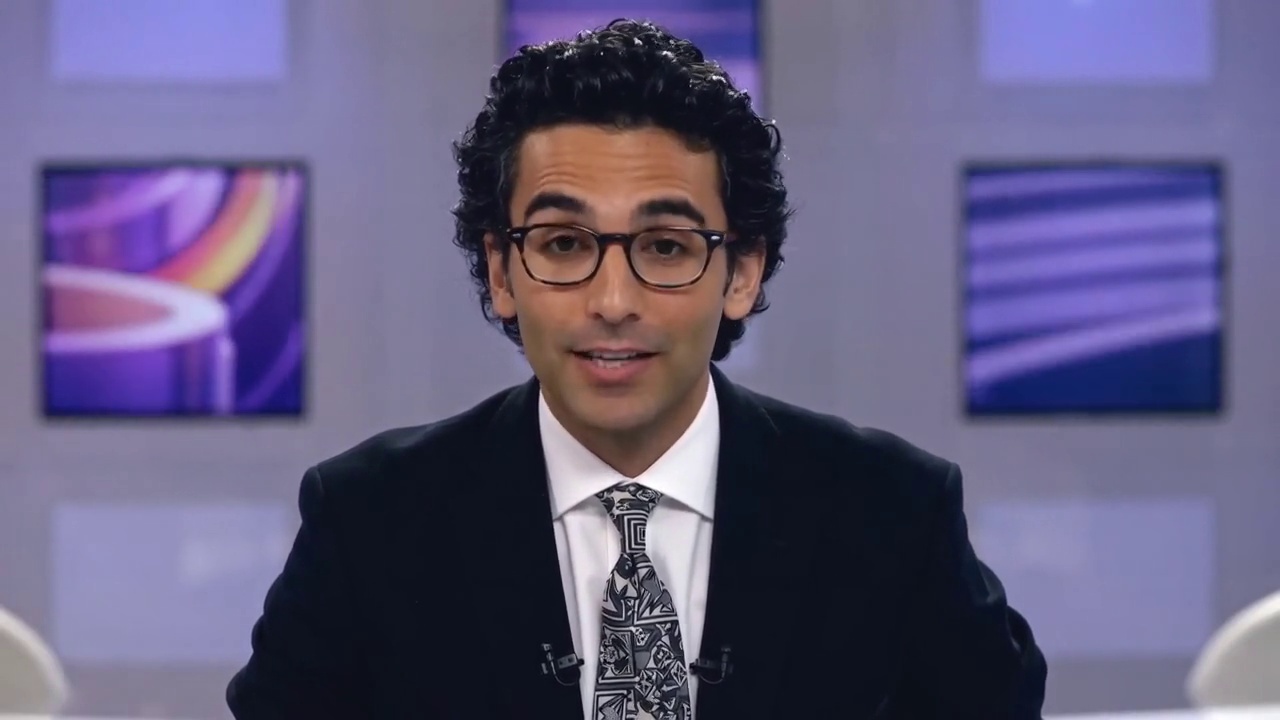}
&
\includegraphics[width=0.155\textwidth]{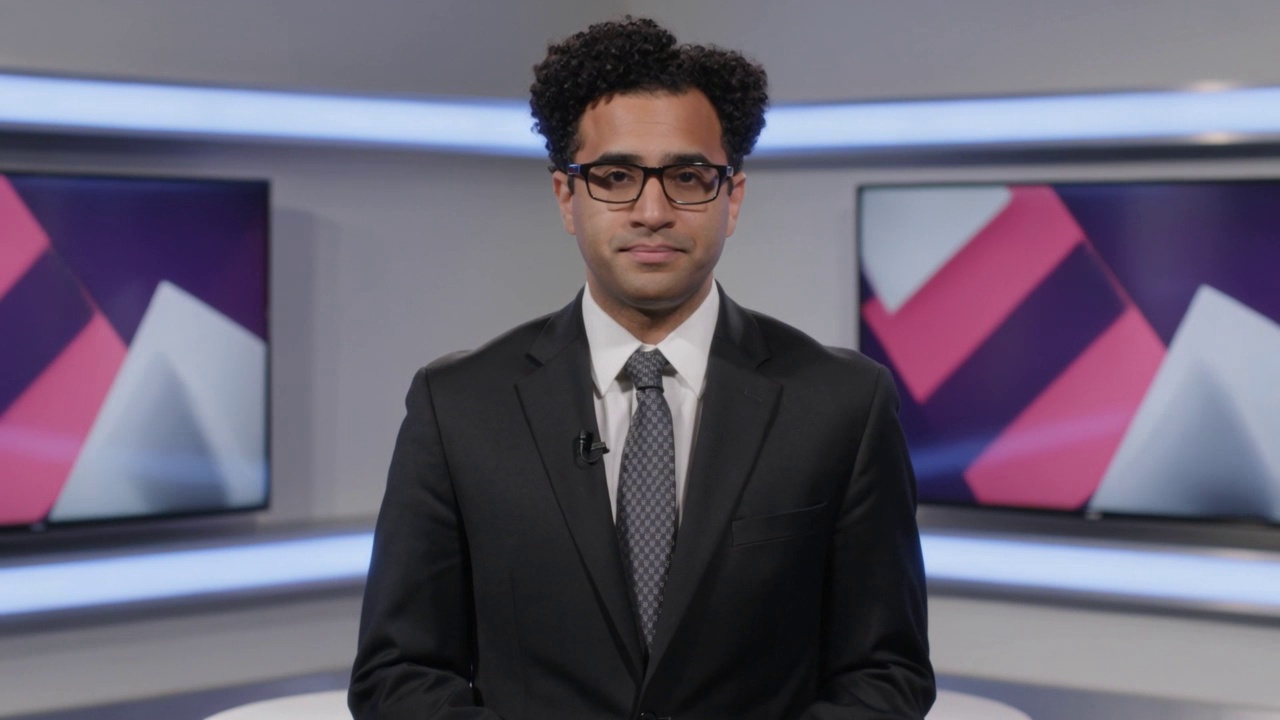}
&
\includegraphics[width=0.155\textwidth]{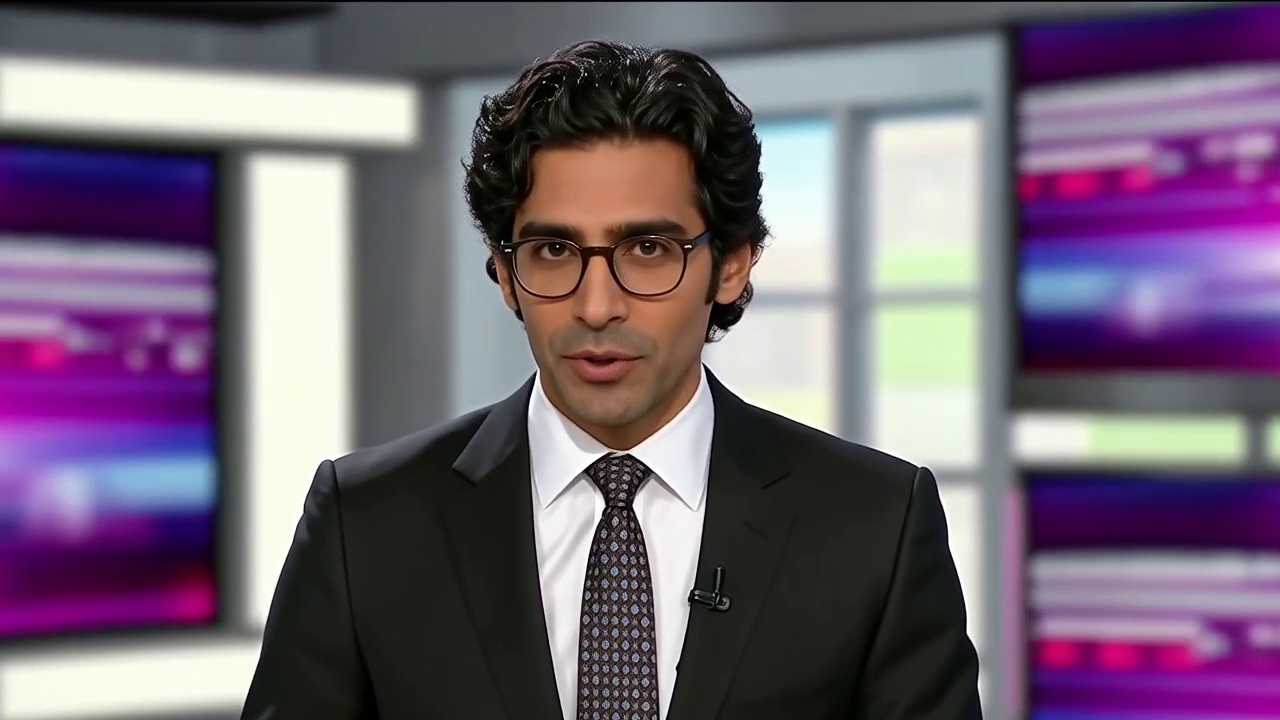}
\\[-1pt]
&
HunyuanVideo T2V
& LTX T2V
& Wan 2.2 T2V
& Kling 3.0
& Wan 2.6
\\
\\

\includegraphics[width=0.155\textwidth]{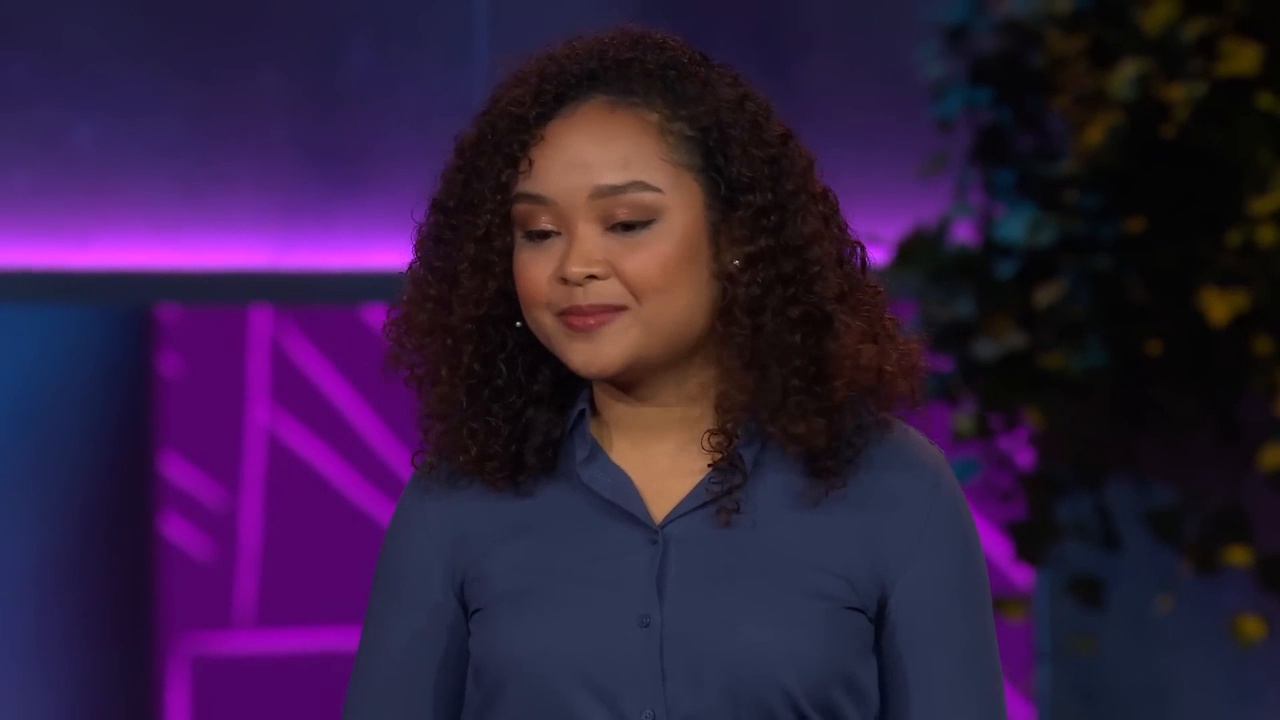}
&
\includegraphics[width=0.155\textwidth]{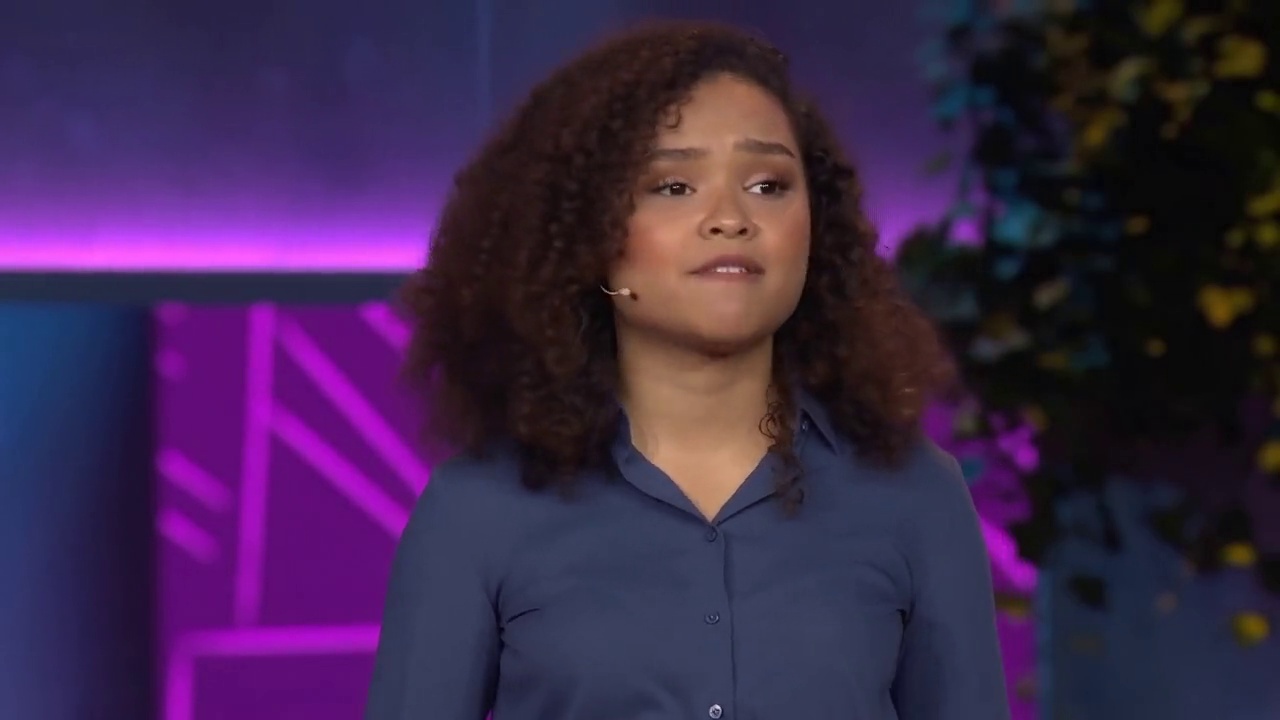}
&
\includegraphics[width=0.155\textwidth]{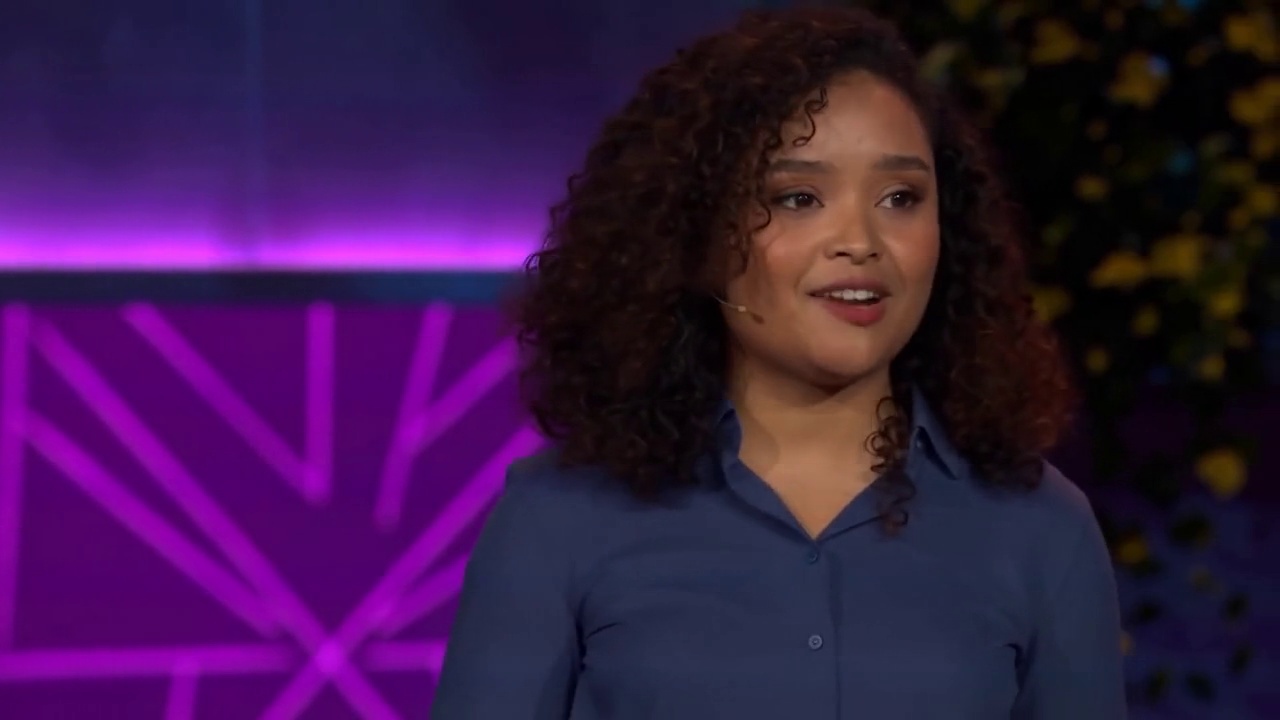}
&
\includegraphics[width=0.155\textwidth]{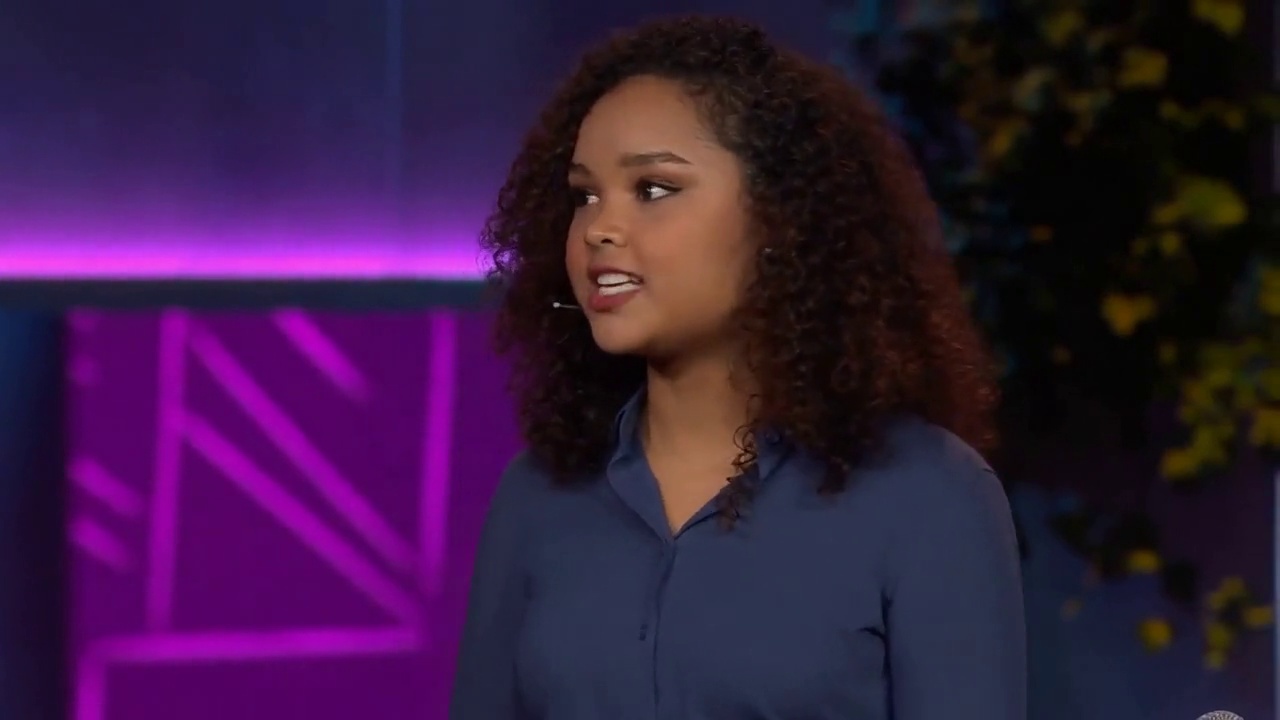}
&
\includegraphics[width=0.155\textwidth]{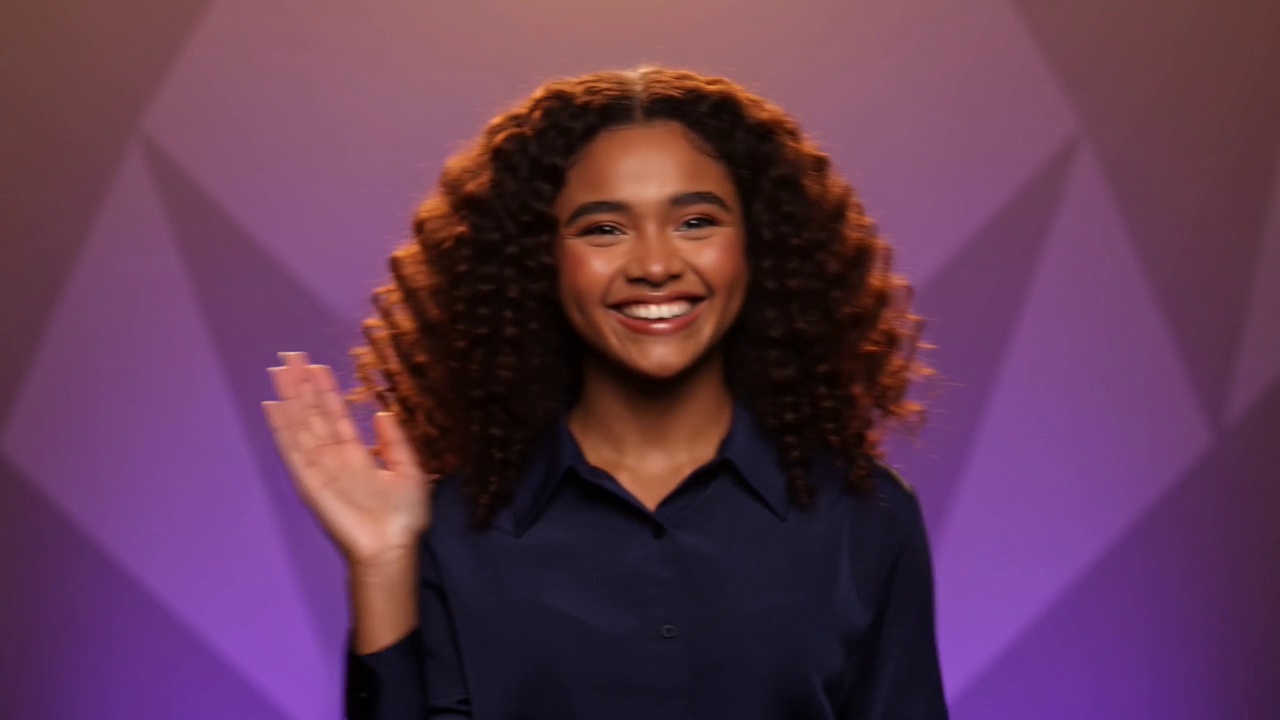}
&
\includegraphics[width=0.155\textwidth]{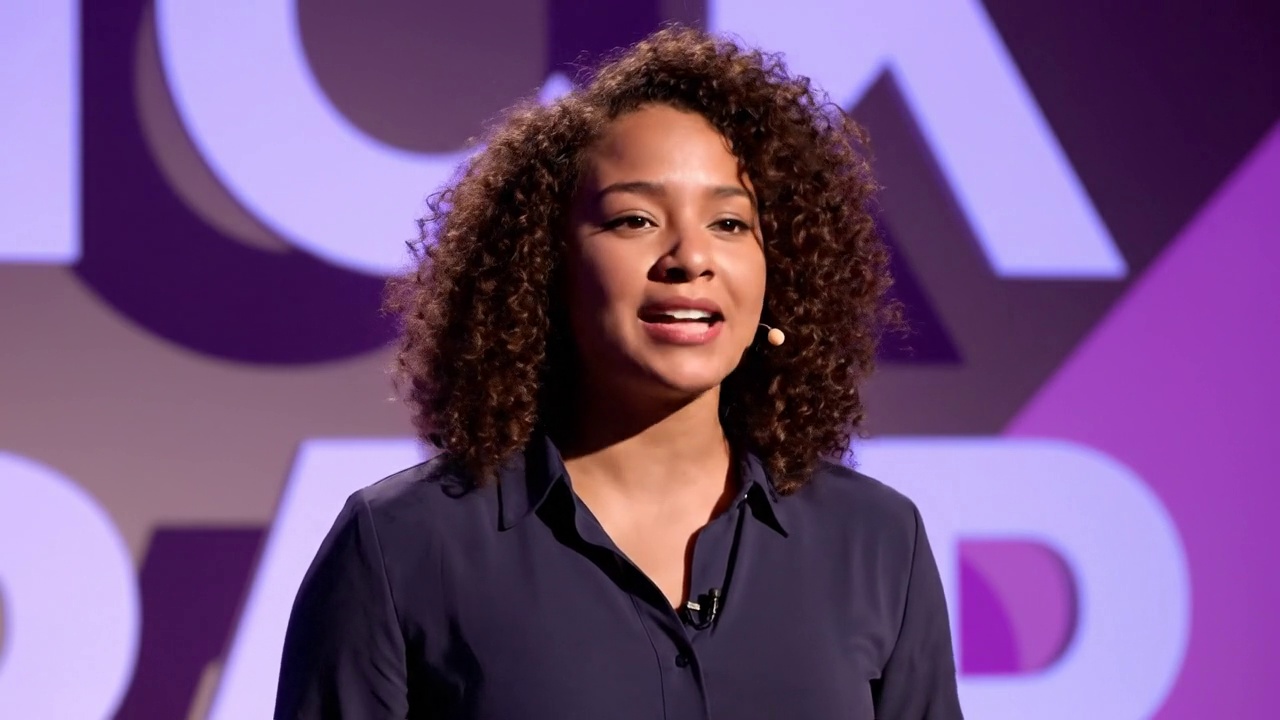}
\\[-1pt]
Real
& HunyuanVideo I2V
& LTX I2V
& Wan 2.2 I2V
& Grok Imagine 1.0
& Veo 3.1
\\[4pt]

&
\includegraphics[width=0.155\textwidth]{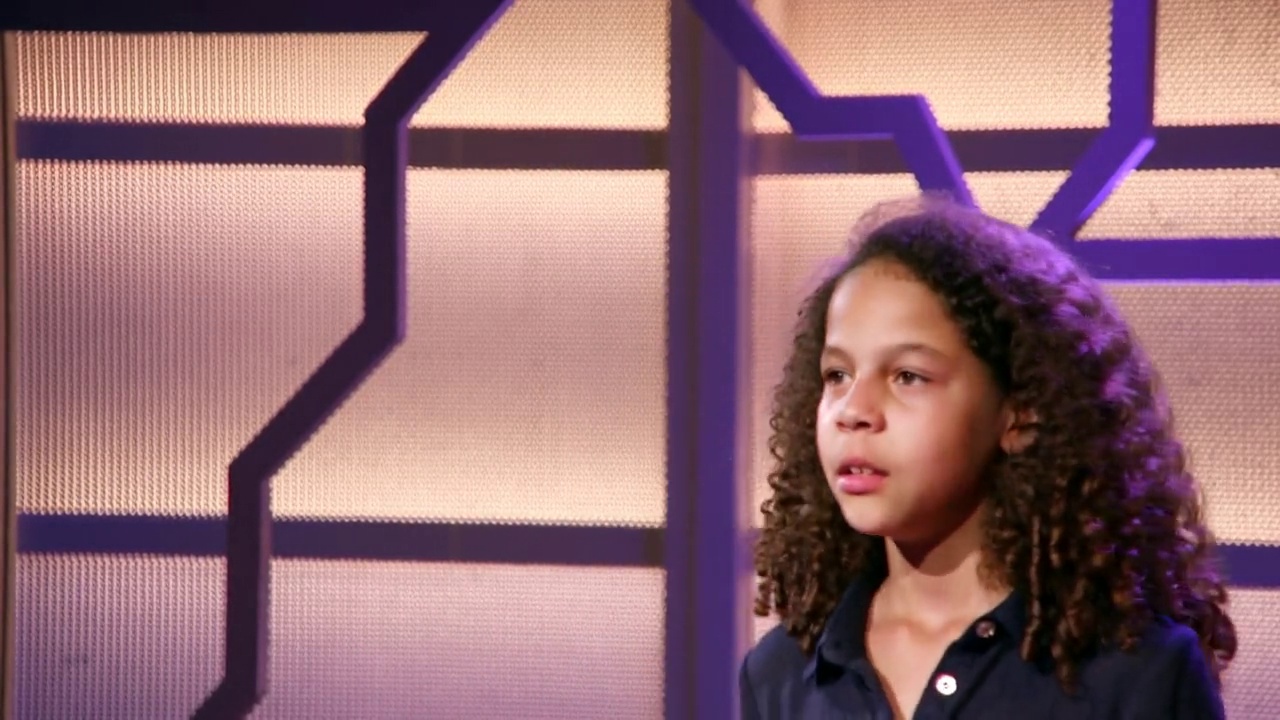}
&
\includegraphics[width=0.155\textwidth]{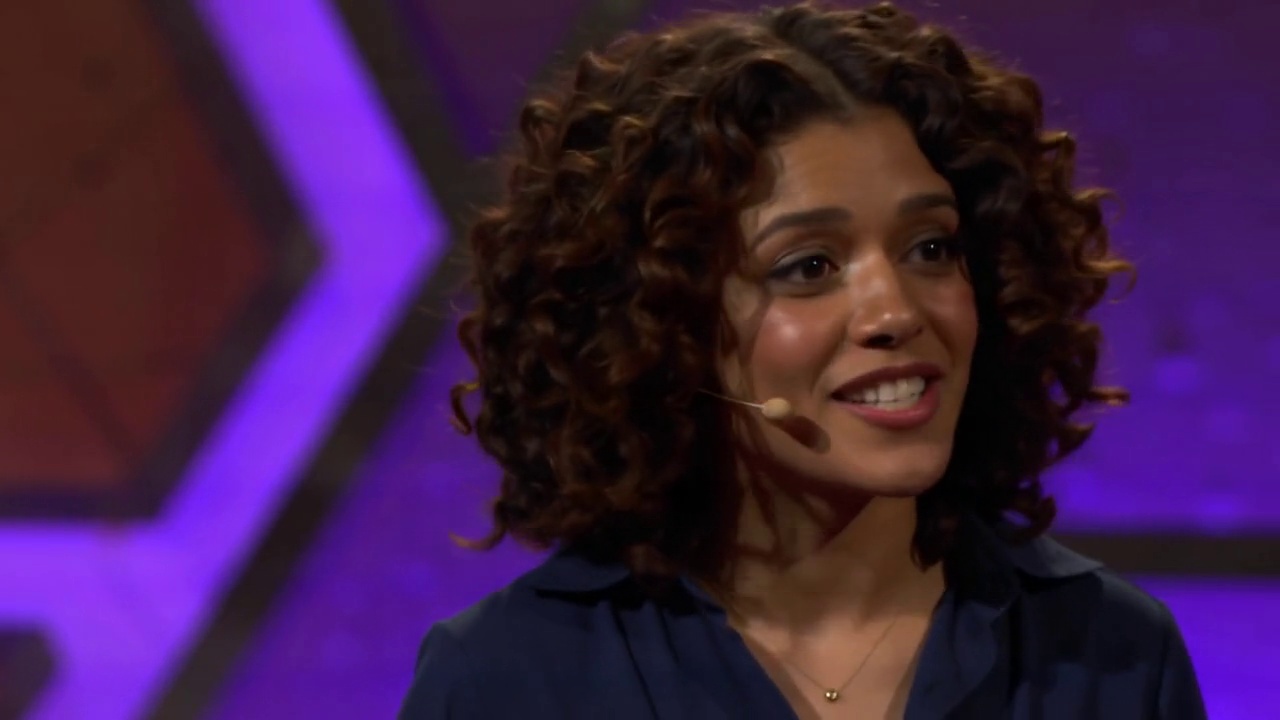}
&
\includegraphics[width=0.155\textwidth]{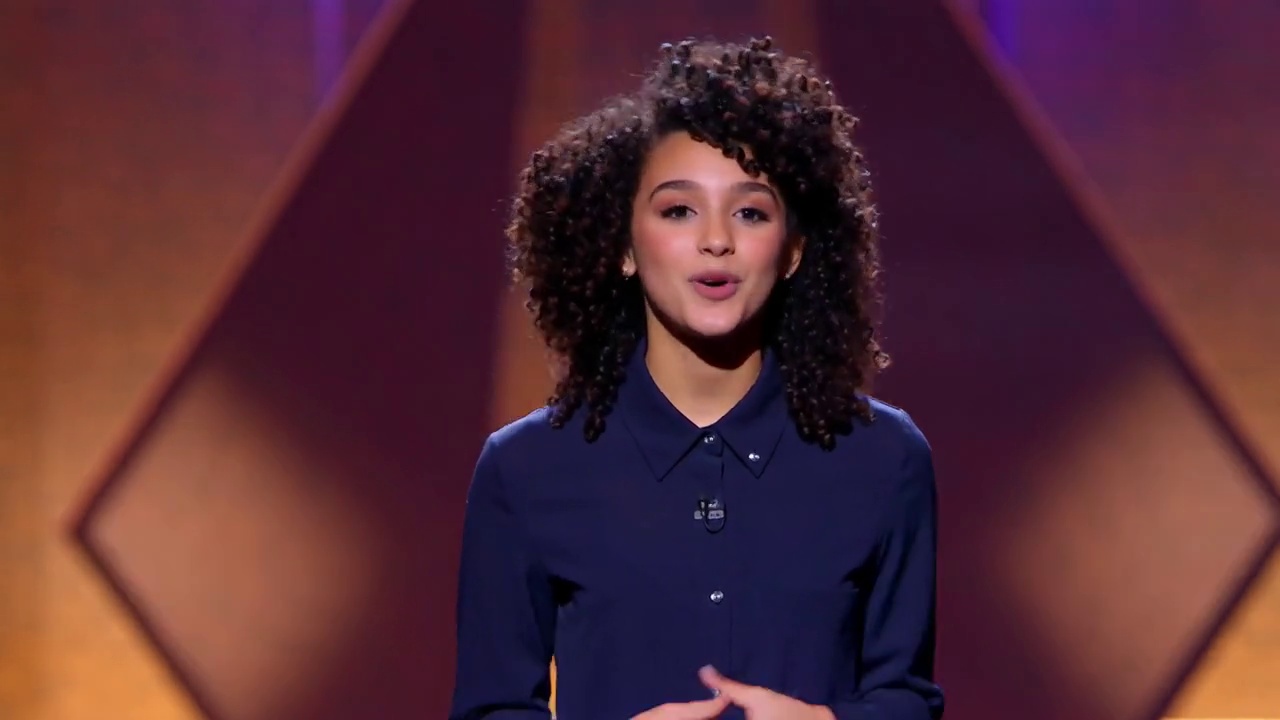}
&
\includegraphics[width=0.155\textwidth]{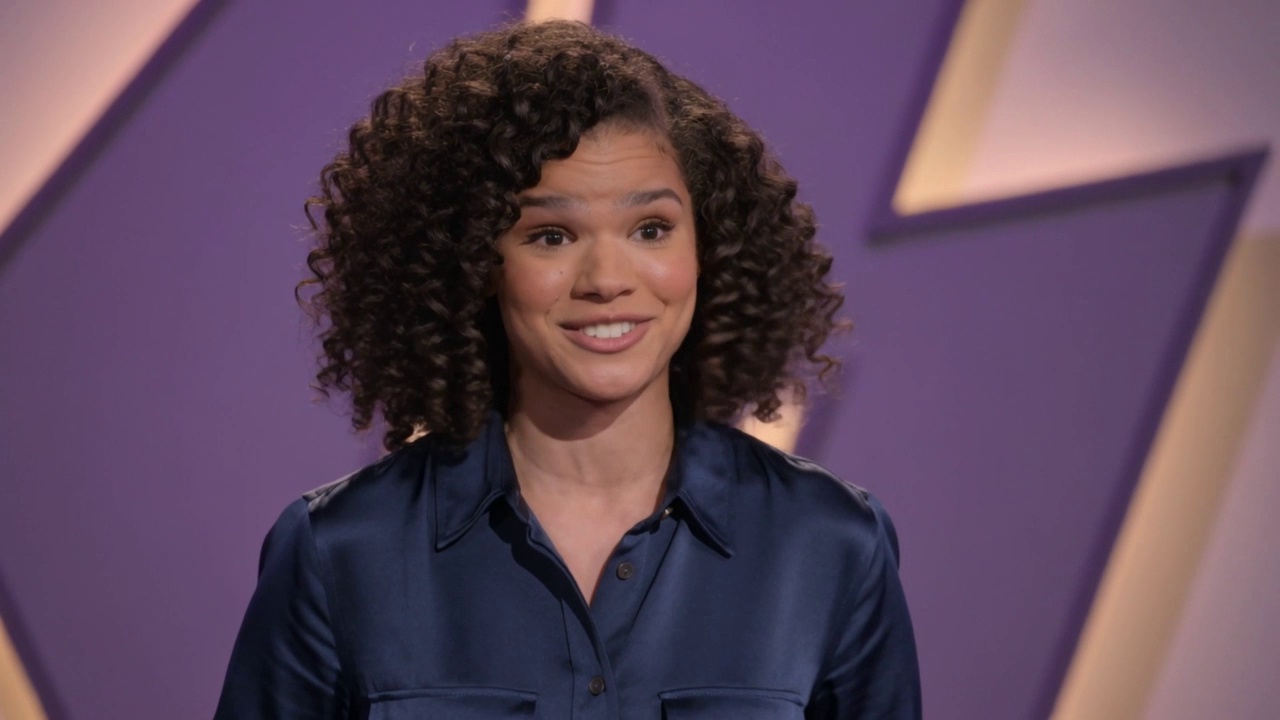}
&
\includegraphics[width=0.155\textwidth]{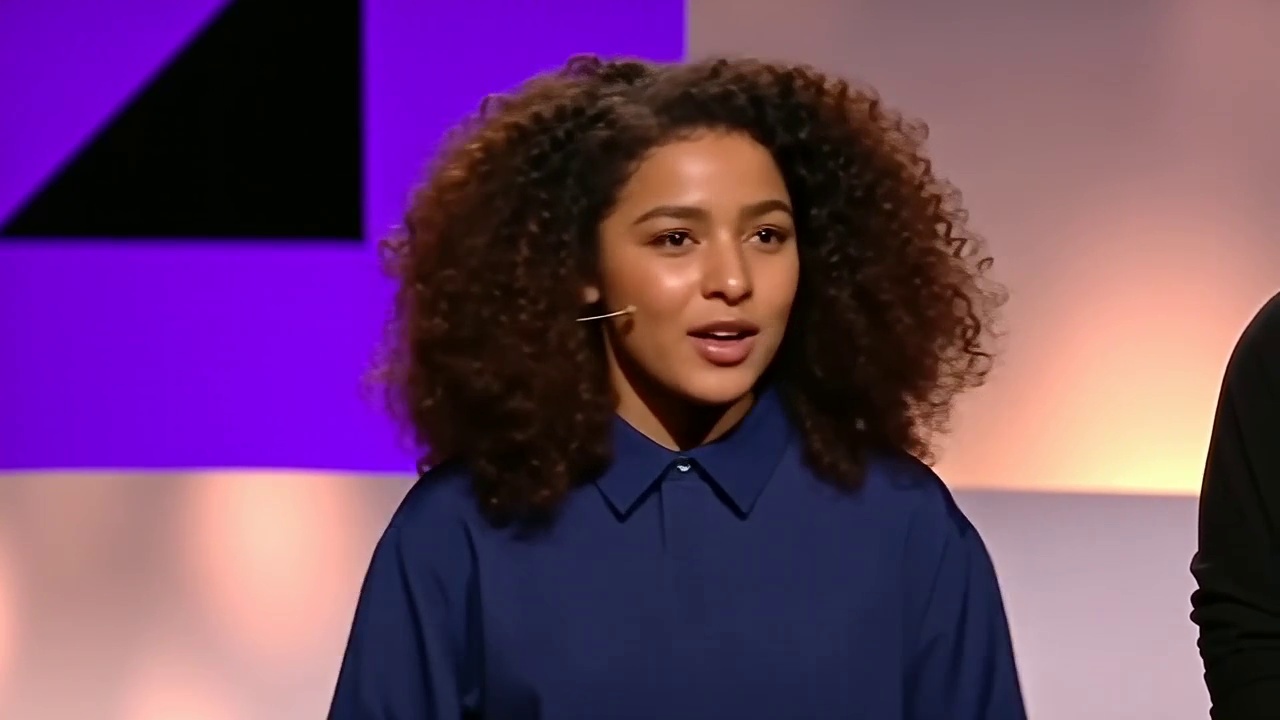}
\\[-1pt]
&
HunyuanVideo T2V
& LTX T2V
& Wan 2.2 T2V
& Kling 3.0
& Wan 2.6
\\

\end{tabular}

\caption{
Samples from \DatasetName generated in image-to-video (I2V) and text-to-video (T2V) setups.
In T2V, videos are generated from semantic prompts derived from the frames of the corresponding real video; in I2V, videos are generated from the first frame of the corresponding real video. All presented samples are taken from the last frames of every video.}
\label{fig:t2v_examples}
\end{figure*}

\begin{table}[t]
\centering
\caption{\label{tab:video-generators-used}Video generators used for the creation of \DatasetName. Elo is estimated by Artificial Analysis as of Aug 2026.}
\begin{tabular}{llrr}
\toprule
\textbf{Generator}    &\textbf{Creator} &\textbf{Released}   &\textbf{Elo} \\
\midrule
\multicolumn{4}{l}{\textit{Closed-source}} \\
\midrule
Wan 2.6    &Alibaba &Dec 2025   &1,185\\
Veo 3.1    &Google &Jan 2026  & 1,213 \\
Grok Imagine 1.0    &SpaceXAI &Jan 2026   &1,221 \\
 Kling 3.0    &KlingAI &Feb 2026   & 1,212\\
 \midrule
 \multicolumn{4}{l}{\textit{Open-source}} \\
 \midrule
 Wan 2.2 A14B   &Alibaba &Jul 2025   & 1,106\\
HunyuanVideo 1.5    &Tencent &Nov 2025  & 1,016\\
LTX 2.3    &Lightricks &Mar 2026  & 1,120\\
\bottomrule
\end{tabular}
\end{table}

We introduce \DatasetName, a video benchmark for modern public-speaking deepfake detection. Our benchmark is created through a multi-step pipeline, including filtering, semantic curation, and prompt-based generation. For each real video, we derive a semantic prompt from its frames and use it to generate matched synthetic clips, creating real/fake pairs with aligned visual and public-speaking context. 
\DatasetName controls the scenario category, real-video source, prompt source, and generation configuration while varying the generator.
The benchmark consists of 2,691 videos: 271 real and 2,420 generated clips across three high-risk public-speaking scenarios: direct-to-camera/casual videos, official footage, and studio interviews. The benchmark covers text-to-video and image-to-video generation for open-source generators where both modes are supported, and text-to-video generation only for the four commercial systems. 

The primary goal of this benchmark is to test visual deepfake detectors for binary classification (frame-based, video-based) in a cross-generator setup. Unlike legacy datasets that focus on face swapping or the manipulation of facial attributes, our benchmark contains fully generated videos produced by three modern open-source models: Wan 2.2~\cite{Wan}, HunyuanVideo 1.5~\cite{Hunyuanvideo}, LTX 2.3 distilled~\cite{hacohen2026ltx}, and four commercial text-to-video systems: Kling 3.0~\cite{kling_2026}, Veo 3.1~\cite{veo_3.1_2026}, Wan 2.6~\cite{wan_26_2026}, Grok Imagine 1.0~\cite{grok_imagine_2026}. Commercial generations are included as a fixed benchmark snapshot. \Cref{tab:video-generators-used} summarizes the used video generators; Elo scores were estimated by the Artificial Analysis as of Aug 2026 and are available in the Text-to-Video Leaderboard (No Audio)\footnote{\raggedright\url{https://artificialanalysis.ai/video/leaderboard/text-to-video?include-non-current=true&audio-output=false}}.

Our key contributions are:

\begin{enumerate}
\item A controlled single-person public-speaking benchmark for generated-video detection. \DatasetName contains 2,691 videos across three public-speaking scenarios and seven modern generators. Open-source generators are evaluated in text-to-video and image-to-video settings where supported, while commercial generators are evaluated only in a text-to-video setting.

\item A reproducible dataset protocol and metadata design. \DatasetName provides source identifiers, scenario labels, generator labels, generation-mode labels, released prompts, technical normalization details, and preprocessing details for dataset construction and detector evaluation.

\item We show that visual state-of-the-art deepfake detectors degrade substantially on modern generators in \DatasetName, with some scoring close to random chance.

\item We conduct a human study comparing the perceptual difficulty of \DatasetName relative to recent CelebDF++ and DeepSpeak v2. Accuracy on deepfakes from \DatasetName is 52.6\%, near chance, compared to 74.5\% and 69.8\% on the other two, while accuracy on real videos is similar.

\end{enumerate}

\section{Related work}

\begin{table*}[t]
\centering
\caption{Positioning of \DatasetName relative to existing deepfake and AI-generated video benchmarks.}
\label{tab:dataset_comparison}
\renewcommand{\arraystretch}{1.15}
\setlength{\tabcolsep}{12pt}
\small
\resizebox{0.92\textwidth}{!}{%
\begin{tabular}{@{}lrlcccc@{}}
\toprule
\textbf{Dataset}  &\textbf{Published}& \textbf{Scope} & \shortstack{\textbf{Gen.}\\\textbf{metadata}} & \shortstack{\textbf{Full-scene}\\\textbf{gen.}} & \shortstack{\textbf{Public}\\\textbf{speaking}} & \shortstack{\textbf{Cross-gen.}\\\textbf{protocol}} \\
\midrule
FaceForensics++~\cite{FF++}  & ICCV'19 & Face manipulation & \cmark & \xmark & \xmark & Limited \\
DFDC~\cite{dolhansky2020deepfakedetectionchallengedfdc}  & arXiv'20 & Face manipulation & Partial & \xmark & \xmark & \xmark \\
Celeb-DF-v2~\cite{li2020celebdflargescalechallengingdataset} & CVPR'20 & Face swap & \cmark & \xmark & Partial & Partial \\
Deepfake-Eval-2024~\cite{Deepfake-Eval-2024}  & arXiv'24 & Online deepfakes & \xmark & Mixed & \xmark & \xmark \\
Celeb-DF++~\cite{CDFv3}  & arXiv'25 & Face swap / talking face & \cmark & \xmark & Partial & Partial \\
TalkingHeadBench~\cite{Talkingheadbench}  & WACV'26 & Talking-head synthesis & \cmark & \xmark & Partial & Partial \\
ViF-Bench~\cite{Skyra}  & CVPR'26 & Open-domain generations & \cmark & \cmark & \xmark & Partial \\
DeepSpeak~\cite{DeepSpeak}  & CVPR'26 & Audiovisual deepfakes & Partial & Partial & Partial & \xmark \\
\textbf{\DatasetName}  & WACV'27 & Public-speaking generations & \cmark & \cmark & \cmark & \cmark \\
\bottomrule
\end{tabular}%
}
\end{table*}

Deepfake and AI-generated video benchmarks have evolved from controlled face manipulation datasets to broader multimodal and generative-video benchmarks. Legacy datasets such as FaceForensics++ \cite{FF++}, DFDC \cite{dolhansky2020deepfakedetectionchallengedfdc}, and CDFv2 \cite{li2020celebdflargescalechallengingdataset} have established standard evaluation settings for detecting face swaps, facial reenactment, and neural rendering artifacts. These datasets remain valuable for controlled training, historical comparison, and cross-dataset evaluation. Yet, their manipulations are largely localized to faces or facial motion, and their artifacts may not reflect the outputs of modern text-to-video and image-to-video systems that synthesize complete scenes.

Recent benchmarks address more realistic or multi-modal settings. Deepfake-Eval-2024 \cite{Deepfake-Eval-2024} evaluates images, audio, and videos collected from social media and detection platforms, providing a realistic online distribution of deepfakes. However, in-the-wild datasets often lack generator metadata, reproducible generation protocols, and controlled source splits, making it difficult to determine whether detector failures arise from generator shift, source mismatch, compression, scene content, or other confounding factors. DeepSpeak~\cite{DeepSpeak} and TalkingHeadBench~\cite{Talkingheadbench} focus on audiovisual deepfakes, lip synchronization, avatar synthesis, and speech-driven facial animation. These are important manipulation settings, but they differ from full-scene public-speaking videos generated by modern foundation video models. ViF-Bench \cite{Skyra} covers open-domain AI-generated videos from recent generators, but its broad scene distribution does not isolate public-speaking scenarios that are particularly relevant to misinformation.

\DatasetName fills this gap by providing a controlled evaluation benchmark for single-person public-speaking videos generated by modern text-to-video and image-to-video systems. Unlike in-the-wild datasets, it provides known generator labels, scenario labels, prompts, and generation-mode metadata. Unlike talking-head benchmarks, it targets full-scene generated videos rather than only speech-driven facial animation or localized synthesis. Unlike broad open-domain AI-video benchmarks, it focuses on a specific misinformation-relevant public-speaking setting. 

This design enables systematic cross-generator evaluation of whether detectors trained or selected elsewhere generalize to unseen modern video generators.

\section{\DatasetName Design}
The dataset is created with the following principles in mind.

\textbf{Disinformation relevance.} All videos depict a single speaking person in contexts commonly exploited for public-facing disinformation: direct-to-camera/casual addresses, official statements, and studio interviews.

 \textbf{Controlled quality.} All clips are normalized to a fixed duration of 5 seconds and a fixed resolution of $1280 \times 720$ using H.264. We apply filtering for text overlays, multiple visible people, missing faces, and severe technical artifacts. These controls reduce trivial cues and make the benchmark focus on visual evidence of synthetic generation rather than unrelated video artifacts.

\textbf{Visual-only.} Audio is excluded from the benchmark and should not be used to judge whether a video is real or fake. In the human study, videos were shown without audio.

\textbf{Evaluation-primary use.} \DatasetName is a held-out benchmark: the main protocol evaluates detectors trained on other data, without fine-tuning on \DatasetName. Training on \DatasetName is reported only as a diagnostic ablation.

\textbf{Reproducibility.} Each sample is a video with metadata describing its source, semantic scenario, generation mode, generator identity, prompt, and preprocessing steps. The processing pipeline code is available. 

\begin{table}[t]
\centering
\caption{
    \label{tab:source_type_and_semantic_performance}
    AUROC and EER for two temporal detectors, stratified by generator source type and semantic class.
    The source-type split shows that both detectors degrade substantially on closed-source generators relative to open-source generators.
    }
    \centering
    \setlength{\tabcolsep}{3.5pt}
    \resizebox{\linewidth}{!}{%
    \begin{tabular}{llrrrr}
    \toprule
    \multirow{2}{*}{\textbf{Group}} &
    \multirow{2}{*}{\textbf{Subset}} &
    \multicolumn{2}{c}{\textbf{DFD-FCG}~\cite{han2025towards}} &
    \multicolumn{2}{c}{\textbf{PwTF-DVD}~\cite{kim2025beyond}} \\
    \cmidrule(lr){3-4}
    \cmidrule(lr){5-6}
    & &
    \textbf{AUROC $\uparrow$} &
    \textbf{EER $\downarrow$} &
    \textbf{AUROC $\uparrow$} &
    \textbf{EER $\downarrow$} \\
    \midrule
    \multicolumn{6}{l}{\textbf{Generator source type}} \\
    \midrule
    Source &
    Open-source &
    57.4 & 45.2 &
    70.5 & 33.8 \\

    Source &
    Closed-source &
    34.5 & 61.7 &
    48.3 & 51.3 \\

    Source &
    \textbf{Closed $-$ Open} &
    \textbf{$-22.9$} & \textbf{$+16.5$} &
    \textbf{$-22.2$} & \textbf{$+17.5$} \\

    \midrule
    \multicolumn{6}{l}{\textbf{Semantic class}} \\
    \midrule
    Semantic &
    Direct-to-camera &
    50.6 & 50.1 &
    60.1 & 43.9 \\

    Semantic &
    Official statement &
    51.7 & 53.2 &
    72.3 & 34.9 \\

    Semantic &
    Studio interview &
    50.5 & 48.1 &
    66.1 & 39.1 \\

    \bottomrule
    \end{tabular}%
    }
\end{table}

\begin{table*}[t]
\centering
\caption{%
  Distribution of real videos in \DatasetName by scenario and source dataset.
}
\label{tab:scenario_source_breakdown}
\begin{tabular}{llcccc}
\toprule
\multirow{2}{*}{\textbf{Scenario}} &
\multirow{2}{*}{\textbf{Description}} &
\multicolumn{3}{c}{\textbf{Source dataset}} &
\multirow{2}{*}{\textbf{Total}} \\
\cmidrule(lr){3-5}
& & \textbf{MAVOS} & \textbf{OpenVid} & \textbf{TalkingCelebs} & \\
\midrule
Direct-to-Camera / Casual & Webcam, vlog, informal address &  3 & 97 &  0 & 100 \\
Official Statement        & Podium, press-room, briefing   &  4 &  5 & 62 &  71 \\
Studio Interview          & News, talk-show, podcast, studio &  5 & 95 &  0 & 100 \\
\midrule
\textbf{Total} & & \textbf{12} & \textbf{197} & \textbf{62} & \textbf{271} \\
\bottomrule
\end{tabular}
\end{table*}

\subsection{Dataset composition}
The dataset contains 271 real video clips organized into three semantic scenarios reflecting common public-speaking disinformation scenarios.
Each real sample consists of: 
(1) a 5-second MP4 video at $1280 \times 720$ resolution, 
(2) a text prompt describing the scene for synthetic-video generation,
(3) four representative keyframes extracted at evenly spaced intervals, and 
(4) metadata including source dataset, semantic scenario, preprocessing information.
For every real video in \DatasetName, we derive a single semantic prompt from the four uniformly sampled frames and use the same prompt for all generators included in the  evaluation benchmark. In all per-generator evaluations, each generator is compared against the same set of real videos.

\subsection{Raw sources and provenance}

The real videos are sourced from three complementary public video datasets. We use these datasets as candidate pools and then apply automatic filtering, semantic classification, and manual selection to obtain the final \DatasetName real-video set.

\textbf{OpenVid}~\cite{nan2025openvid1mlargescalehighqualitydataset} -- OpenVid-1M is a large-scale open-scenario video dataset containing more than one million text-video pairs. We use OpenVid-1M as a broad candidate pool for sourcing real videos in \DatasetName, primarily for the Direct-to-Camera/Casual and Studio Interview scenarios. Candidate clips were first automatically filtered for relevance to the target public-speaking settings and then manually reviewed. Manual selection verifies that each retained clip satisfies our single-person public-speaking definition, has sufficient visual quality, and matches one of the target semantic scenarios. OpenVid-1M is released under a CC-BY-4.0 license.

\textbf{TalkingCelebs}~\cite{TalkingCelebs} -- provides videos of politicians and public figures, making it useful for the Official Statement scenario. Because this source involves public figures and has non-commercial licensing constraints, we treat these samples carefully in the release policy.

\textbf{MAVOS-DD}~\cite{Croitoru-ArXiv-2025} -- provides diverse videos used to supplement all three semantic scenarios. MAVOS-DD is released under a CC BY-NC-SA 4.0 license. Samples derived from this source follow the corresponding non-commercial and share-alike constraints.

The distribution of these videos is shown below in \cref{tab:scenario_source_breakdown}.

\section{Dataset creation pipeline}

The dataset was produced through a multi-stage pipeline with strict objective filtering first, followed by semantic classification, manual curation, prompt generation, synthetic-video generation, and quality control.

\textbf{Stage A: Video filtering and normalization}

The filtering stage converts a large, noisy pool of raw videos into a high-quality candidate set by applying three phases of checks:

\textbf{Phase 1: Metadata filtering} We enforce resolution constraints (allowed: $1280 \times 720$ or $1920 \times 1080$) and minimum duration ($\geq 5.0$~seconds).

\textbf{Phase 2: Content checks} We apply two visual-content constraints:

\textit{Text overlay rejection:} Ten evenly-spaced frames are analyzed using OCR. A video is rejected if any frame contains text. This strict criterion prevents text overlays (logos, tickers, captions) from leaking shortcut cues for detectors and allows text overlays to be studied separately as robustness augmentations.
    
\textit{Single-person constraint:} Face detection is used to enforce the single-speaker setting. A video is retained only if no sampled frame contains more than one detected face and at least 80\% of the sampled frames contain exactly one detected face. This prevents identity ambiguity and reduces ambiguity in the public-speaking setup.

\begin{figure*}[t]
    \centering
    \begin{subfigure}[t]{0.33\textwidth}
        \centering
        \includegraphics[width=\linewidth]{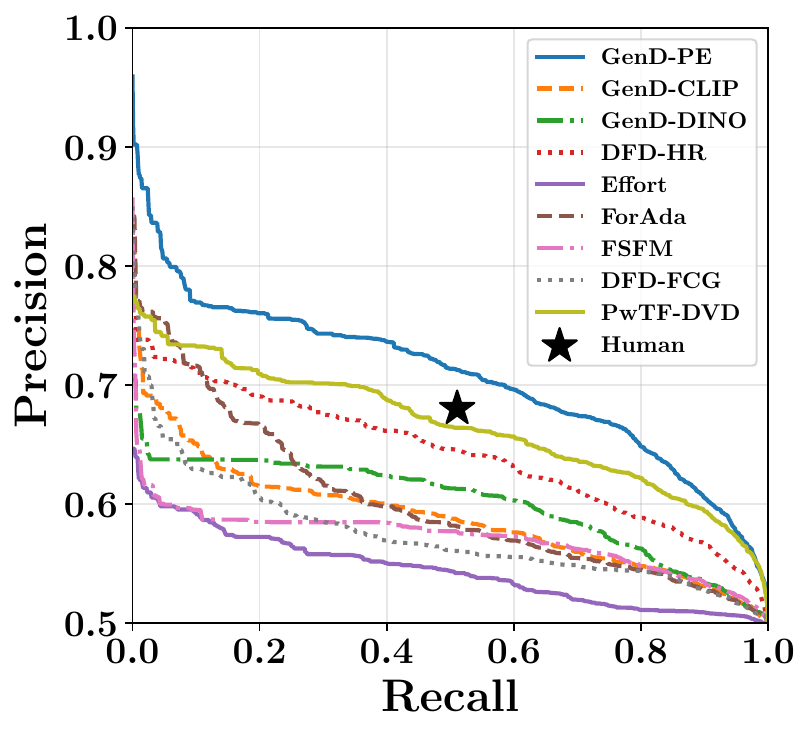}
        \caption{Full \DatasetName: OS and CS generators together}
    \end{subfigure}
    \hfill
    \begin{subfigure}[t]{0.33\textwidth}
        \centering
        \includegraphics[width=\linewidth]{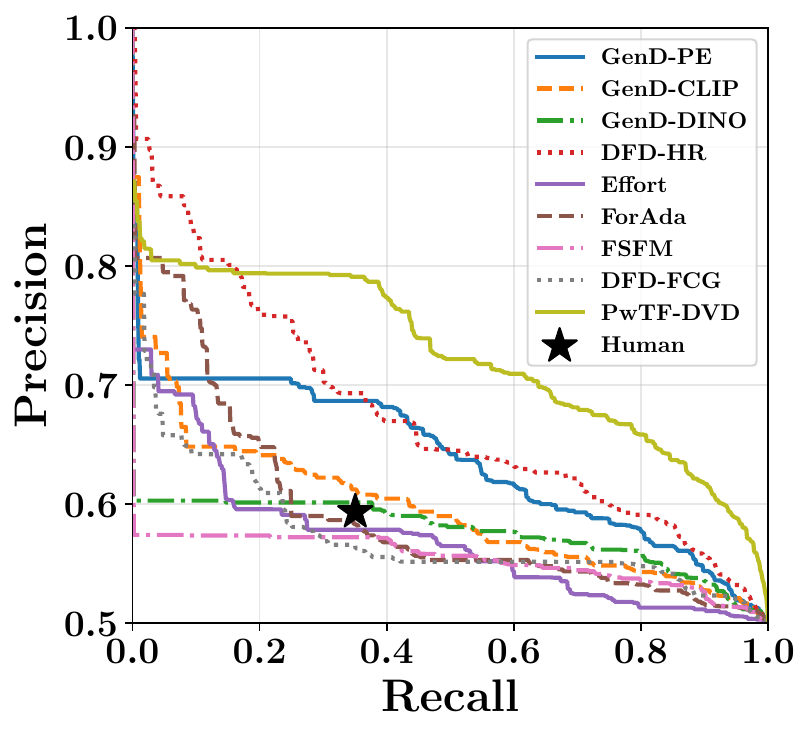}
        \caption{Open-source I2V generators from \DatasetName}
    \end{subfigure}
    \hfill
    \begin{subfigure}[t]{0.33\textwidth}
        \centering
        \includegraphics[width=\linewidth]{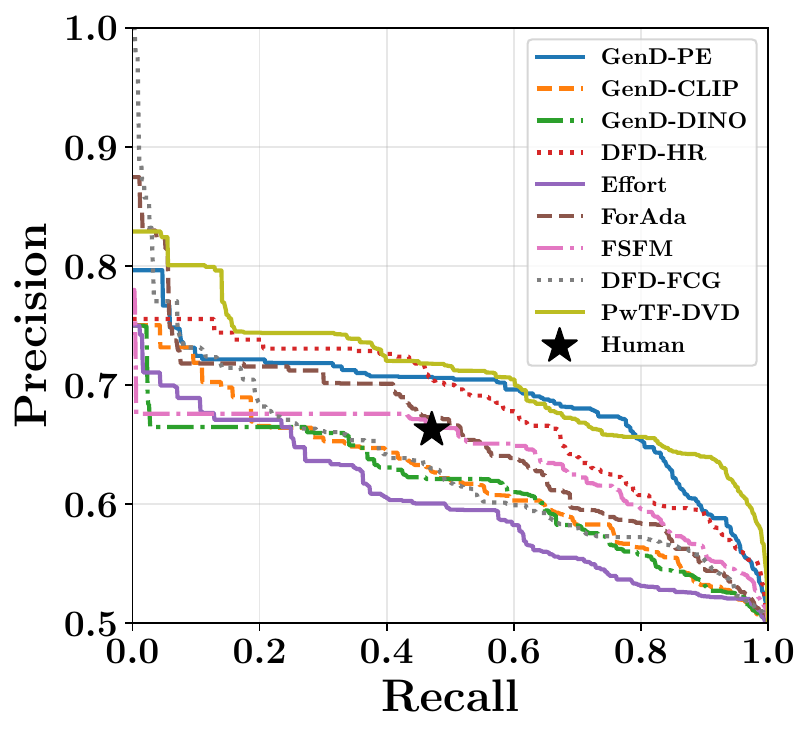}
        \caption{Open-source T2V generators from \DatasetName}
    \end{subfigure}
    
    \caption{
    \label{fig:precision-recall}
    Precision–recall curves on \DatasetName: (a) all generators, (b) open-source (OS) I2V, (c) OS T2V. Curves are macro-averaged over generators, each scored against same-scenario real videos (${\approx}$1:1, no resampling). Human precision assumes equal numbers of real and fake responses.
    }
    
\end{figure*}

\textbf{Phase 3: Normalization} Retained videos are trimmed to 5 seconds and resized to $1280 \times 720$ if they were of $1920 \times 1080$. 

\textbf{Stage B: Semantic classification}

Each filtered candidate video is classified using Gemini 2.5 VLM~\cite{Gemini-2.5}. We used a single VLM to maintain a consistent prompt style across all generators. This design allowed us to generate matched videos from the same prompt using multiple video generators, making the generator difference the target variable for the evaluation. The model returns a semantic scenario label from the following set: Official Statement, Studio Interview, Direct-to-Camera / Casual, or Other. The VLM label is used as a candidate label during filtering. Final labeling is confirmed during manual curation.

\textbf{Stage C: Manual curation protocol}.
We checked whether each candidate clip: 
(i) contains a single visible speaking person; 
(ii) matches one of the three target public-speaking scenarios; 
(iii) lacks large visible text overlays, watermarks, captions, or news tickers; 
(iv) satisfies the duration and resolution requirements; 
(v) has sufficient visual quality;
and (vi) does not contain severe occlusion, corruption, or obvious technical artifacts.

Within each scenario, the reviewer selected clips to encourage diversity in the source dataset, speaker appearance, background, camera framing, lighting, and recording style. Based on manual visual inspection, the real-video subset contains near-unique speaker identities, with the exception of the Official Statement category, where identities are heavily reused. This category contains 71 clips because fewer candidates satisfied all filtering and quality-control criteria.

\textbf{Stage D: Prompt generation}

Prompt-based matching is an intentional control mechanism in \DatasetName. For each real video, Gemini 2.5 VLM \cite{Gemini-2.5} produced a detailed prompt that is derived from the four uniformly sampled frames and is designed to describe the visual scene.  This reduces content mismatch between real and generated samples and makes the generator shift the primary variable of interest. 
The prompt template describes only observable scene-level and appearance-level properties, such as person appearance, background, lighting, framing, camera style, and speaking context. It does not include source-dataset identifiers or benchmark labels. All prompts are released to approved benchmark users together with the corresponding metadata. 

\begin{table*}[t]
\centering
\caption{%
  Cross-dataset video-level AUROC (\%) pooled across all samples and EER (\%) of state-of-the-art deepfake detectors. All detectors were trained on FF++~\cite{FF++} dataset. Most detectors perform well on CDFv3~\cite{CDFv3} but degrade on the harder \DatasetName; $\Delta$ shows the drop.
}
\label{tab:cross_benchmark_transfer_gap}
\setlength{\tabcolsep}{4.5pt}
\renewcommand{\arraystretch}{1.08}
\begin{tabular}{lrlcccccc}
\toprule
\multirow{2}{*}{\textbf{Detector}} &
\multirow{2}{*}{\textbf{Venue}} &
\multirow{2}{*}{\textbf{Type}} &
\multicolumn{2}{c}{\textbf{CDFv3}} &
\multicolumn{2}{c}{\textbf{\DatasetName}} &
\multicolumn{2}{c}{\textbf{DF26 $-$ CDFv3}} \\
\cmidrule(lr){4-5} \cmidrule(lr){6-7} \cmidrule(lr){8-9}
& & &
\textbf{AUROC $\uparrow$} & \textbf{EER $\downarrow$} &
\textbf{AUROC $\uparrow$} & \textbf{EER $\downarrow$} &
\textbf{$\Delta$AUROC} & \textbf{$\Delta$EER} \\
\midrule
DFD-FCG~\cite{han2025towards}  & CVPR'25  & Temporal & \textbf{94.3} & \textbf{12.3} & 49.8 & 49.9 & $-44.5$ & $+37.6$ \\
PwTF-DVD~\cite{kim2025beyond}  & ICCV'25  & Temporal & 92.3 & 15.7 & 63.1 & 41.0 & $-29.2$ & $+25.3$ \\
ForAda~\cite{ForAda}           & CVPR'25  & Frame & 75.6 & 32.1 & 56.6 & 45.8 & $-19.0$ & $\mathbf{+13.7}$ \\
Effort~\cite{Effort}           & ICML'25  & Frame & 78.7 & 29.8 & 44.0 & 55.7 & $-34.7$ & $+25.9$ \\
FSFM~\cite{FSFM}               & CVPR'25  & Frame & 79.6 & 27.5 & 52.6 & 48.1 & $-27.0$ & $+20.6$ \\
GenD-CLIP~\cite{GenD}          & WACV'26  & Frame & 85.2 & 23.0 & 54.2 & 47.7 & $-31.0$ & $+24.7$ \\
GenD-PE~\cite{GenD}            & WACV'26  & Frame & 89.9 & 16.9 & \textbf{69.7} & \textbf{35.9} & $-20.2$ & $+19.0$ \\
GenD-DINO~\cite{GenD}          & WACV'26  & Frame & 82.6 & 24.5 & 54.7 & 47.3 & $-27.9$ & $+22.8$ \\
DFD-HR~\cite{DFD-HR}           & CVPR'26  & Frame & 81.6 & 27.0 & 63.5 & 40.7 & $\mathbf{-18.1}$ & $\mathbf{+13.7}$ \\
\bottomrule
\end{tabular}
\end{table*}

\section{Deepfake generation}
\label{sec:generation}

We selected generators to cover three axes of modern video generation: 
(i) open-source models with reproducible checkpoints and documented inference settings; 
(ii) commercial systems accessible to end users through a common video generation interface; and 
(iii) both prompt-only and image-conditioned generation settings where controllable. 

Using the prompts generated in Stage~D, we produce generated counterparts of the real videos using seven recent video AI generation systems. The generator set includes both commercial and open-source systems and covers text-to-video (T2V) and image-to-video (I2V) generation modes (open-source models setup). 

For open-source models, single video generation uses a single NVIDIA H200 GPU with 141GB VRAM and takes approximately 33 minutes for Wan 2.2 A14B, 12 minutes for HunyuanVideo 1.5, and 3 minutes for the distilled LTX 2.3. Videos were generated in parallel on an internal research cluster using $3\times$ NVIDIA H200 GPUs. In total, generating 1,626 videos from open-source models required around 440 GPU-hours.

All commercial video models were accessed through paid access to the Higgsfield AI platform in March 2026. Higgsfield is a unified AI video generation platform that provides access to multiple video generators and allows users to switch between models ~\cite{higgsfield_ai_video}. For commercial models, we generated videos only for two scenario classes: Direct-to-Camera / Casual and Studio Interview. We excluded commercial generations for the Official Statement scenario due to platform restrictions to avoid violating platform terms.
Therefore, commercial-generator results are reported only for the Direct-to-Camera / Casual and Studio Interview subsets, with six rejected generation requests in total across providers.

Additionally, we ensured that the generated commercial videos contain no visible watermarks, model watermarks, logos, platform overlays, or other visible export markers. Open-source models were run using documented checkpoints and inference settings where available.

\section{Benchmark}

The benchmark measures how robust AI-video detectors are to unseen generators. We evaluate recent frame-based and temporal detectors from their public checkpoints, break down results by generator, generator source, generation mode, and scenario, and compare them with human perception. We report the area under the ROC curve (AUROC) as the primary metric and the equal error rate (EER) as a complementary one. All subsets use the same fixed split without changing hyperparameters or thresholds.

\subsection{Frame-based and temporal deepfake detectors}

For frame-based detectors, we preprocess images using the pipelines supplied by each model. Video scores are obtained by averaging the scores of 32 evenly sampled frames.

For temporal deepfake detectors, we evaluate DFD-FCG~\cite{han2025towards} and PwTF-DVD~\cite{kim2025beyond}. All models are evaluated without fine-tuning.
\Cref{tab:cross_benchmark_transfer_gap} shows that multiple state-of-the-art detectors achieve strong performance on the CelebDF++~\cite{CDFv3} (CDFv3) benchmark, with temporal methods reaching an AUROC of 94.3 and 92.3. However, their performance drops substantially on \DatasetName, to 49.8 and 63.1 AUROC, respectively. Most methods degrade substantially, remaining near chance; the highest AUROC of 69.7 is achieved by GenD-PE. This shows that \DatasetName is a more challenging benchmark for state-of-the-art detectors.

For further insight, we provide precision-recall curves in \cref{fig:precision-recall}. Human precision and recall, based on our study described in \cref{sec:human_study}, are marked for comparison with deepfake detectors. While \cref{fig:precision-recall} suggests that GenD-PE~\cite{GenD} performs the best across all generators, the setting-specific evaluation reveals performance differences, with detection performance varying between open-source image-to-video (I2V) and text-to-video (T2V) generators. The I2V setting is a more challenging scenario for both humans and deepfake detectors. Notably, the best performing detector depends on the generation setting: in the I2V setting, the temporal model PwTF-DVD~\cite{kim2025beyond} performs best, whereas in the T2V setting, PwTF-DVD and the frame-based GenD-PE~\cite{GenD} perform comparably.

\subsection{Per-generator analysis of detectors}\label{sec:per-generator-amalysis}

\begin{table*}[t]
\centering
\caption{\label{tab:frame-based-SOTA-models-AUROC}%
  Cross-generator video-level AUROC (\%) of state-of-the-art deepfake detectors trained on {FF++}. Mean is macro-averaged.
}
\begin{tabular}{l|cccc|cccccc|c}
\toprule
& \multicolumn{4}{c|}{\textbf{Commercial generators}} & \multicolumn{6}{c|}{\textbf{Open-source generators}} & \\
\cmidrule(lr){2-5}\cmidrule(lr){6-11}
\multirow{2}{*}{\textbf{Detector}} & \textbf{Grok} & \textbf{Kling} & \textbf{Veo} & \textbf{Wan} & \multicolumn{2}{c}{\textbf{HV}} & \multicolumn{2}{c}{\textbf{Wan}} & \multicolumn{2}{c|}{\textbf{LTX}} & \multirow{2}{*}{\textbf{Mean}} \\
& \textbf{1.0} & \textbf{3.0} & \textbf{3.1} & \textbf{2.6} & \multicolumn{2}{c}{\textbf{1.5}} & \multicolumn{2}{c}{\textbf{2.2}} & \multicolumn{2}{c|}{\textbf{2.3}} & \\
& \textbf{T2V} & \textbf{T2V} & \textbf{T2V} & \textbf{T2V} & \textbf{I2V} & \textbf{T2V} & \textbf{I2V} & \textbf{T2V} & \textbf{I2V} & \textbf{T2V} & \\
\midrule
DFD-FCG~\cite{han2025towards} & 20.5 & 27.4 & 29.6 & 60.3 & 41.1 & 76.5 & 61.3 & 58.4 & 61.1 & 45.9 & 48.2 \\
PwTF-DVD~\cite{kim2025beyond} & 34.5 & 60.4 & 26.7 & 71.6 & 41.9 & 56.7 & \textbf{87.7} & \textbf{92.9} & 79.5 & 64.0 & 61.6 \\
ForAda~\cite{ForAda} & 38.9 & 53.4 & 39.9 & 63.8 & 50.8 & 78.0 & 62.6 & 48.4 & 70.8 & 57.7 & 56.4 \\
Effort~\cite{Effort} & 13.6 & 32.3 & 21.3 & 29.0 & 44.3 & 68.8 & 61.7 & 30.4 & 61.6 & 45.5 & 40.9 \\
FSFM~\cite{FSFM} & 36.3 & 49.7 & 38.2 & 69.2 & 46.3 & 75.4 & 62.7 & 55.0 & 65.2 & 41.3 & 53.9 \\
GenD-CLIP~\cite{GenD} & 47.7 & 62.5 & 39.6 & 62.6 & 52.2 & 73.9 & 66.5 & 35.5 & 72.2 & 43.5 & 55.6 \\
GenD-PE~\cite{GenD} & \textbf{66.5} & 71.0 & \textbf{76.6} & \textbf{88.3} & \textbf{64.8} & 82.7 & 66.1 & 58.6 & 81.5 & \textbf{71.4} & \textbf{72.7} \\
GenD-DINO~\cite{GenD} & 35.4 & 70.5 & 42.4 & 76.6 & 55.6 & 73.0 & 63.0 & 34.7 & 72.0 & 46.5 & 57.0 \\ 
DFD-HR~\cite{DFD-HR} & 43.9 & \textbf{72.1} & 49.9 & 64.4 & 61.3 & \textbf{84.5} & 73.1 & 48.3 & \textbf{83.8} & 56.5 & 63.8 \\
\bottomrule
\end{tabular}
\end{table*}

Table~\ref{tab:frame-based-SOTA-models-AUROC} provides a per-generator difficulty of \DatasetName for the current state-of-the-art models. The proposed \DatasetName is not only difficult, with the best evaluated model achieving a mean AUROC of 72.7, but also diverse in the generative traces left to be detected. For instance, PwTF-DVD achieves an AUROC of 92.9 on Wan 2.2 T2V, but on HunyuanVideo (HV) 1.5 it performs no better than chance.

\subsection{Diagnostic ablation: training on \DatasetName}
\label{sec:retraining}
This experiment is a diagnostic ablation outside the main evaluation protocol.
To separate the effect of legacy training data from the intrinsic difficulty of DF26, we trained the best-scoring model from \cref{tab:frame-based-SOTA-models-AUROC}~-- GenD-PE~\cite{GenD}, initialized only with a pretrained backbone, on the Official Statement videos of one open-source I2V generator. We evaluate on the other two scenarios for all generators, covering both the training generator (in-distribution, gray cells in \cref{tab:retrained}) and unseen ones (cross-generator).

\Cref{tab:retrained} shows that retraining GenD-PE on the open-source HunyuanVideo 1.5 I2V improves cross-generator AUROC on commercial generators, detecting samples from Grok Imagine 1.0, Veo 3.1, and Wan 2.6, with an AUROC of at least 93.1. We therefore attribute much of the failure in \cref{tab:frame-based-SOTA-models-AUROC} to legacy training data. However, gains depend on the choice of training generator: Wan 2.2 I2V leaves the mean AUROC nearly unchanged, whereas LTX 2.3 I2V lowers it.

These results suggest that training only on legacy manipulation datasets, such as FF++, is no longer sufficient for the detection of modern generated videos. Next generation detectors should be trained on data collected from newer generators and evaluated in a cross-generator protocol.

\subsection{Open-source vs.\ commercial generators and scenario analysis}

\Cref{tab:source_type_and_semantic_performance} shows a consistent gap between open-source and commercial generators. Both temporal detectors perform substantially better on open-source generators than on closed-source generators. 

These results suggest that commercial generators represent a distinctly challenging evaluation setup. We emphasize that this analysis is diagnostic rather than causal: source type may correlate with other factors, including model family, post-processing, visual quality, and generation modality. The consistent drop across detectors highlights the importance of including commercial systems in evaluation-only benchmarks.

Additionally, we evaluate detector performance across semantic video scenarios in \cref{tab:source_type_and_semantic_performance}. 
These scenarios differ in framing, camera motion, background complexity, facial motion, and speaking style. For DFD-FCG~\cite{han2025towards}, AUROC remains close to chance across all three scenarios. PwTF-DVD~\cite{kim2025beyond} shows moderate variation, with the highest AUROC on official statements and lower performance on direct-to-camera videos. This suggests that detector failures on \DatasetName are driven more by generator shift than by the public-speaking semantic class.

\subsection{Human study}

\label{sec:human_study}
\begin{figure}[t]
\centering
\includegraphics[width=1.0\linewidth]{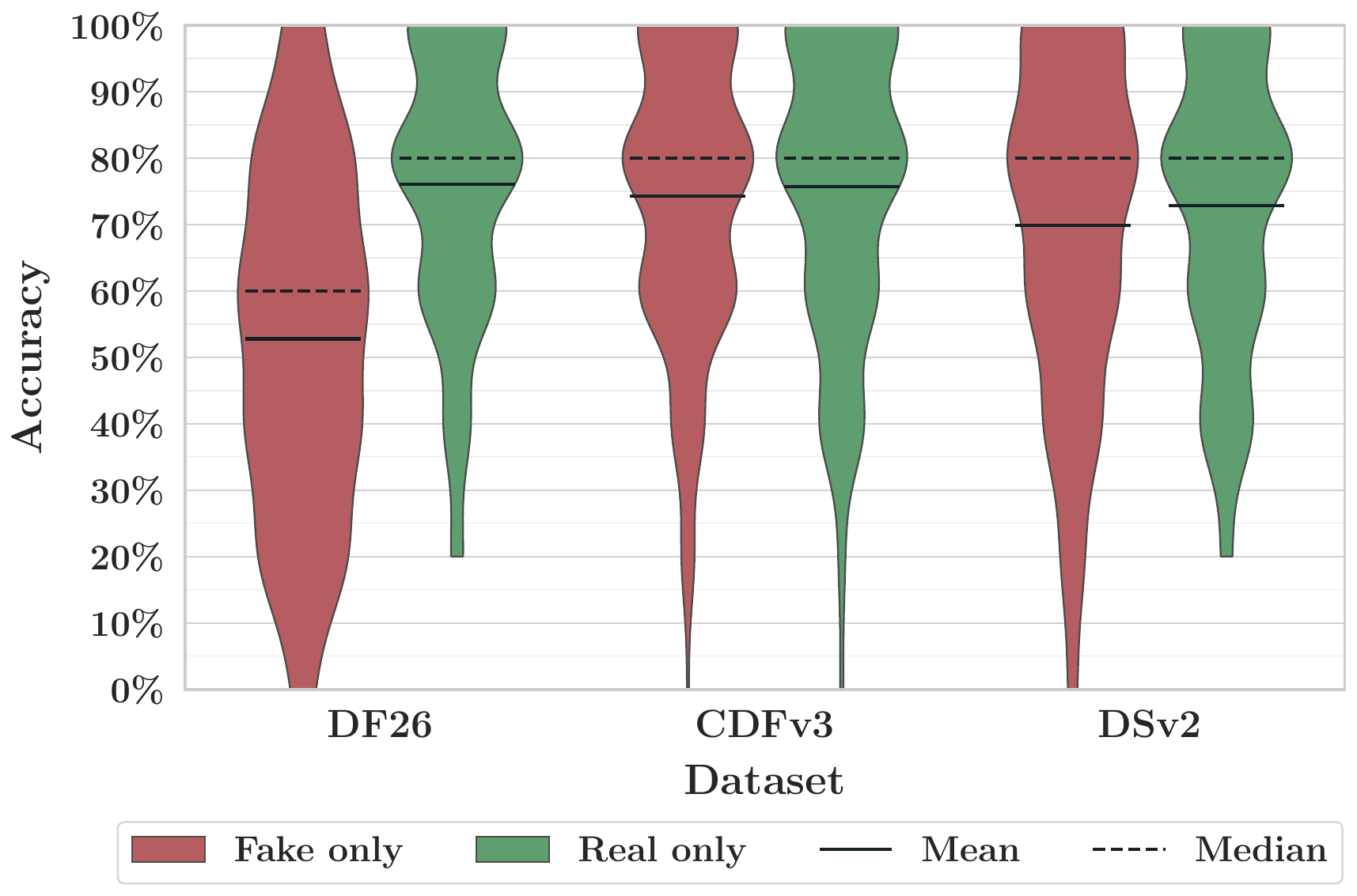}
\caption{Human accuracy distributions on real and fake videos over 232 labeling sessions on \DatasetName, CDFv3~\cite{CDFv3}, and DSv2~\cite{DeepSpeak}.}
\label{fig:study_dataset_accuracy_violin}
\end{figure}

\begin{table}[t]
\centering
\caption{Human accuracy by dataset, split by real and fake videos.}
\label{tab:human-study}
\small
\begin{tabular}{lcccc}
\toprule
\multirow{2}{*}{\textbf{Dataset}} 
& \multicolumn{2}{c}{\textbf{Accuracy (\%)}} 
& \multicolumn{2}{c}{\textbf{\# Videos}} \\
\cmidrule(lr){2-3} \cmidrule(lr){4-5}
& \textbf{Real} & \textbf{Fake} & \textbf{Real} & \textbf{Fake} \\
\midrule
\DatasetName & 76.0 & 52.6 & 200  & 1685 \\
CDFv3~\cite{CDFv3}        & 75.5 & 74.5 & 178  & 5240 \\
DSv2~\cite{DeepSpeak}         & 72.8 & 69.8 & 1863 & 1416 \\
\bottomrule
\end{tabular}
\end{table}
\begin{table*}[t]
\centering
\caption{\label{tab:retrained}%
  Diagnostic ablation. Video-level AUROC (\%) of GenD-PE trained on FF++ (original) or on the Official Statement videos of one \DatasetName generator (retrained), evaluated on the other two scenarios. Gray cells: in-distribution. Mean is macro-averaged over all columns.
}
\begin{tabular}{r|cccc|cccccc|c}
\toprule
& \multicolumn{4}{c|}{\textbf{Commercial generators}} & \multicolumn{6}{c|}{\textbf{Open-source generators}} & \\
\cmidrule(lr){2-5}\cmidrule(lr){6-11}
\textbf{Train set} & \textbf{Grok 1.0} & \textbf{Kling 3.0} & \textbf{Veo 3.1} & \textbf{Wan 2.6} &
\multicolumn{2}{c}{\textbf{HV 1.5}} & \multicolumn{2}{c}{\textbf{Wan 2.2}} & \multicolumn{2}{c|}{\textbf{LTX 2.3}} & \textbf{Mean} \\
& \textbf{T2V} & \textbf{T2V} & \textbf{T2V} & \textbf{T2V} &
\textbf{I2V} & \textbf{T2V} & \textbf{I2V} & \textbf{T2V} & \textbf{I2V} & \textbf{T2V} & \\
\midrule
FF++  & 66.5 & \textbf{71.0} & 76.6 & 88.3 & 68.5 & \textbf{88.7} & 69.1 & 65.2 & \textbf{86.5} & 79.8 & 76.0 \\
HV I2V  & \textbf{96.8} & 69.9 & \textbf{95.6} & \textbf{93.1} & \cellcolor{gray!25}\textbf{83.2} & \cellcolor{gray!25}85.3 & 65.8 & \textbf{84.1} & 67.0 & 82.4 & \textbf{82.3} \\
{Wan I2V} & 88.5 & 50.6 & 80.5 & 87.8 & 67.3 & 78.1 & \cellcolor{gray!25}\textbf{78.4} & \cellcolor{gray!25}83.6 & 70.6 & 80.3 & 76.6 \\
{LTX I2V} & 74.7 & 65.5 & 73.0 & 74.3 & 50.6 & 80.3 & 60.4 & 71.3 & \cellcolor{gray!25}78.5 & \cellcolor{gray!25}\textbf{94.0} & 72.3 \\
\bottomrule
\end{tabular}
\end{table*}

We evaluated the ability of people to recognize deepfakes in \DatasetName, as well as in two of the most closely related datasets: DeepSpeak v2 (DSv2)~\cite{DeepSpeak} and CelebDF++ (CDFv3)~\cite{CDFv3}. For each dataset, we used only the test split; pool sizes are reported in \cref{tab:human-study}. All videos were trimmed to 5 seconds and muted so that the amount and type of evidence available per decision are constant across datasets.

Participants were volunteer students, mostly from technical universities. Each participant labeled 30 videos as real or fake, where fake covers both AI-generated and AI-manipulated videos; 15 of the 30 were fake, and participants were not informed of the class distribution in advance. The 30 videos were stratified across the three datasets and the two classes, giving 5 videos per dataset per class, sampled uniformly at random without replacement within each stratum. Participants could rewatch each video at most 10 times. In total, we collected 232 labeling sessions. We did not record participant identities, so we estimate the number of unique participants from session timestamps, assuming that a participant who repeats the study starts the next session within 10 minutes of finishing the previous one, and that concurrent sessions come from different people. This gives an estimated 118 participants.

Results are shown in \cref{fig:study_dataset_accuracy_violin} and \cref{tab:human-study}. Accuracy on real videos is nearly identical across the three datasets (76.0\%, 75.5\%, 72.8\%), whereas accuracy on fake videos drops to 52.6\% on \DatasetName, barely above chance, compared with 74.5\% on CDFv3 and 69.8\% on DSv2. The asymmetry indicates that participants did not find the real subset of \DatasetName more confusing, but they failed to find enough evidence for the generated samples. Response times, per-generator difficulty, the distribution of the two response options, and screenshots of the labeling interface are provided in the supplementary material.

\section{Dataset Limitations}

Although \DatasetName provides a benchmark for modern synthetic-video detection, we acknowledge two limitations.

\textbf{Dataset scale.} \DatasetName contains 2,691 videos, which is small relative to large-scale benchmarks. We consider scaling the dataset through paid actors, similarly to DSv2~\cite{DeepSpeak}.

\textbf{Limited to the visual modality.}  \DatasetName is visual-only; audio is not part of the benchmark, so it does not evaluate audio realism, speech quality, lip-sync consistency, or audio-visual synchronization. Extending \DatasetName to multimodal detection is one of the directions for future work.

\section{Release policy}

\DatasetName is a controlled-access benchmark for evaluating deepfake and AI-generated video detectors. Access requires a request form stating affiliation, research purpose, intended use, and agreement to the terms of use. By default, \DatasetName is for held-out evaluation of detectors trained on other data. Training and fine-tuning are permitted only on the open-source subset, under the cross-generator protocol of \cref{sec:retraining}. Per Higgsfield's terms of use, the closed-source subset is evaluation-only and may not be used for training, fine-tuning, distillation, or any other model improvement.

\section{Conclusion}

We introduce \DatasetName~-- a video benchmark designed to evaluate vision-based deepfake detectors when challenged by modern video generation models. \DatasetName covers four closed-source text-to-video and three open-source image-to-video and text-to-video setups in public-speaking scenes.

Our evaluation shows that the performance of state-of-the-art detectors on \DatasetName is low, with several performing near random chance. Temporal detectors, while achieving high scores on CelebDF++, perform poorly on \DatasetName. Per-generator analysis shows that aggregate benchmark scores can hide severe failure modes, with both frame-based and temporal detectors performing well on certain generators while approaching or falling below chance on others.

In our human study, people detect \DatasetName fakes less reliably than those from CelebDF++ and DeepSpeak v2, with similar accuracy on real videos.

\DatasetName highlights a gap in current deepfake benchmarks by introducing modern generated videos that are difficult for both humans and machines to classify. This work motivates the development of better evaluation protocols and robust detection mechanisms to mitigate disinformation.

{\paragraph{Acknowledgments.} The research was supported by the EC Digital Europe Programme projects CEDMO 2.0 no.\ 101158609, CEDMO 3.0 no.\ 101330137, and by the CTU Student Grant SGS26/074/OHK3/1T/13.
}

{
    \small
    \bibliographystyle{ieeenat_fullname}
    \bibliography{main}
}

\clearpage
\setcounter{page}{1}
\twocolumn[
\begin{center}
    {\LARGE \textbf{Supplementary Material}}
    \vspace{1.5em}
\end{center}
]

\renewcommand{\thefigure}{S\arabic{figure}}
\renewcommand{\thetable}{S\arabic{table}}
\renewcommand{\theequation}{S\arabic{equation}}
\renewcommand{\thesection}{S\arabic{section}}

\setcounter{section}{0}
\setcounter{table}{0}
\setcounter{equation}{0}
\setcounter{figure}{0}

\begin{strip}
    \centering
    \includegraphics[width=0.85\textwidth]{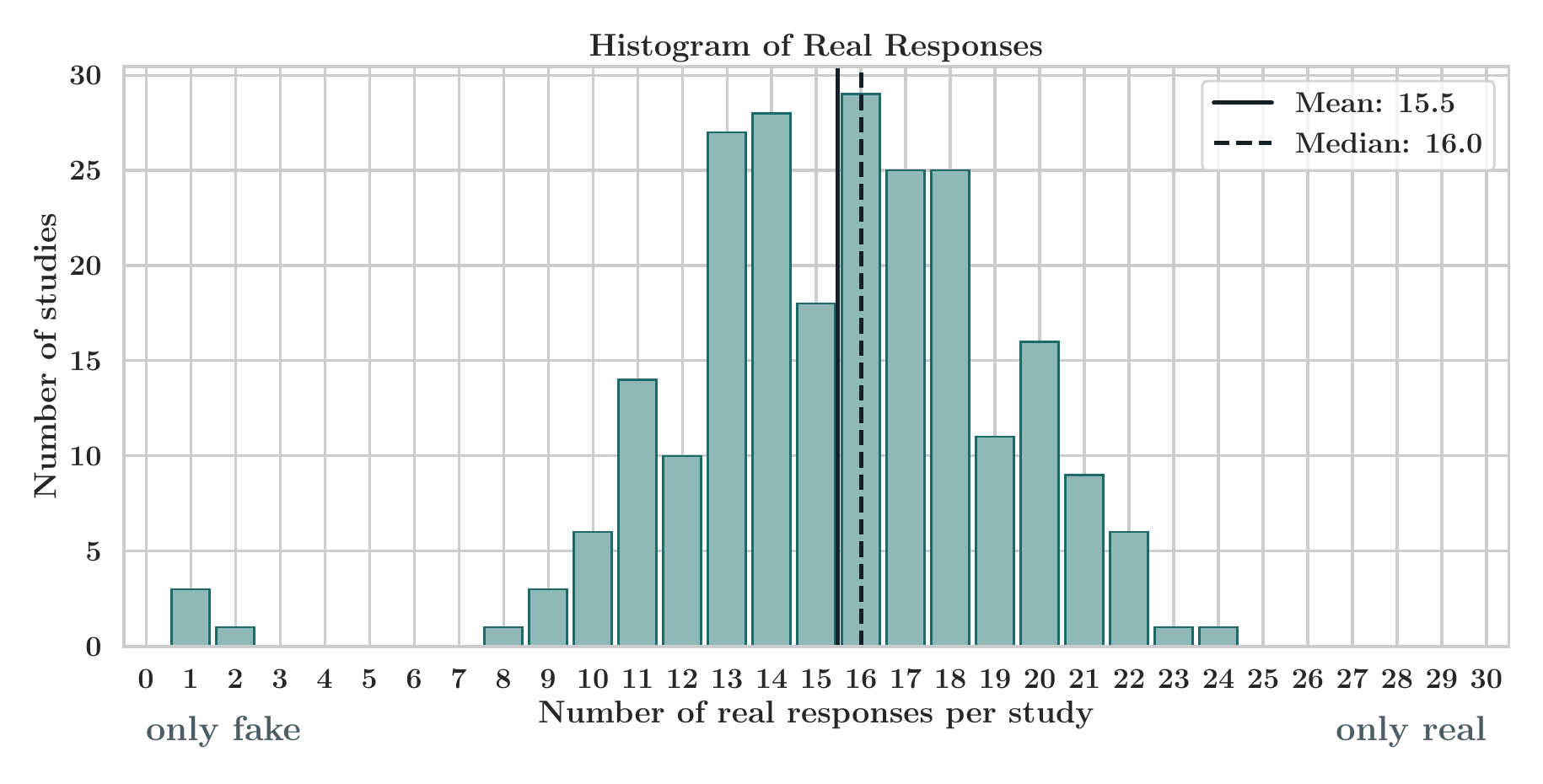}
    \captionof{figure}{\label{fig:real_response_histogram} 
    Human study -- histogram of real responses per trial. Leftmost bins correspond to trials where participants selected only fake videos; rightmost bins correspond to trials with only real selections. Participants were not informed about the true prior distribution, but largely converged around the true number of real videos per trial, which was 15.
    }
\end{strip}

\section{Additional results of the human study}

\begin{figure*}[t]
    \centering
    \includegraphics[width=0.65\linewidth]{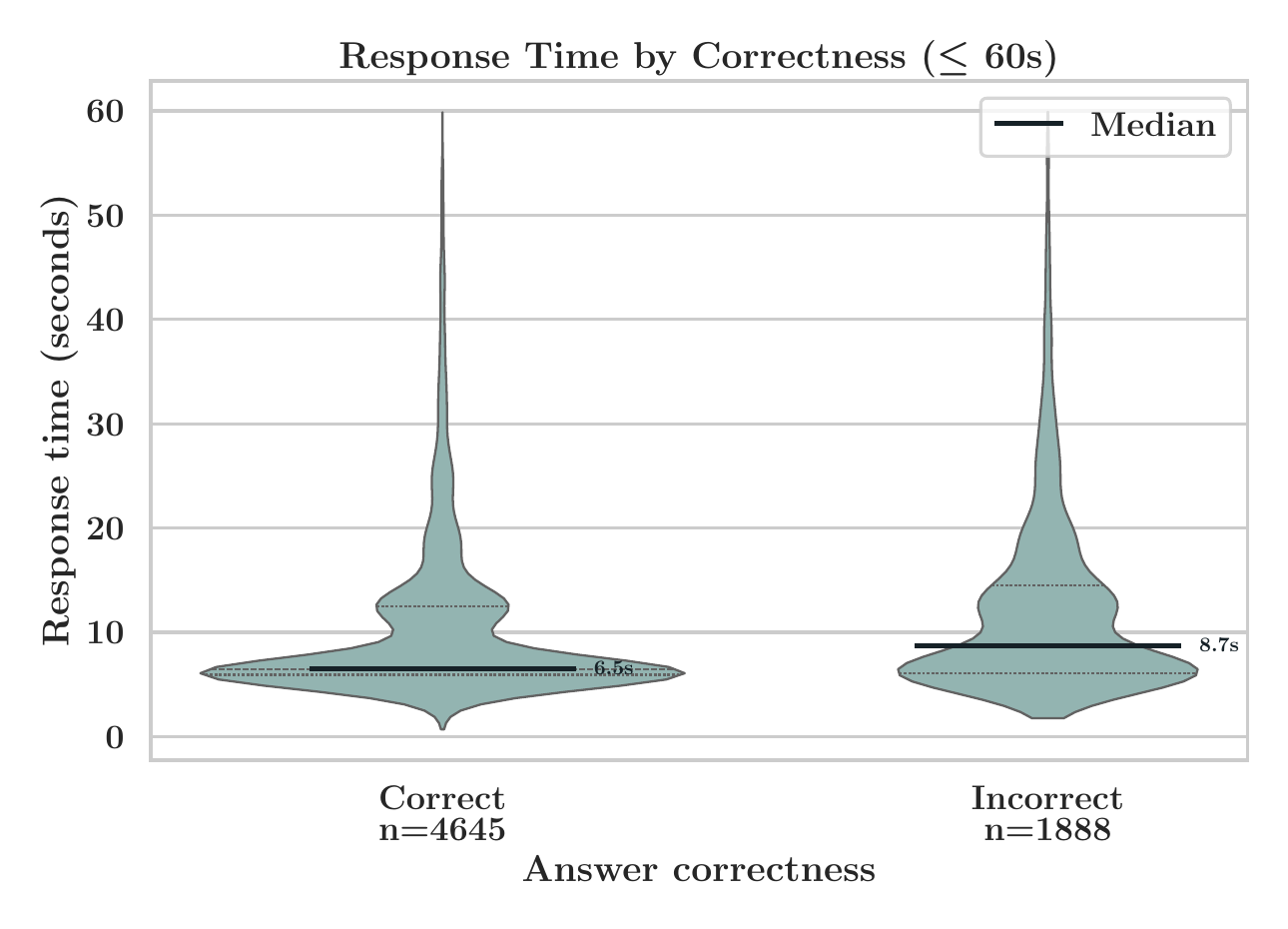}
    \captionof{figure}{\label{fig:hs-response_time_correct_vs_incorrect} 
    Distribution of human response times for correct (median is 6.5s) and incorrect (median is 8.7s) answers.
    }
\end{figure*}

\begin{figure*}[t]
    \centering
    \includegraphics[width=0.65\textwidth]{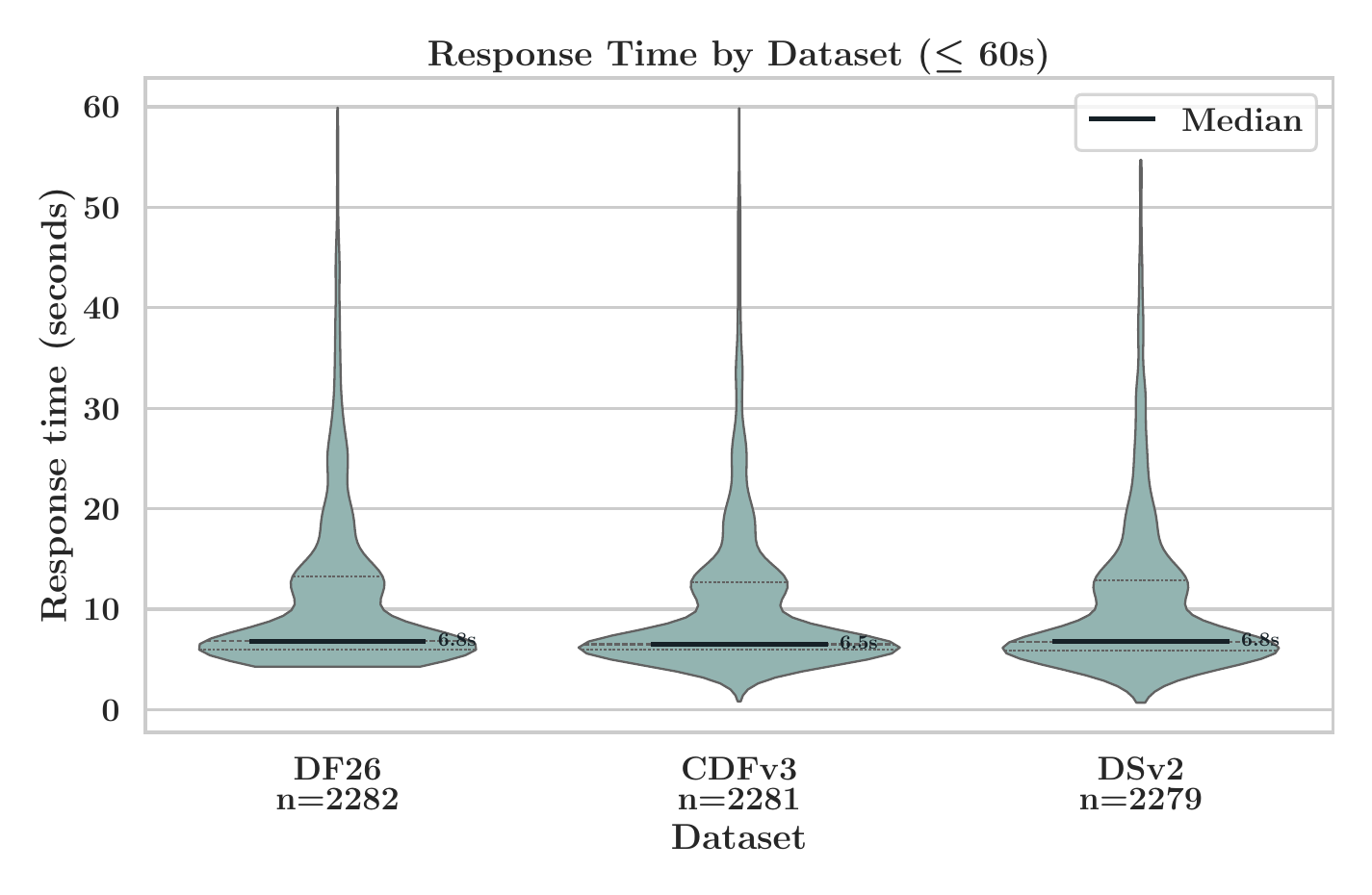}
    \caption{\label{fig:hs-response_time_per_datasets} 
    Distribution of human response times per dataset.
    }
\end{figure*}

The section shows additional results of the human study: the distribution of accuracies on fake and real subsets for all three tested datasets, DF26, CDFv3, and DSv2; see \cref{fig:study_dataset_accuracy_violin}. The histogram of the number of real responses is shown in \cref{fig:real_response_histogram}. Response times for correct and incorrect answers are presented in \cref{fig:hs-response_time_correct_vs_incorrect}, and per dataset in \cref{fig:hs-response_time_per_datasets}. Finally, detailed results per generator are shown in \cref{tab:hs-details}.

LTX 2.3 distilled I2V yields the lowest human accuracy at 25.2\%. For a generated clip carrying no perceptible evidence of synthesis, the expected accuracy is not 0\% but rather the rate at which participants label a genuine video as fake, which is 24.0\% on the DF26 real subset. Human accuracy on this generator is therefore at the level expected for clips that are perceptually indistinguishable from real footage, and the comparison is well matched, since the I2V clips are conditioned on the first frames of those same real videos. Notably, this generator is among the easier ones for automated detectors, reaching 79.5, 81.5, and 83.8 AUROC for PwTF-DVD, GenD-PE, and DFD-HR respectively (\cref{tab:frame-based-SOTA-models-AUROC}), so the traces that detectors exploit here are not the ones people can see.

\begin{table*}[t]
\centering
\small
\setlength{\tabcolsep}{6pt}
\renewcommand{\arraystretch}{1.08}

\caption{\label{tab:hs-details}
Detailed results of human answers by dataset and source. Real class has gray background.}

\begin{tabular}{
  ll
  S[table-format=4.0]
  S[table-format=3.0]
  S[table-format=2.1]
}
\toprule
\textbf{Dataset} & \textbf{Source}
& {\textbf{Answered}}
& {\textbf{Correct}}
& {\textbf{Accuracy (\%)}} \\
\midrule

\multirow{23}{*}{CDFv3}
& AniTalker        & 57   & 49  & 86.0 \\
& BlendFace        & 45   & 36  & 80.0 \\
& CDFv2            & 84   & 40  & 47.6 \\
& DaGAN            & 52   & 40  & 76.9 \\
& EDTalk           & 68   & 61  & 89.7 \\
& EchoMimic        & 69   & 54  & 78.3 \\
& FLOAT            & 75   & 66  & 88.0 \\
& FSRT             & 42   & 34  & 81.0 \\
& GHOST            & 39   & 30  & 76.9 \\
& HifiFace         & 48   & 20  & 41.7 \\
& HyperReenact     & 34   & 33  & 97.1 \\
& IP LAP           & 58   & 36  & 62.1 \\
& InSwapper        & 46   & 20  & 43.5 \\
& LIA              & 44   & 30  & 68.2 \\
& LivePortrait     & 52   & 49  & 94.2 \\
& MCNET            & 45   & 40  & 88.9 \\
& MobileFaceSwap   & 53   & 40  & 75.5 \\
& Real3DPortrait   & 66   & 60  & 90.9 \\
& SadTalker        & 61   & 49  & 80.3 \\
& SimSwap          & 38   & 19  & 50.0 \\
& TPSMM            & 35   & 26  & 74.3 \\
& UniFace          & 41   & 26  & 63.4 \\
\rowcolor{gray!25}
& Real    &  1152 &  870 &  75.5 \\
\midrule

\multirow{11}{*}{\DatasetName}
& Grok Imagine 1.0                    & 132 & 101 & 76.5 \\
& HunyuanVideo 1.5 I2V     & 113 & 52  & 46.0 \\
& HunyuanVideo 1.5 T2V     & 96  & 41  & 42.7 \\
& Kling 3.0               & 118 & 69  & 58.5 \\
& LTX 2.3 distilled I2V             & 107 & 27  & 25.2 \\
& LTX 2.3 distilled T2V             & 97  & 44  & 45.4 \\
& Veo 3.1                 & 118 & 56  & 47.5 \\
& Wan 2.2 A14B I2V             & 128 & 43  & 33.6 \\
& Wan 2.2 A14B T2V             & 117 & 61  & 52.1 \\
& Wan 2.6                 & 125 & 94  & 75.2 \\
\rowcolor{gray!25}
& Real           &  1150 &  874 &  76.0 \\
\midrule

\multirow{7}{*}{DSv2}
& Diff2Lip        & 138  & 114 & 82.6 \\
& FaceFusion      & 188  & 130 & 69.1 \\
& HelloMeme       & 197  & 159 & 80.7 \\
& LatentSync      & 223  & 101 & 45.3 \\
& LivePortrait    & 198  & 153 & 77.3 \\
& MEMO            & 207  & 146 & 70.5 \\
\rowcolor{gray!25}
& Real   &  1151 &  838 &  72.8 \\

\bottomrule
\end{tabular}
\end{table*}

\clearpage

\begin{strip}
    \centering
    \includegraphics[width=0.6\linewidth]{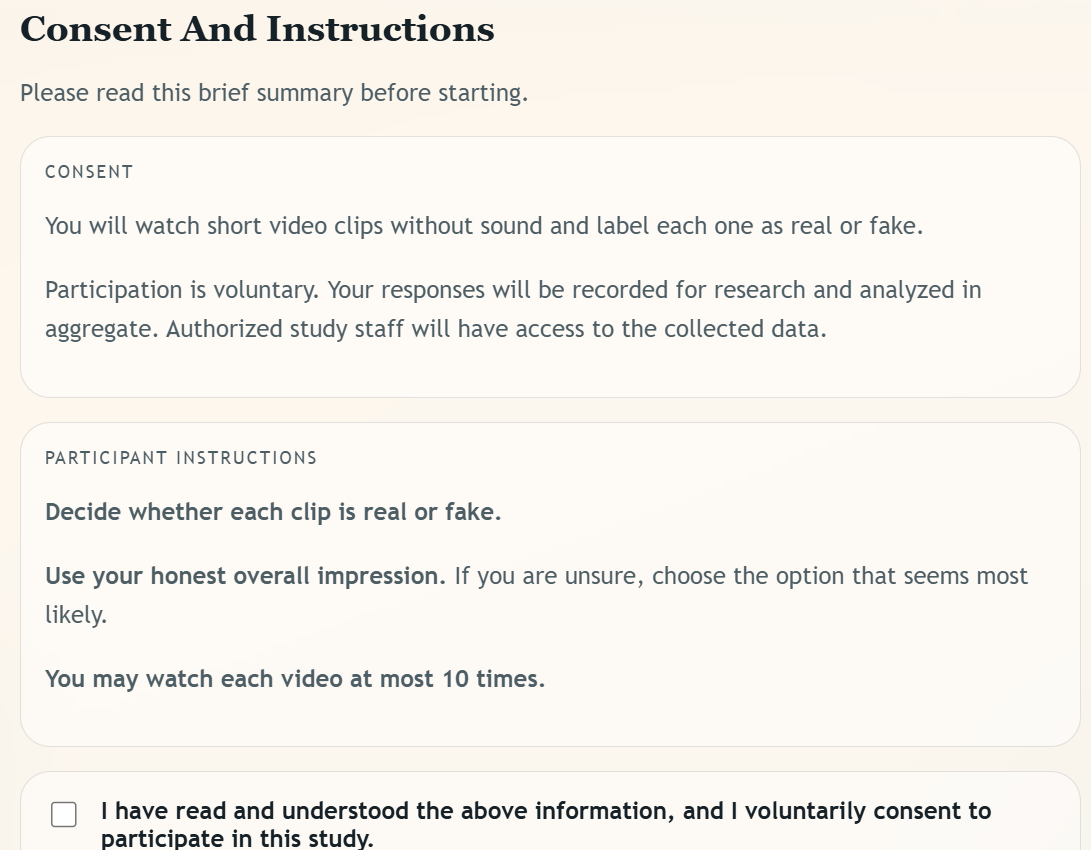}
    \captionof{figure}{Instructions shown to participants at the beginning of the study.}
    \label{fig:participant-instructions}
\end{strip}

\begin{strip}
    \centering
    \includegraphics[width=0.6\linewidth]{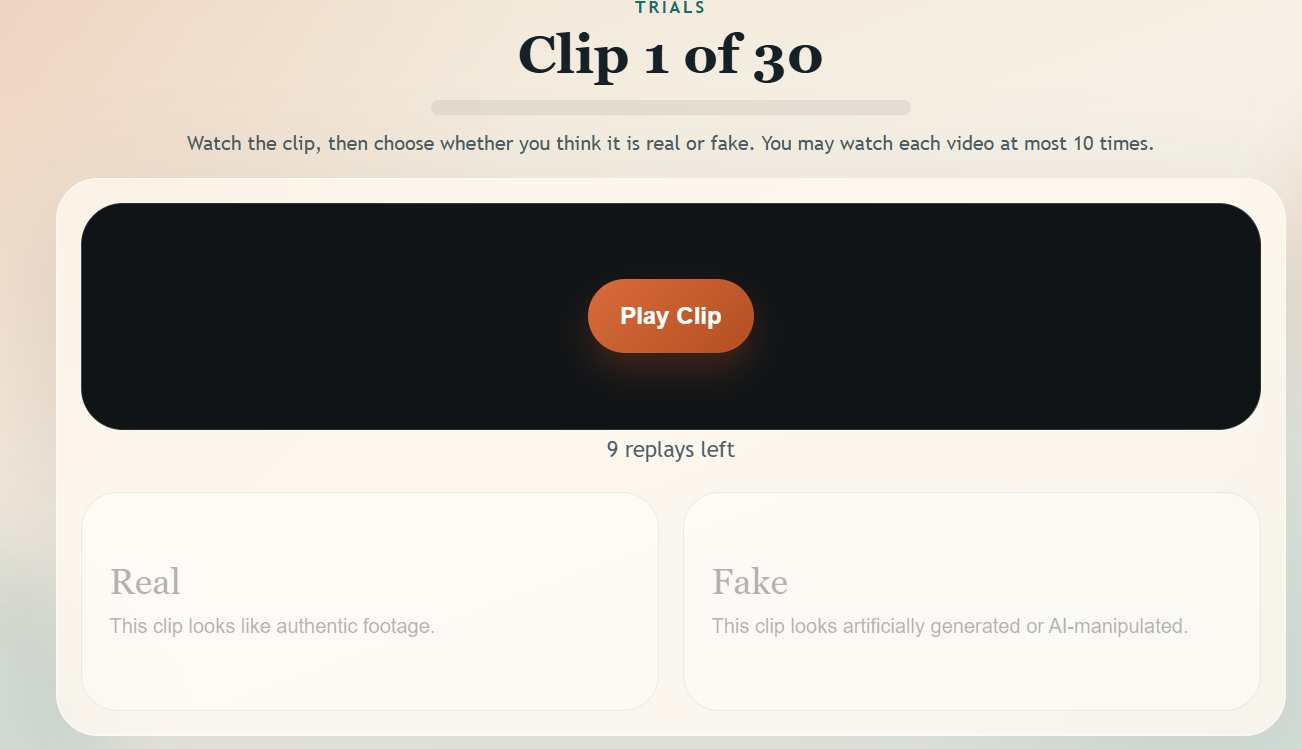}
    \captionof{figure}{Screenshot of the task interface shown to participants.}
    \label{fig:participant-task}
\end{strip}

\section{Human study: instructions and screenshots}

\Cref{fig:participant-instructions} shows the consent form and instructions presented before the first trial. Participants were told only that they would see short clips without sound and should give their honest overall impression, with no information about the class prior or the number of source datasets. \Cref{fig:participant-task} shows the trial interface. Each clip could be replayed a maximum of 10 times, with the remaining replay budget displayed, and the response was recorded along with the time from the first play to selection. The interface was identical for all three datasets, and clips were presented in randomized order.

\clearpage

\begin{strip}
\centering
\setlength{\tabcolsep}{2pt}

\begin{tabular}{c c c c c c}

\includegraphics[width=0.155\textwidth]{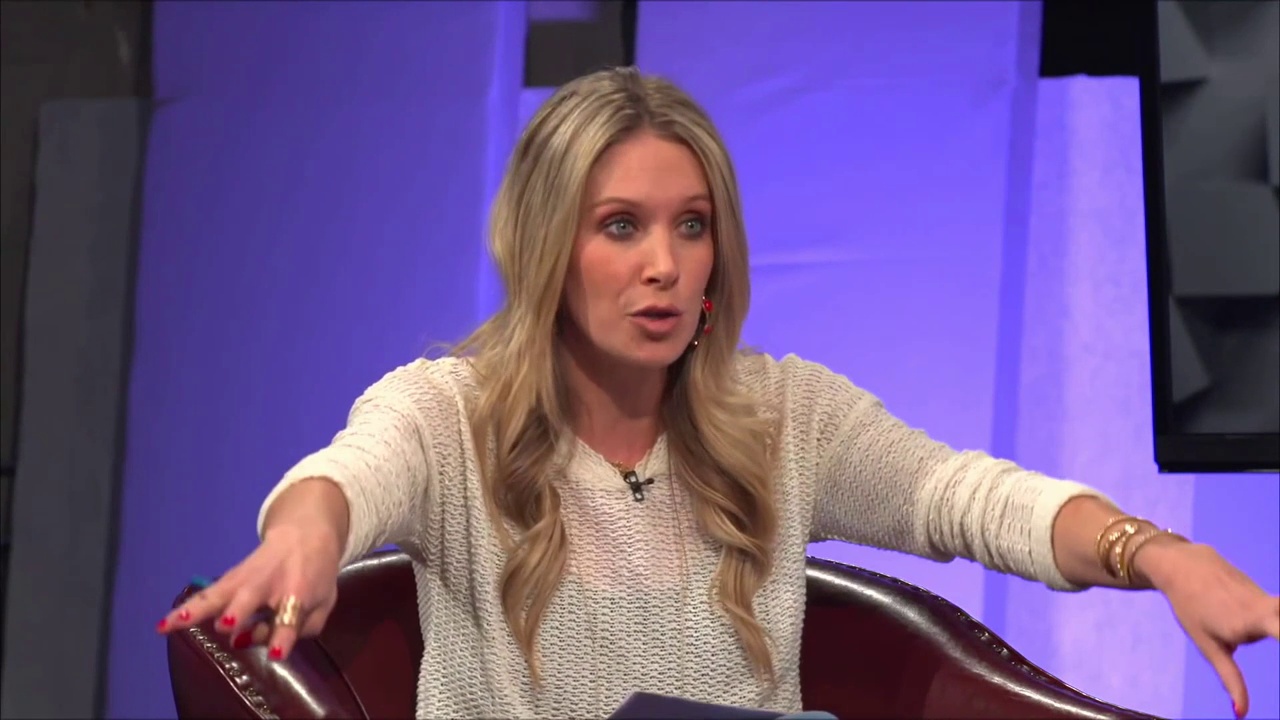}
&
\includegraphics[width=0.155\textwidth]{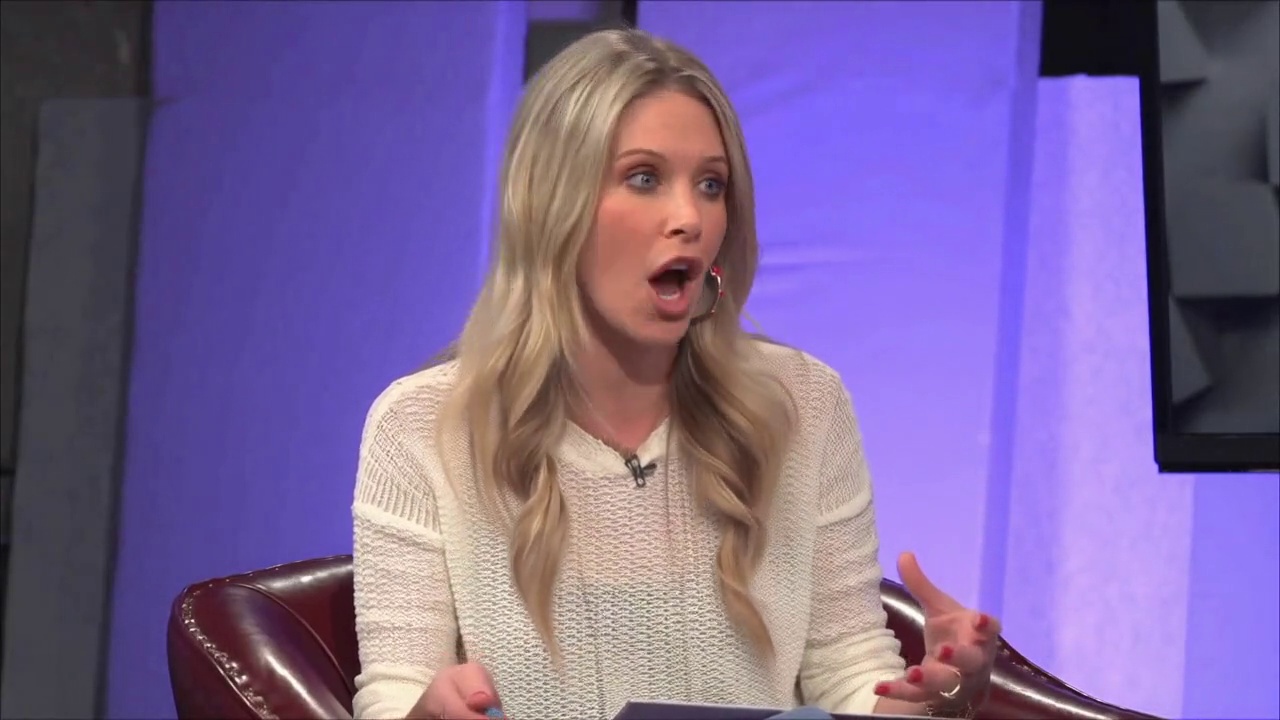}
&
\includegraphics[width=0.155\textwidth]{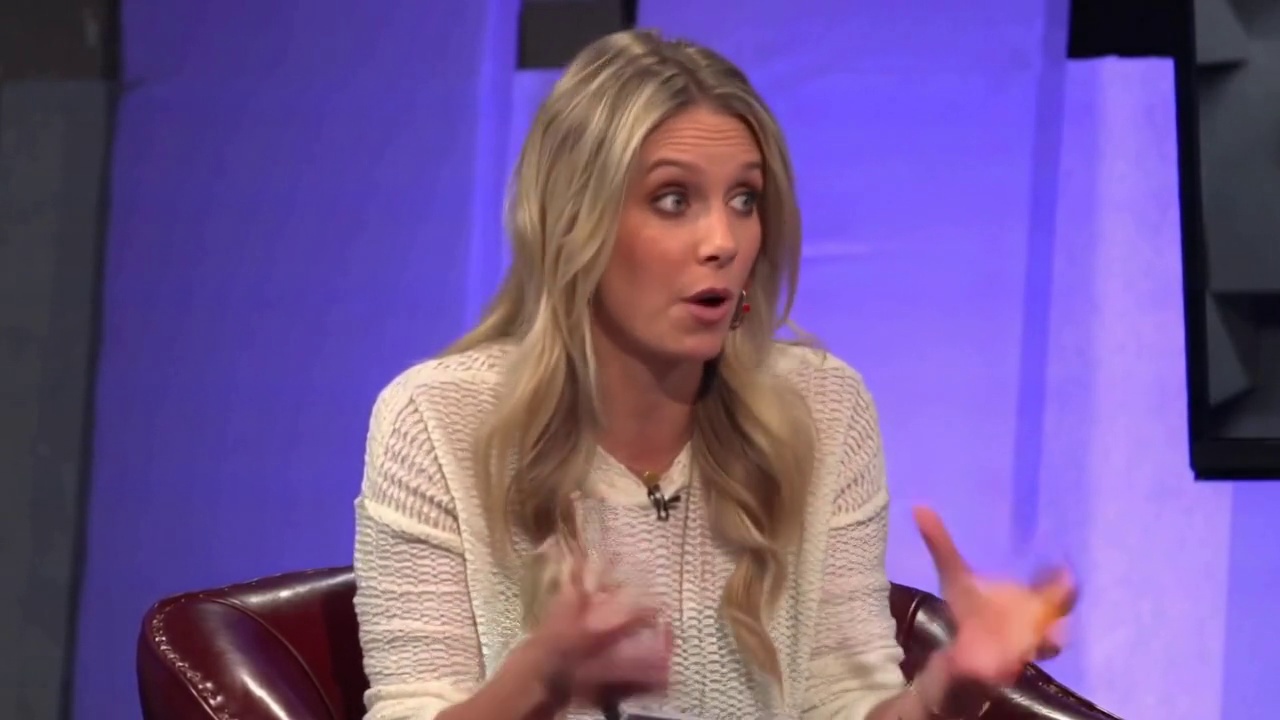}
&
\includegraphics[width=0.155\textwidth]{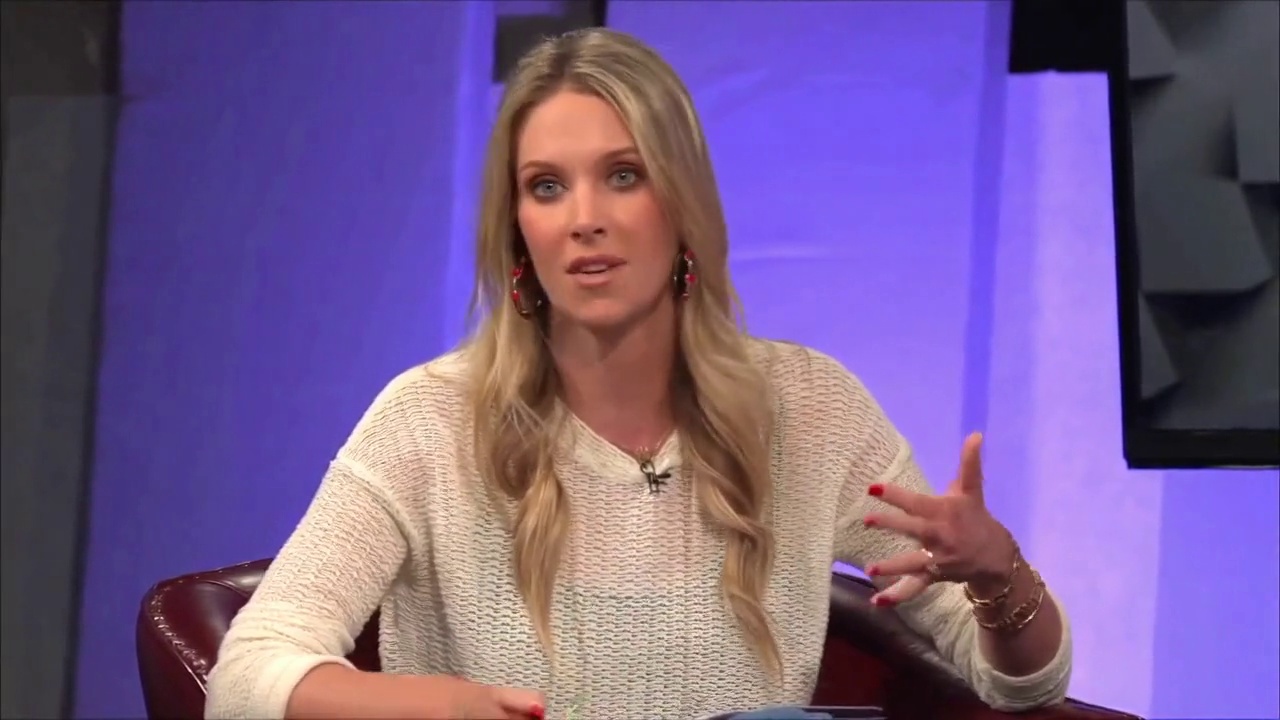}
&
\includegraphics[width=0.155\textwidth]{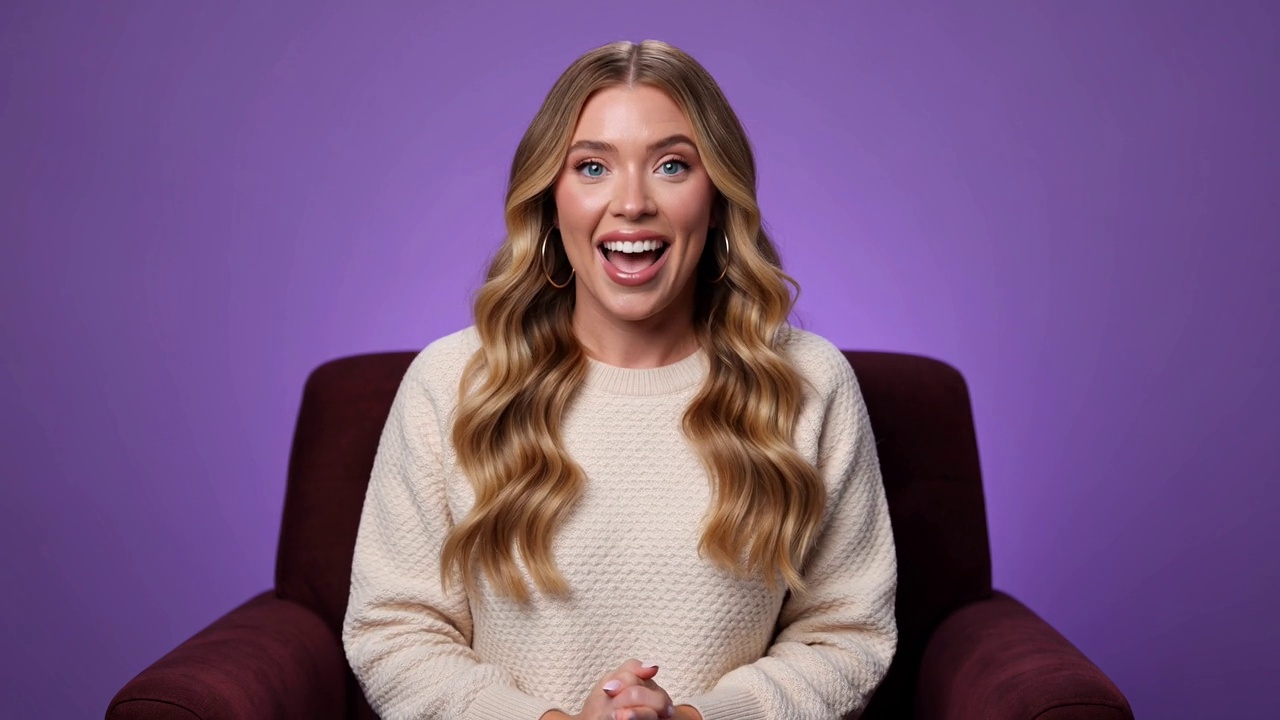}
&
\includegraphics[width=0.155\textwidth]{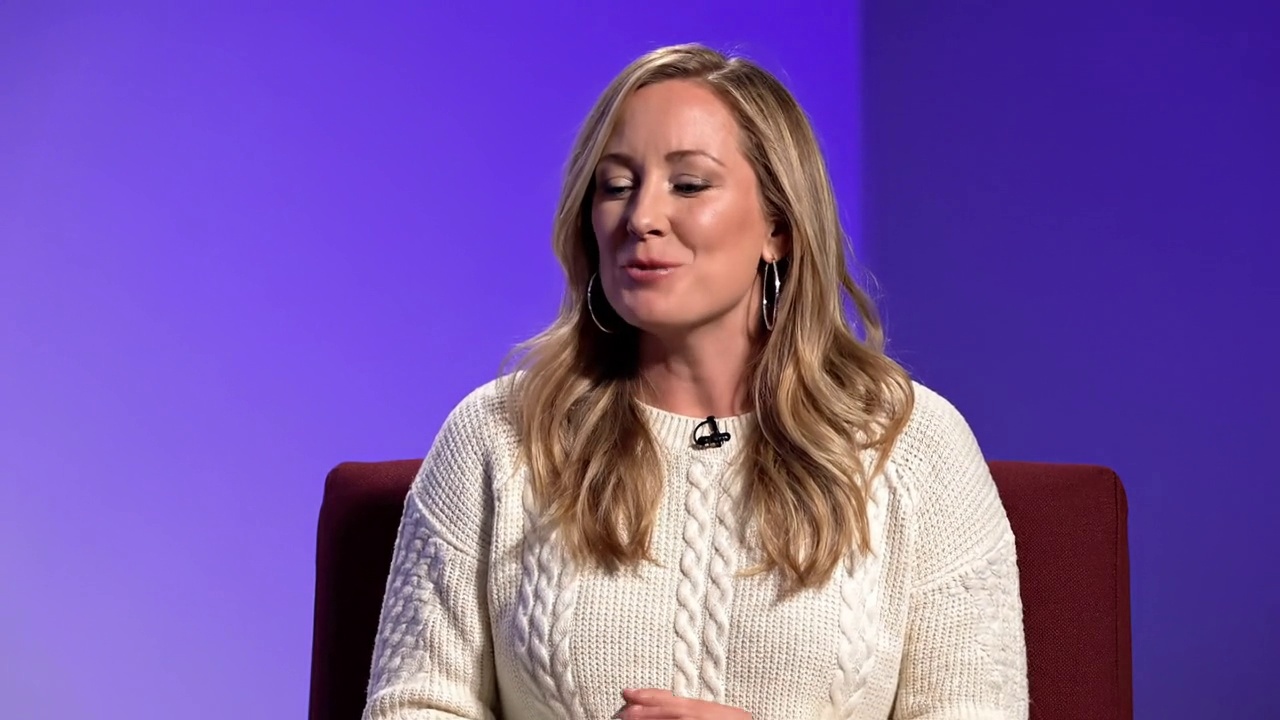}
\\[-1pt]
Real
& HunyuanVideo I2V
& LTX 2.3 I2V
& Wan 2.2 A14B I2V
& Grok Imagine 1.0
& Veo 3.1
\\[4pt]

&
\includegraphics[width=0.155\textwidth]{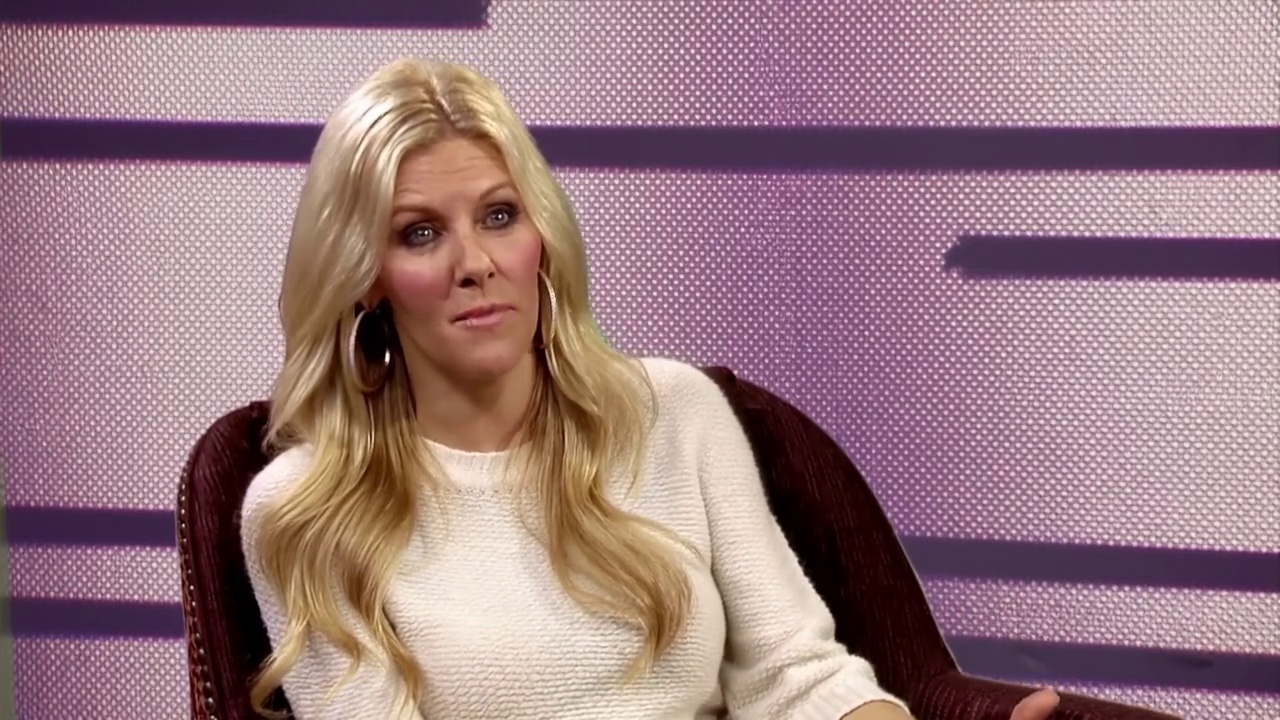}
&
\includegraphics[width=0.155\textwidth]{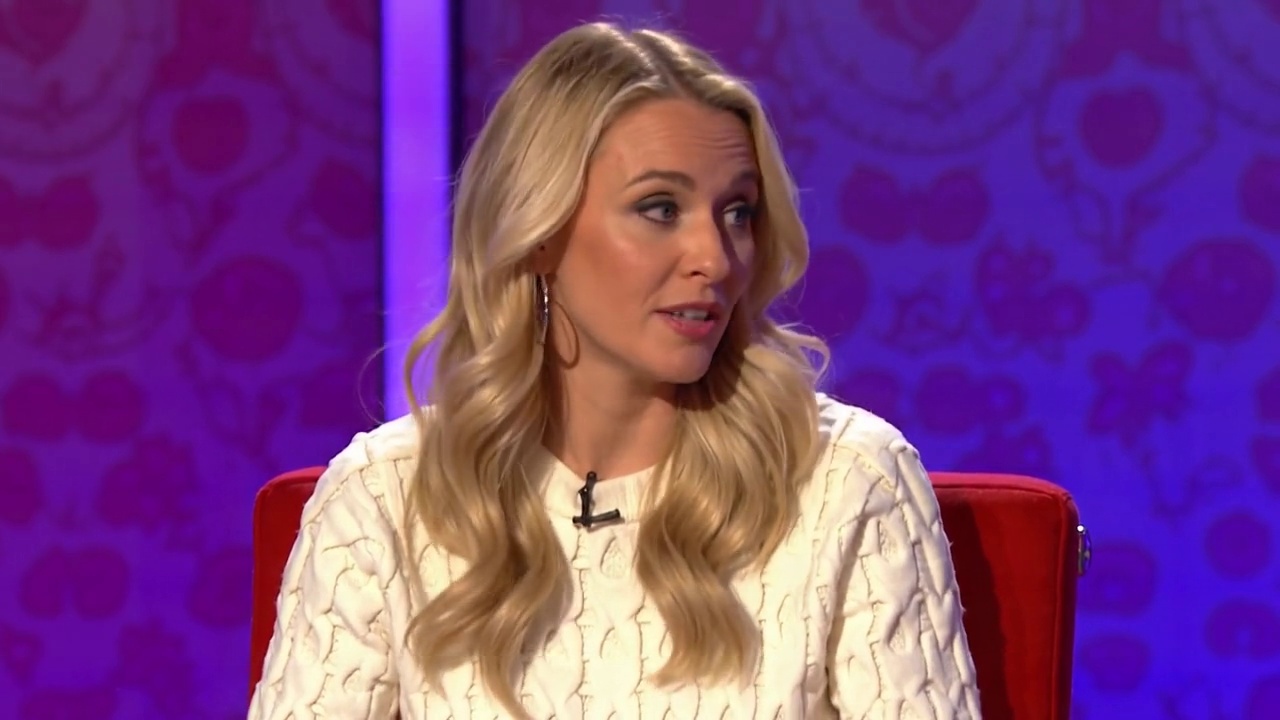}
&
\includegraphics[width=0.155\textwidth]{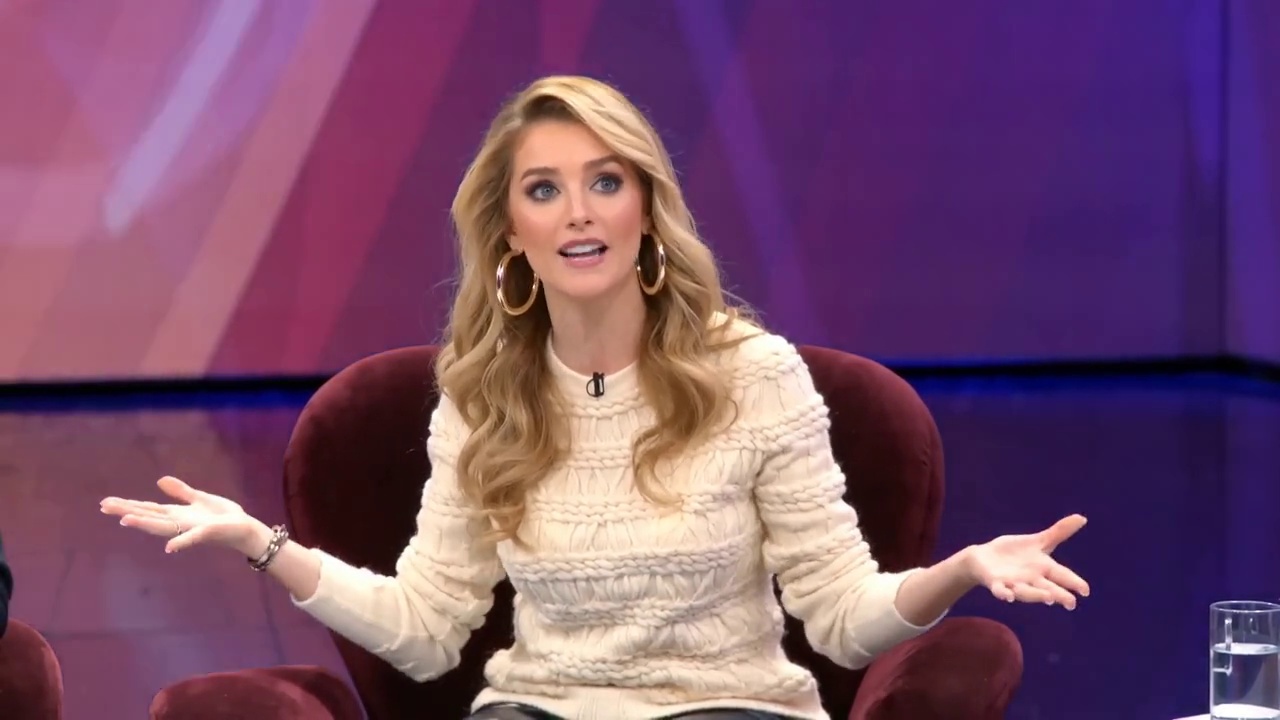}
&
\includegraphics[width=0.155\textwidth]{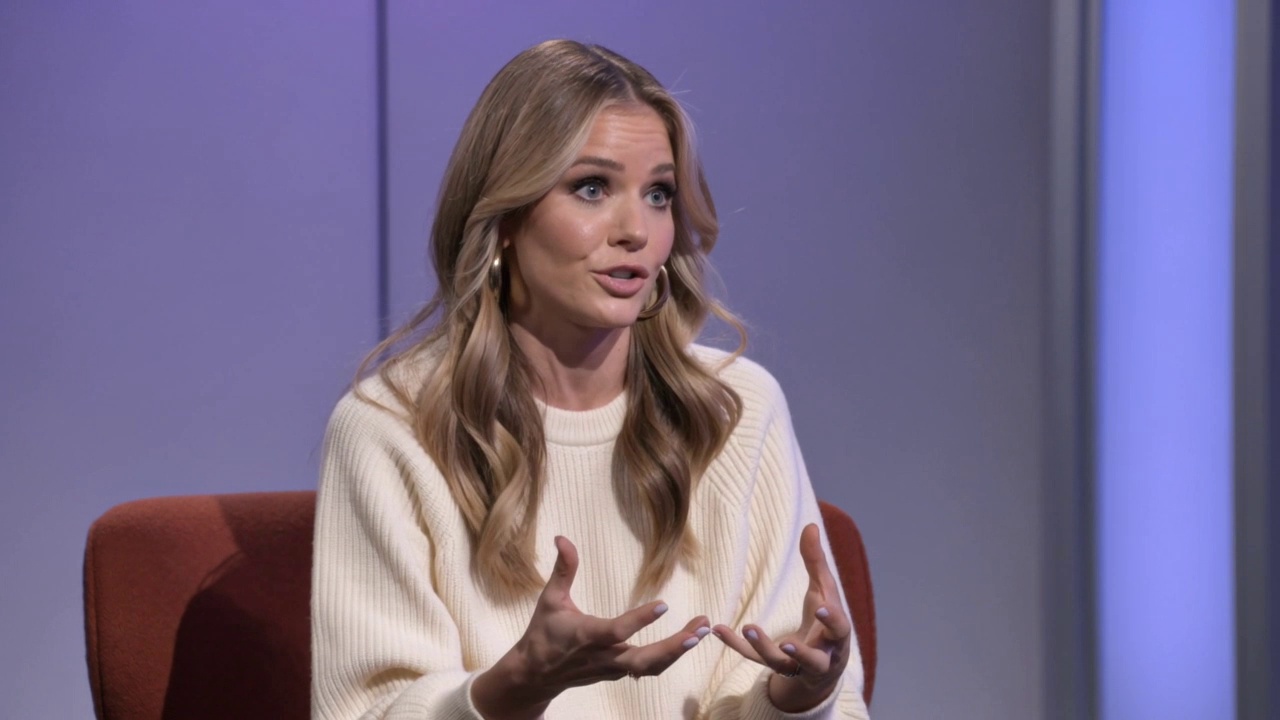}
&
\includegraphics[width=0.155\textwidth]{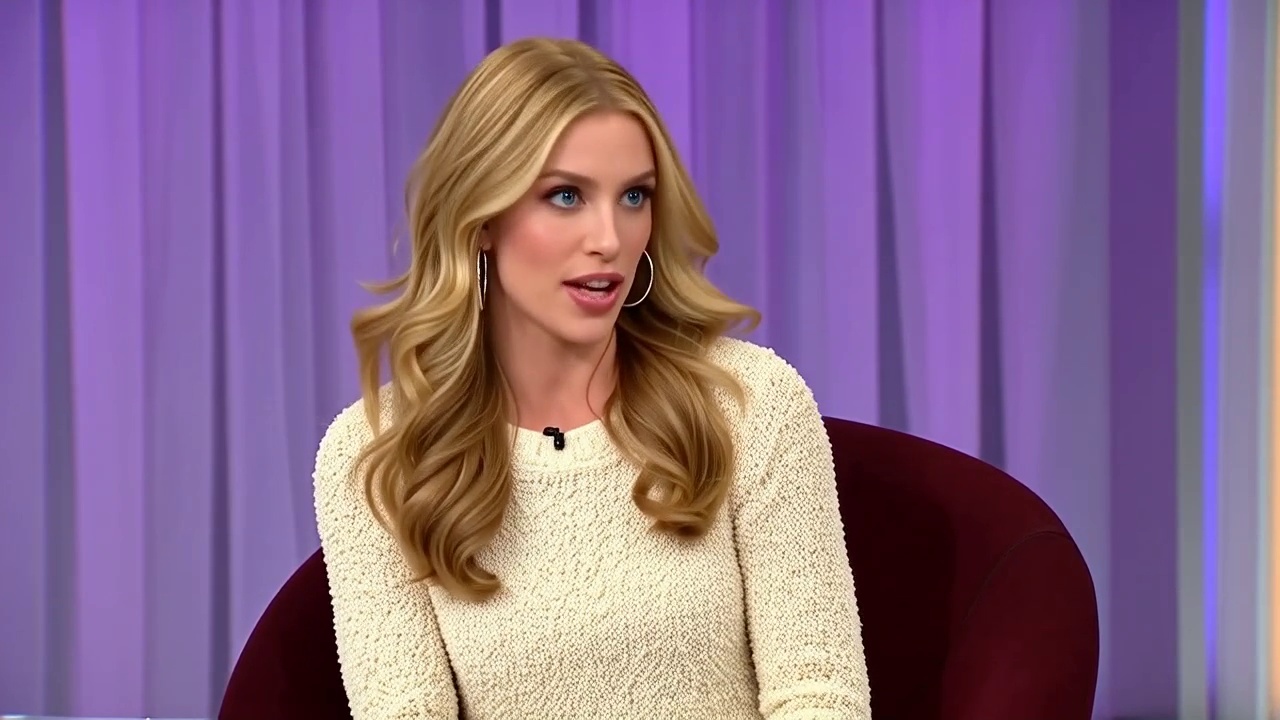}
\\[-1pt]
& HunyuanVideo T2V
& LTX 2.3 T2V
& Wan 2.2 A14B T2V
& Kling 3.0
& Wan 2.6
\\

\\
\includegraphics[width=0.155\textwidth]{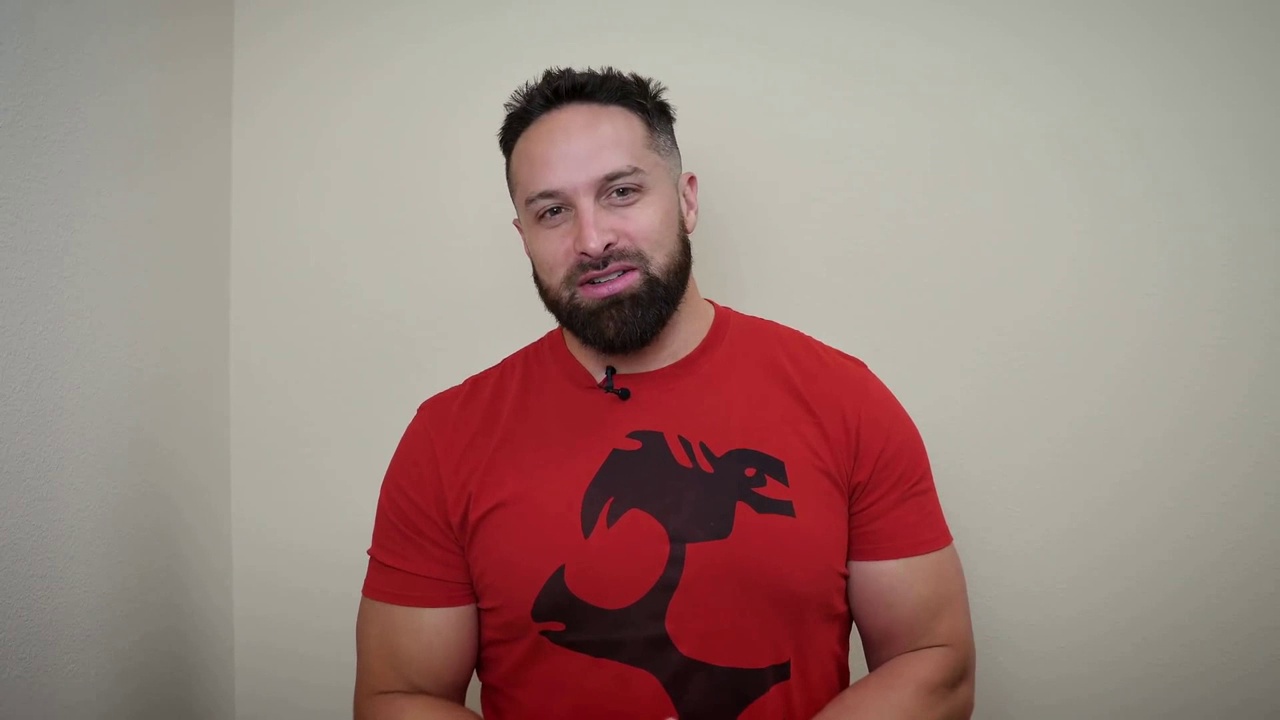}
&
\includegraphics[width=0.155\textwidth]{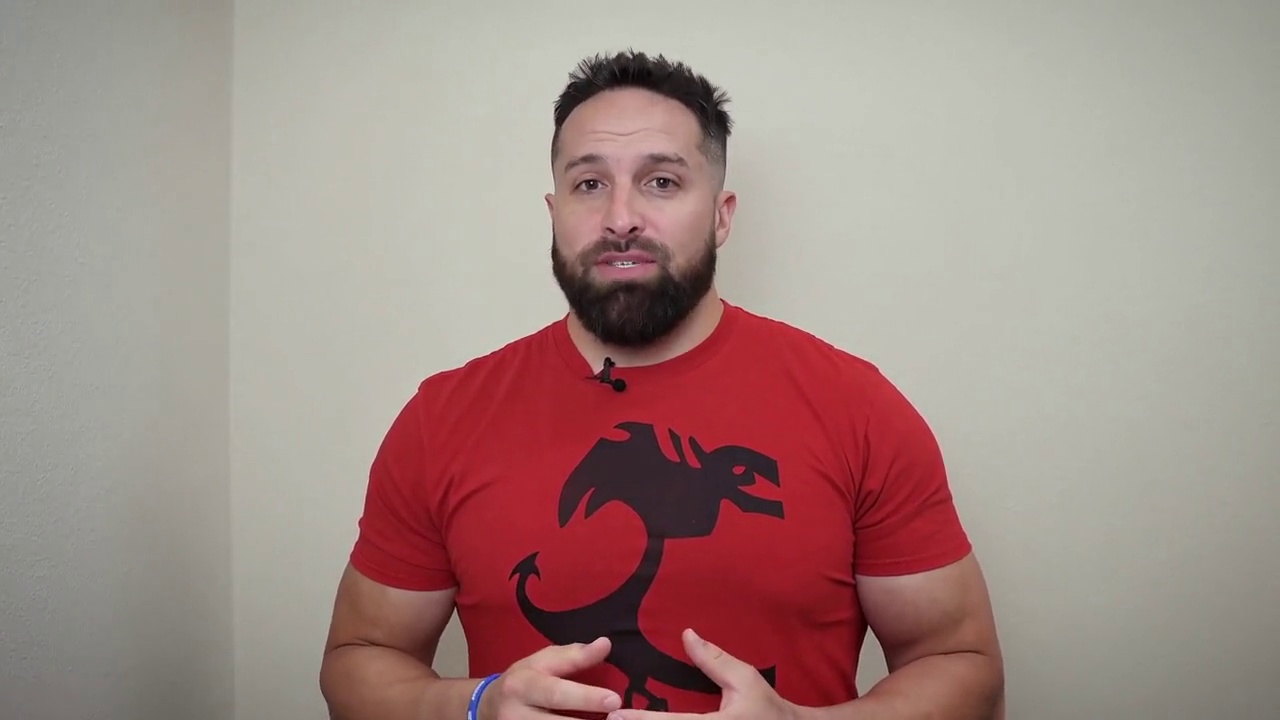}
&
\includegraphics[width=0.155\textwidth]{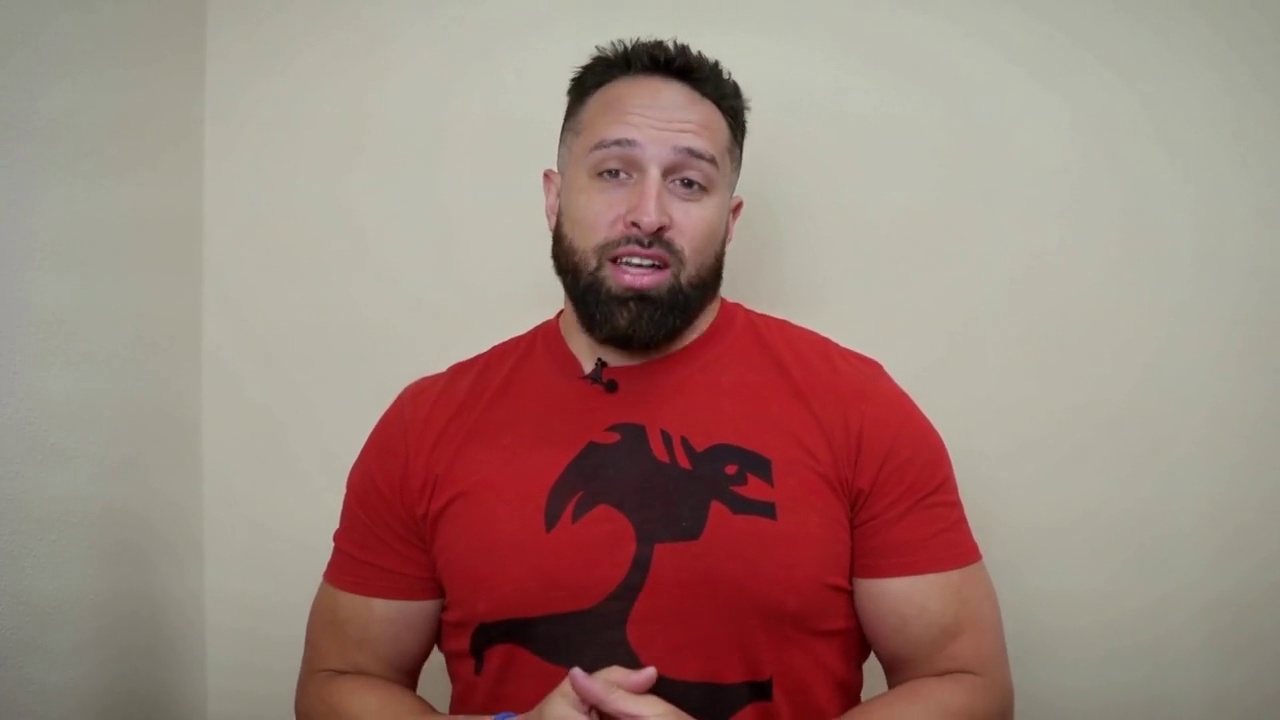}
&
\includegraphics[width=0.155\textwidth]{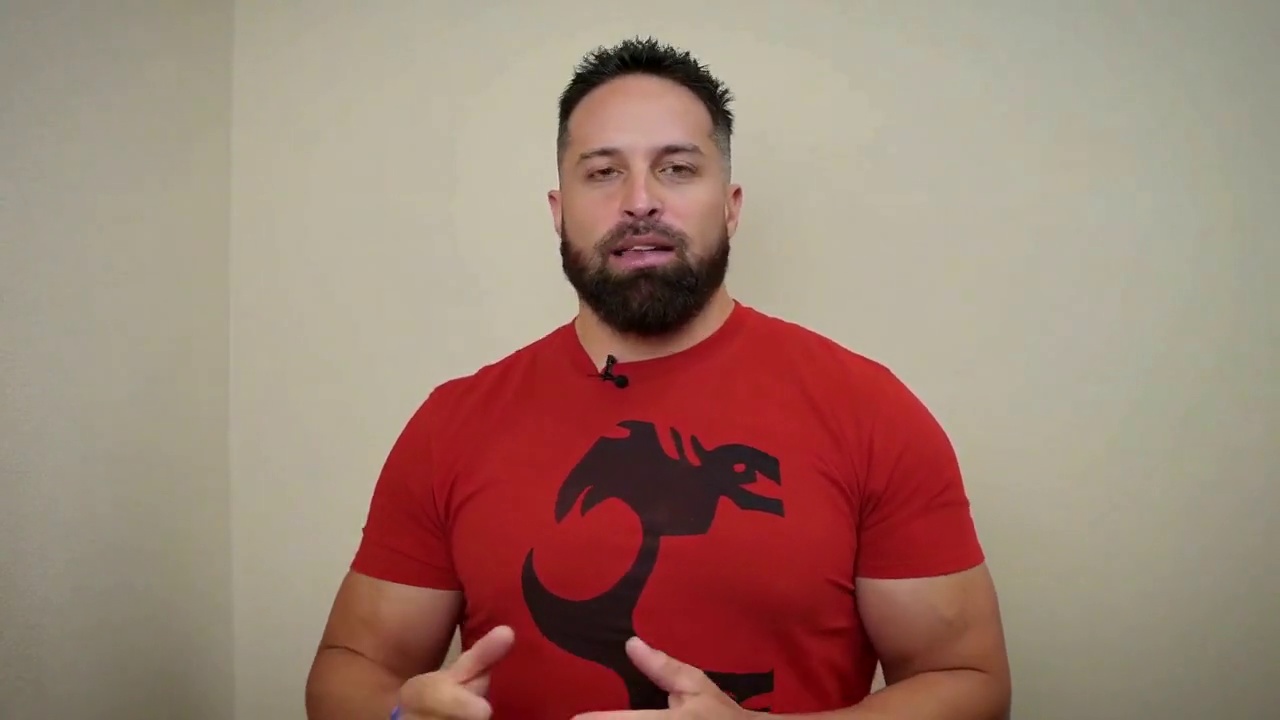}
&
\includegraphics[width=0.155\textwidth]{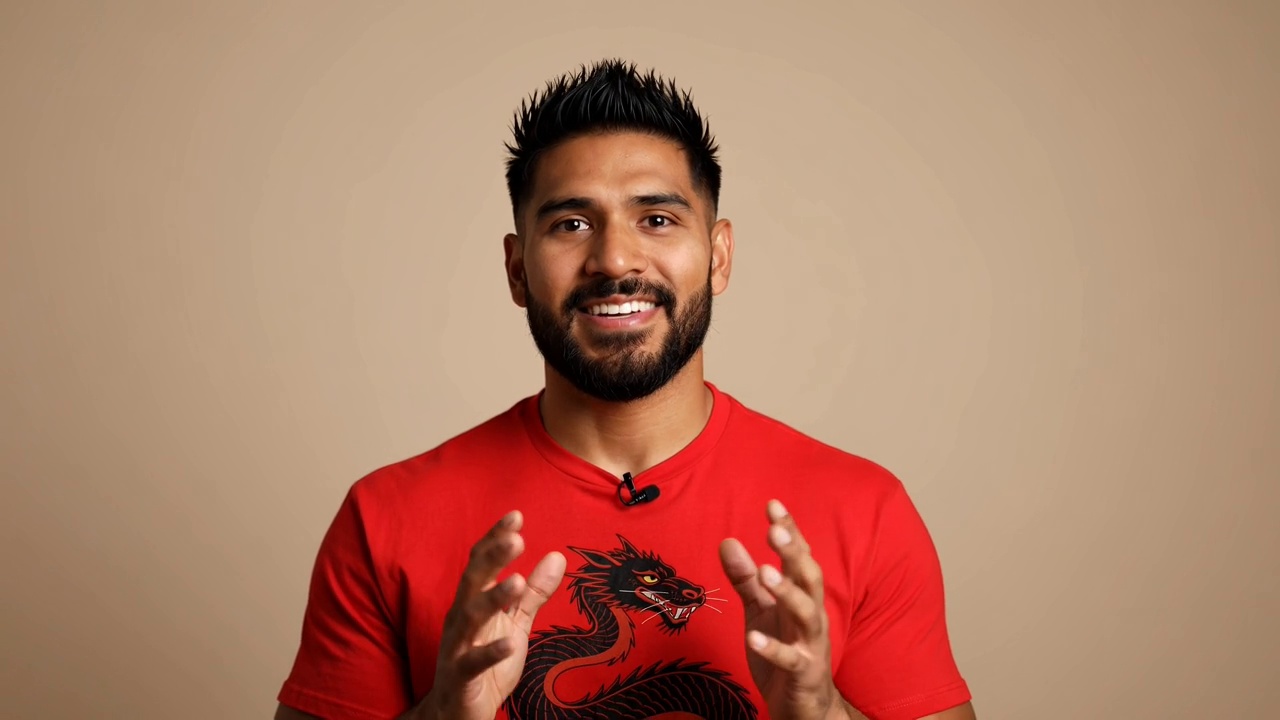}
&
\includegraphics[width=0.155\textwidth]{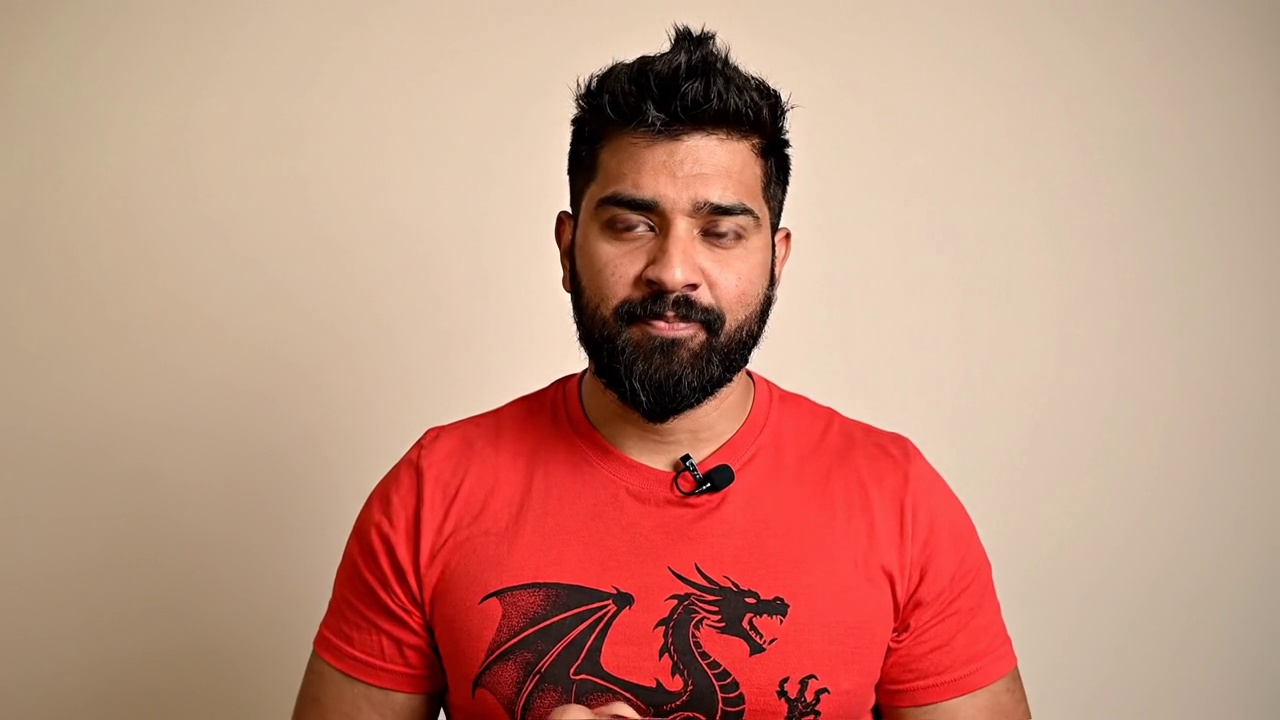}
\\[-1pt]
Real
& HunyuanVideo I2V
& LTX 2.3 I2V
& Wan 2.2 A14B I2V
& Grok Imagine 1.0
& Veo 3.1
\\[4pt]

&
\includegraphics[width=0.155\textwidth]{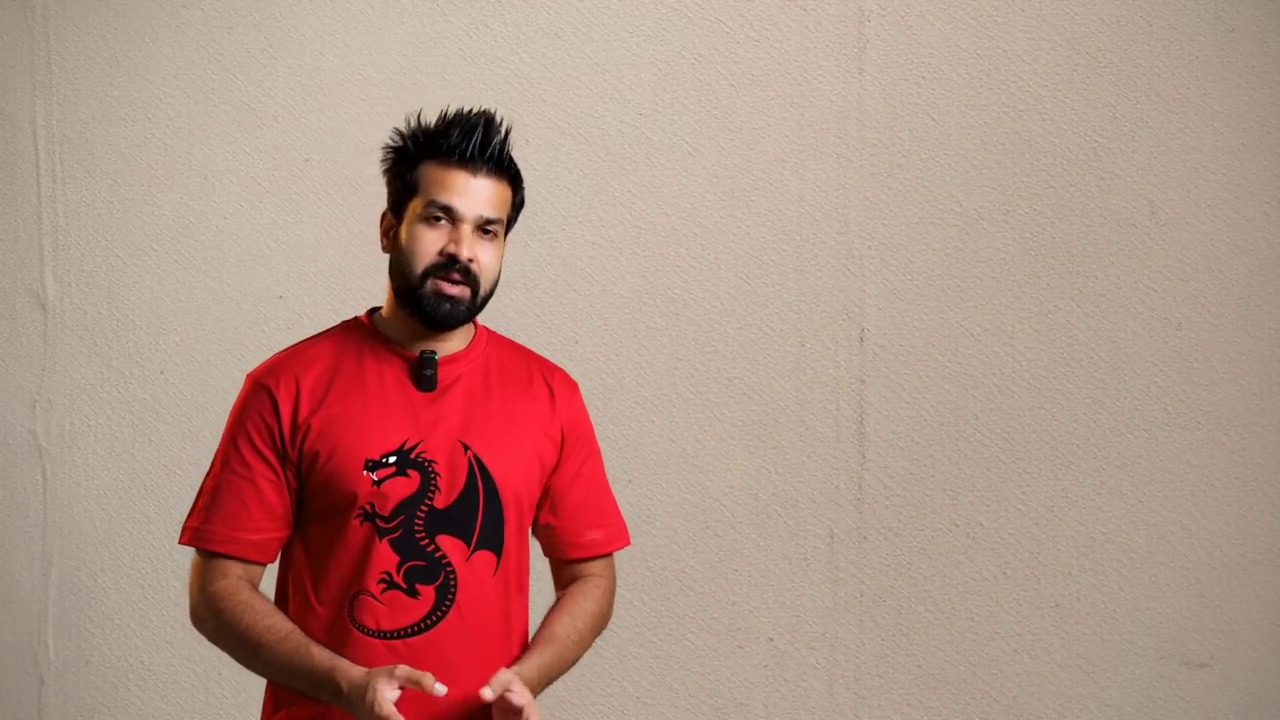}
&
\includegraphics[width=0.155\textwidth]{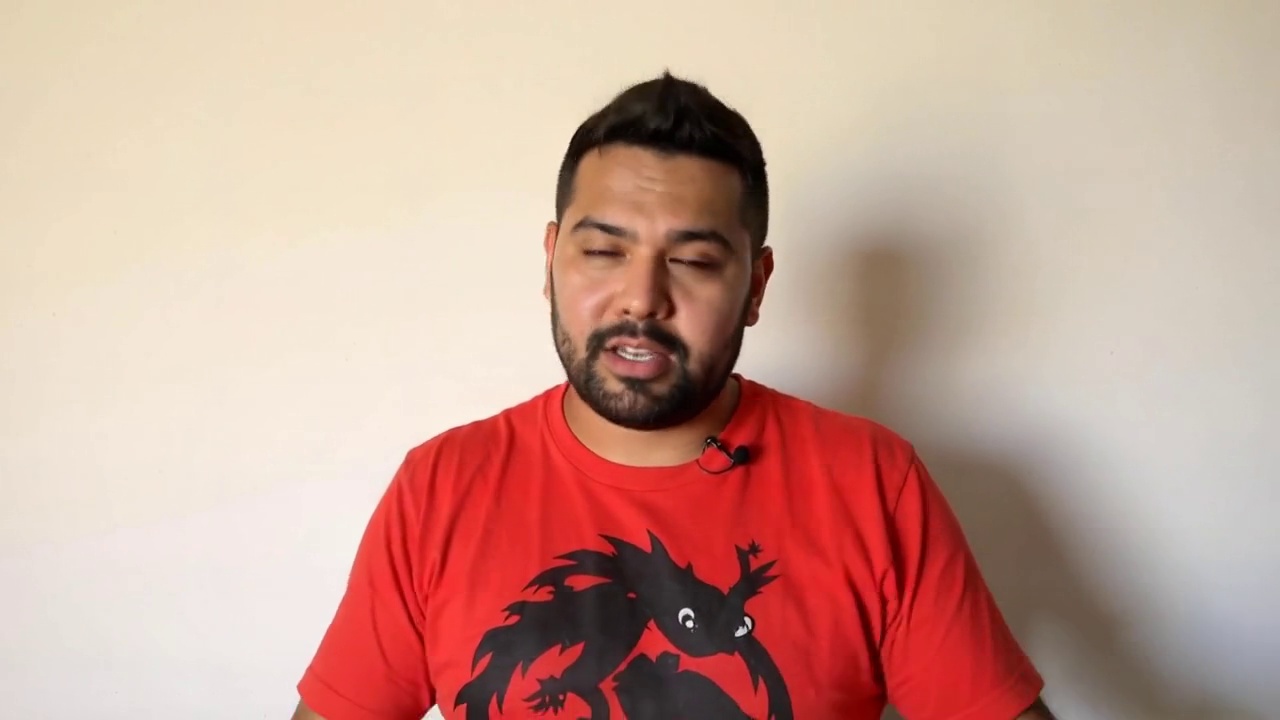}
&
\includegraphics[width=0.155\textwidth]{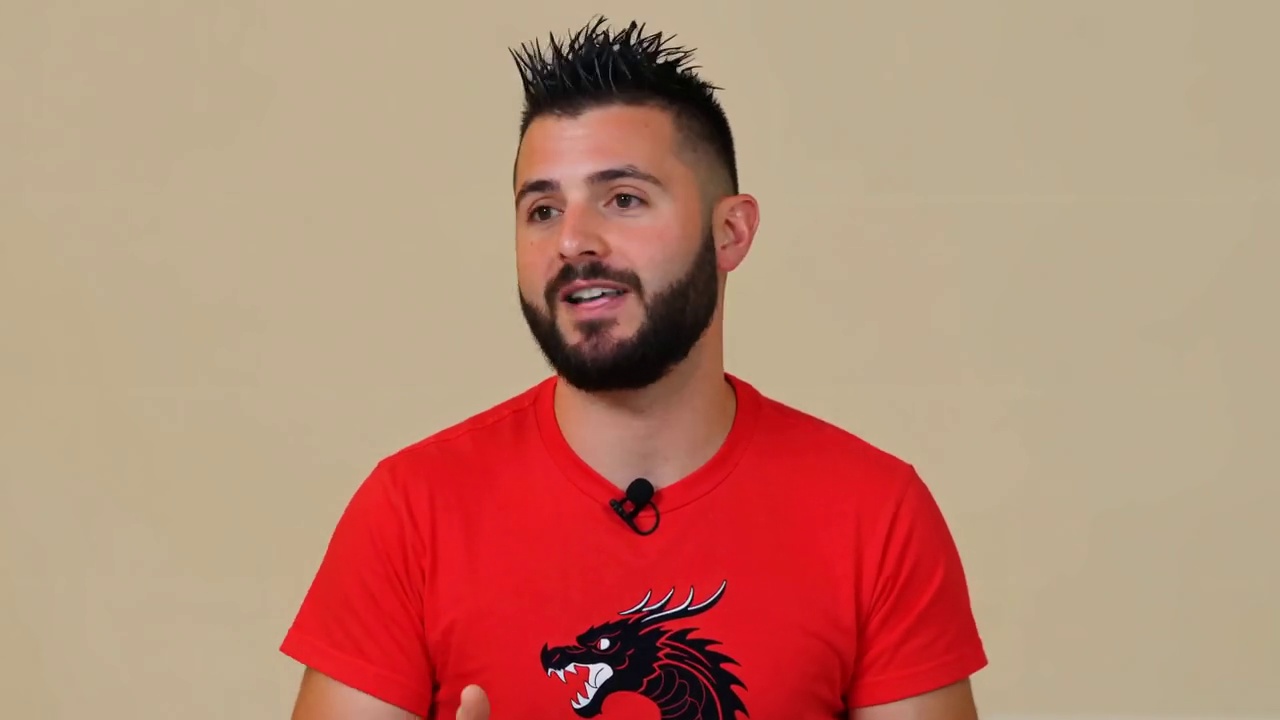}
&
\includegraphics[width=0.155\textwidth]{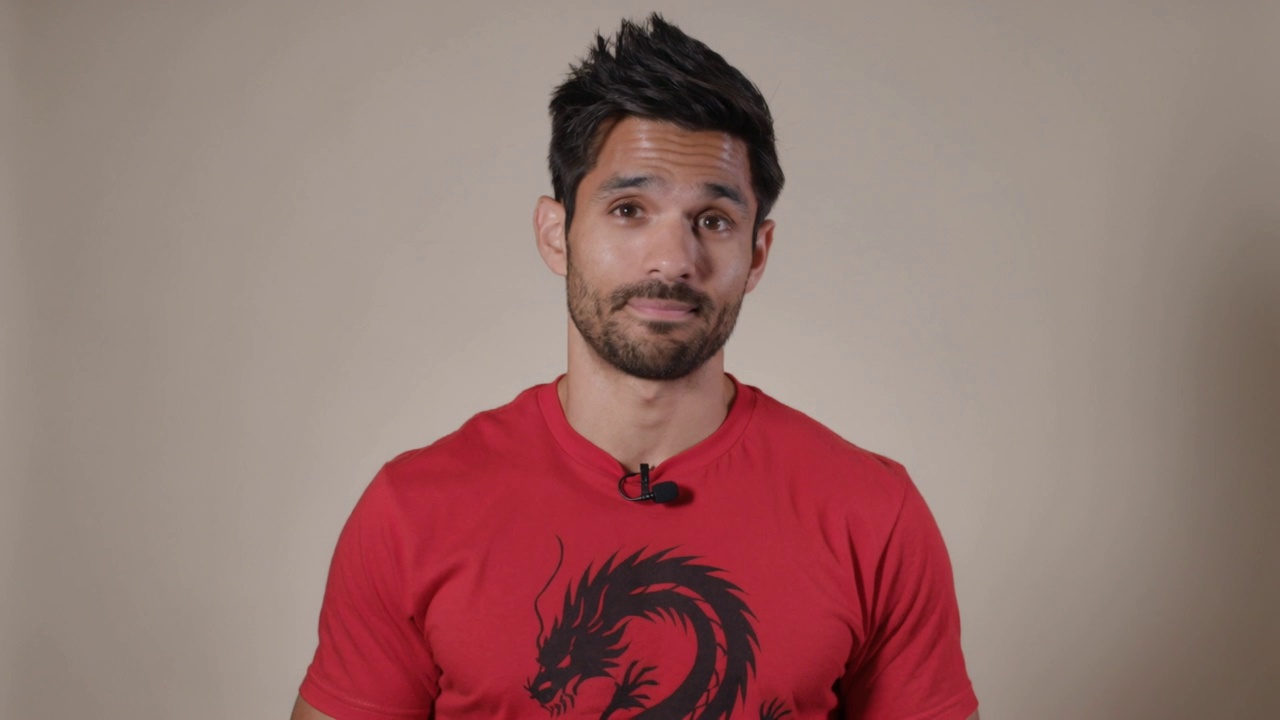}
&
\includegraphics[width=0.155\textwidth]{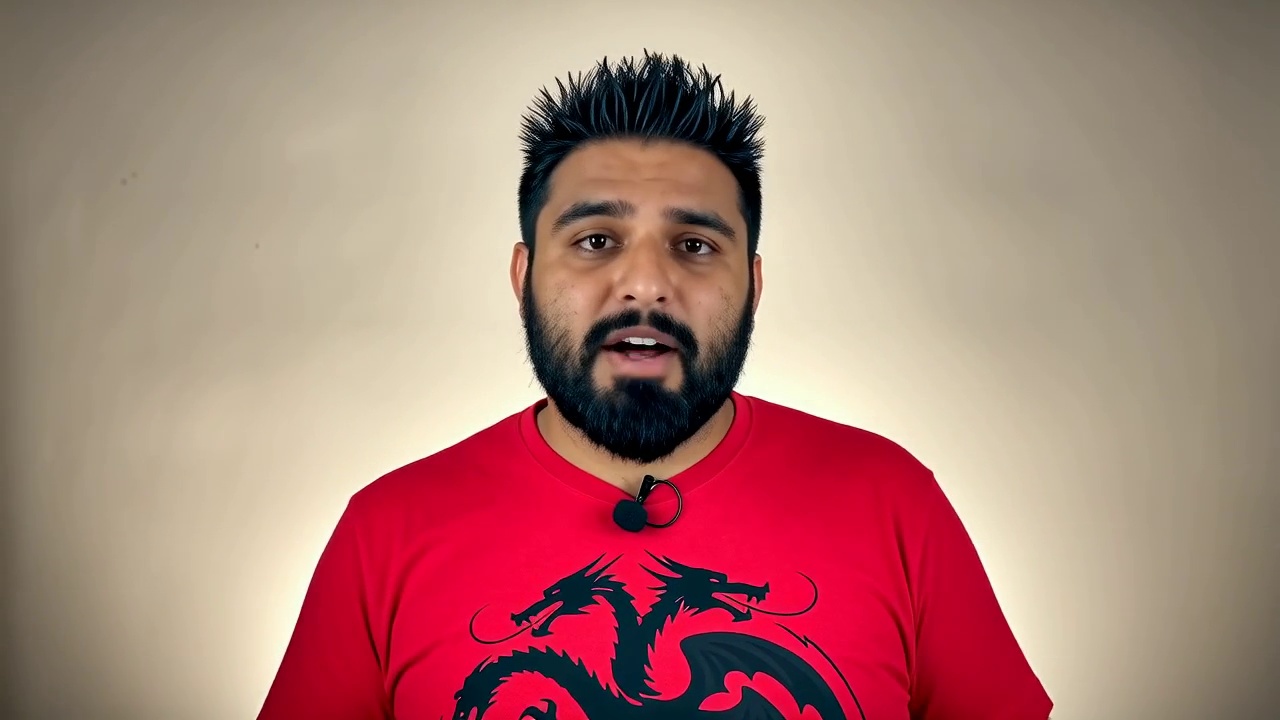}
\\[-1pt]
& HunyuanVideo T2V
& LTX 2.3 T2V
& Wan 2.2 A14B T2V
& Kling 3.0
& Wan 2.6
\\

\end{tabular}

\captionof{figure}{Last-frame comparison between real videos and videos generated by different models. Prompts used for video generation from text.
Example 1: ``A medium shot of a woman with long, wavy blonde hair, fair skin, and blue eyes, wearing a cream-colored, textured sweater and large hoop earrings. She is seated in a dark red chair in a studio with a purple backdrop. The woman is actively speaking, gesturing with her hands, and her mouth is moving dynamically. The lighting is bright and even, typical of a television studio, with soft shadows.''
Example 2: ``A man with short, dark, spiky hair, a full beard, and medium skin tone is speaking directly to the camera. He is wearing a bright red t-shirt with a black dragon graphic on the front and a small black lavalier microphone clipped to the collar. He gestures with his hands as he talks, his mouth moving naturally with his speech, and his head makes slight movements. The background is a plain, light beige wall with soft, even studio lighting that casts minimal shadows. The camera is at a medium shot, framing him from the chest up, and remains static throughout the clip.''}
\label{fig:last-frame-comparison-more-examples-df26}
\end{strip}

\section{Additional samples from \DatasetName}

\Cref{fig:last-frame-comparison-more-examples-df26} shows two more real videos alongside their generated counterparts from all seven generators in both I2V and T2V modes where supported. As in \cref{fig:t2v_examples}, all frames are taken from the last frame of each clip where the drift from the conditioning frame is the largest. The full text prompts used for the T2V generations are given in the caption, illustrating the level of scene, appearance, and camera detail produced by the prompt template described in Stage D.

\clearpage

\section{Results for temporal-based detectors reported  along with bootstrapped 95\% CI}

To quantify the uncertainty induced by the size of DF26, we report bootstrapped 95\% confidence intervals for the two temporal detectors. Intervals are obtained by resampling videos with replacement (1,000 resamples). \Cref{tab:df26_overall} provides the pooled metrics, \cref{tab:df26_per_generator} the per-generator breakdown, and \cref{tab:df26_per_video_type} presents the breakdown by semantic scenario. The intervals are narrow enough that the per-generator spread reported in \cref{sec:per-generator-amalysis} is not attributable to sampling noise: for PwTF-DVD, the interval on Wan 2.2 T2V does not overlap the interval on HunyuanVideo 1.5 in either mode. Per-scenario intervals overlap substantially for DFD-FCG, consistent with the claim that failures are driven by generator shift rather than by the public-speaking scenario.

\begin{table}[h]
\centering
\caption{AUROC (\%) pooled across all samples and EER (\%) metrics for DFD-FCG~\cite{han2025towards} and PwTF-DVD~\cite{kim2025beyond} on \DatasetName. Bootstrapped 95\% confidence intervals are shown in brackets.}
\label{tab:df26_overall}
\begin{tabular}{lcc}
\toprule
\textbf{Detector} & \textbf{AUROC} $\uparrow$ & \textbf{EER} $\downarrow$ \\
\midrule
DFD-FCG~\cite{han2025towards} 
& \ci{49.8}{46.4, 53.5} 
& \ci{49.9}{46.7, 53.7} \\
PwTF-DVD~\cite{kim2025beyond} 
& \ci{\textbf{63.1}}{60.1, 66.2} 
& \ci{\textbf{41.0}}{38.1, 43.9} \\
\bottomrule
\end{tabular}
\end{table}

\begin{table}[h]
\centering
\caption{Per-generator pooled AUROC (\%) and EER (\%) metrics for DFD-FCG~\cite{han2025towards} and PwTF-DVD~\cite{kim2025beyond} on \DatasetName. Bootstrapped 95\% confidence intervals are shown in brackets.}
\label{tab:df26_per_generator}
\resizebox{\linewidth}{!}{%
\begin{tabular}{clcccc}
\toprule
\multirow{2}{*}{\textbf{Type}} & \multirow{2}{*}{\textbf{Generator}} 
& \multicolumn{2}{c}{\textbf{DFD-FCG}} 
& \multicolumn{2}{c}{\textbf{PwTF-DVD}} \\
\cmidrule(lr){3-4} \cmidrule(lr){5-6}
& & \textbf{AUROC} $\uparrow$ & \textbf{EER} $\downarrow$ & \textbf{AUROC} $\uparrow$ & \textbf{EER} $\downarrow$ \\
\midrule
\multirow{3}{*}{I2V}
& HunyuanVideo 1.5 
& \ci{41.1}{36.6, 45.8} 
& \ci{55.0}{50.7, 59.1} 
& \ci{41.9}{37.1, 46.6} 
& \ci{55.0}{50.9, 59.6} \\
& LTX 2.3 distilled 
& \ci{61.1}{56.2, 65.8} 
& \ci{\textbf{43.0}}{38.9, 48.2} 
& \ci{79.5}{75.6, 83.1} 
& \ci{29.5}{25.6, 33.6} \\
& Wan 2.2 A14B 
& \ci{\textbf{61.3}}{56.6, 65.7} 
& \ci{43.7}{39.7, 48.5} 
& \ci{\textbf{87.7}}{84.6, 90.4} 
& \ci{\textbf{19.0}}{16.2, 23.1} \\
\midrule
\multirow{7}{*}{T2V}
& Grok Imagine 1.0 
& \ci{20.5}{16.7, 24.6} 
& \ci{74.0}{69.2, 77.2} 
& \ci{34.5}{29.7, 39.3} 
& \ci{58.7}{54.9, 63.6} \\
& HunyuanVideo 1.5 
& \ci{\textbf{76.5}}{72.4, 80.2} 
& \ci{\textbf{32.8}}{28.6, 36.7} 
& \ci{56.7}{51.9, 61.7} 
& \ci{42.8}{38.6, 48.2} \\
& Kling 3.0 
& \ci{27.4}{23.1, 32.2} 
& \ci{64.6}{60.3, 69.1} 
& \ci{60.4}{55.1, 65.3} 
& \ci{43.1}{38.0, 47.4} \\
& LTX 2.3 distilled 
& \ci{45.9}{41.1, 50.7} 
& \ci{52.2}{47.6, 56.3} 
& \ci{64.0}{59.6, 68.3} 
& \ci{41.3}{37.3, 45.4} \\
& Veo 3.1 
& \ci{29.6}{25.2, 34.2} 
& \ci{64.5}{60.0, 68.6} 
& \ci{26.7}{22.2, 31.0} 
& \ci{68.8}{64.3, 72.6} \\
& Wan 2.2 A14B 
& \ci{58.4}{53.4, 62.9} 
& \ci{44.8}{41.3, 50.2} 
& \ci{\textbf{92.9}}{90.5, 94.9} 
& \ci{\textbf{15.1}}{12.0, 18.4} \\
& Wan 2.6 
& \ci{60.3}{55.3, 65.1} 
& \ci{43.8}{39.1, 48.5} 
& \ci{71.6}{66.8, 76.1} 
& \ci{34.6}{30.6, 39.4} \\
\bottomrule
\end{tabular}
}
\end{table}
\begin{table}[h]
\centering
\caption{Per-video-type pooled AUROC (\%) and EER (\%) metrics for DFD-FCG~\cite{han2025towards} and PwTF-DVD~\cite{kim2025beyond} on \DatasetName. Bootstrapped 95\% confidence intervals are shown in brackets.}
\label{tab:df26_per_video_type}
\resizebox{\linewidth}{!}{%
\begin{tabular}{lcccc}
\toprule
\multirow{2}{*}{\textbf{Video Type}} 
& \multicolumn{2}{c}{\textbf{DFD-FCG}} 
& \multicolumn{2}{c}{\textbf{PwTF-DVD}} \\
\cmidrule(lr){2-3} \cmidrule(lr){4-5}
& \textbf{AUROC} $\uparrow$ & \textbf{EER} $\downarrow$ & \textbf{AUROC} $\uparrow$ & \textbf{EER} $\downarrow$ \\
\midrule
Direct to Camera
& \ci{50.6}{44.8, 56.6} 
& \ci{50.1}{44.0, 55.2} 
& \ci{60.1}{54.7, 65.6} 
& \ci{43.9}{38.2, 48.1} \\
Official Statement 
& \ci{\textbf{51.7}}{44.6, 59.3} 
& \ci{53.2}{46.5, 59.0} 
& \ci{\textbf{72.3}}{66.4, 78.0} 
& \ci{\textbf{34.9}}{29.2, 39.2} \\
Studio Interview 
& \ci{50.5}{44.9, 56.3} 
& \ci{\textbf{48.1}}{42.8, 54.2} 
& \ci{66.1}{61.1, 71.3} 
& \ci{39.1}{34.5, 43.4} \\
\bottomrule
\end{tabular}
}
\end{table}

\section{Technical statistics, audio and encoding.}
Real clips were trimmed to 5\,s from their source videos and re-encoded to 720p
H.264. Generated clips are
released exactly as each model or provider produced them: commercial clips keep the
providers' own encoding, and open-weight clips were encoded once by the generation
pipeline. Although they come from different encoders, all clips share the same H.264
level, keyframe structure, and constant frame rate. Frame rates, durations, and
bitrates are not equalized across sources (Table~\ref{tab:supp_stats}); in
particular, open-weight clips have 3--4$\times$ lower bitrates than real clips.
Audio is not part of \DatasetName, which is a visual-only benchmark. The released
files still carry audio tracks in commercial clips and in 18 real clips, but not in
open-weight clips. Users of \DatasetName should discard audio tracks, since their
presence alone separates most commercial clips from real ones. No evaluated detector
uses audio, and clips were muted in the human study.

\begin{table}[t]
\centering
\footnotesize
\setlength{\tabcolsep}{3.5pt}
\caption{Per-source technical statistics. All clips: 1280$\times$720 progressive 8-bit
H.264, level~3.1, constant frame rate, single keyframe. Real clips were trimmed and
re-encoded by us; generated clips are as produced by each model or provider.
Duration = frame count / fps. Bitrate (Mbps) is that of the video stream: mean
[2.5th--97.5th percentile]. Audio = number of clips with an audio track in the release.}
\label{tab:supp_stats}
\begin{tabular}{@{}lrrrlr@{}}
\toprule
Source & $n$ & fps & Dur.\ (s) & Bitrate (Mbps) & Audio \\
\midrule
Real        & 271 & mixed & 5.00       & 3.26 [0.88, 6.39]  & 18 \\
\midrule
Veo 3.1     & 200 & 24       & 4.00 & 2.91 [1.71, 5.42]  & 200 \\
Kling 3.0   & 198 & 24       & 5.04       & 4.88 [3.48, 7.06]  & 187 \\
Grok 1.0    & 199 & 24       & 6.04       & 2.59 [1.63, 4.46]  & 199 \\
Wan 2.6     & 197 & 30       & 5.00       & 4.50 [2.04, 11.65] & 197 \\
\midrule
HV 1.5 I2V  & 271 & 24 & 5.04       & 0.78 [0.42, 1.53]  & 0 \\
HV 1.5 T2V  & 271 & 24       & 5.04       & 1.21 [0.68, 2.01]  & 0 \\
LTX 2.3 I2V & 271 & 24       & 5.04       & 0.91 [0.47, 1.66]  & 0 \\
LTX 2.3 T2V & 271 & 24       & 5.04       & 1.03 [0.56, 1.87]  & 0 \\
Wan 2.2 I2V & 271 & 16       & 5.06       & 0.70 [0.35, 1.44]  & 0 \\
Wan 2.2 T2V & 271 & 16       & 5.06       & 0.71 [0.40, 1.27]  & 0 \\
\bottomrule
\end{tabular}
\end{table}
\clearpage

\end{document}